\newif\iffinaldraft
\documentclass[a4paper,fleqn]{cas-sc}
\usepackage[authoryear]{natbib}

\usepackage{tipa}
\usepackage{amsmath, amssymb}   % Core AMS packages
\usepackage{bm}                 % Bold math
\usepackage{amssymb}

\usepackage{graphicx}
\usepackage{tikz}
\usepackage{tikz-3dplot}
\usepackage{pgfplots}
\pgfplotsset{compat=1.16}

\usetikzlibrary{
  decorations.pathreplacing,
  shapes.geometric,
  arrows.meta,
  positioning,
  calc
}

\usepackage{hyperref}
\usepackage{url}
\usepackage{footmisc}

\usepackage{booktabs}
\usepackage{multirow}
\usepackage{makecell}
\usepackage{threeparttable}
\usepackage{adjustbox}

\usepackage[justification=justified,singlelinecheck=false]{caption}
\usepackage{subcaption}
\usepackage{pdflscape}
\usepackage{rotating}
\usepackage{float}
\usepackage{placeins}

\usepackage{xcolor}
\usepackage[dvipsnames]{xcolor}
\usepackage{soul}
\usepackage{xspace}
\usepackage{fancyvrb}
\usepackage{enumitem}
\usepackage{cancel}
\usepackage{microtype}

\usepackage{listings}
\pgfplotsset{compat=1.17}
\tdplotsetmaincoords{60}{120}

\def\tsc#1{\csdef{#1}{\textsc{\lowercase{#1}}\xspace}}
\tsc{WGM}
\tsc{QE}
\tsc{EP}
\tsc{PMS}
\tsc{BEC}
\tsc{DE}
\usepackage{pifont}

\newcommand{\xmark}{\ding{55}} % Define \xmark for clarity
\usepackage{tikz}
\usepackage{tikz-3dplot}
\usetikzlibrary{arrows.meta,positioning,fit,backgrounds}

\newcommand{\Ocomplex}{\mathcal{O}}

\usetikzlibrary{shapes, arrows.meta, positioning, shapes.geometric}

\newcount\Comments  % 1 suppresses notes to selves in text
\usepackage{color}
\definecolor{darkgreen}{rgb}{0,0.5,0}
\definecolor{purple}{rgb}{1,0,1}
\usepackage{amsmath}

\newcommand{\kibitz}[2]{\ifnum\Comments=0\textcolor{#1}{#2}\fi}

\newcommand{\midbar}[1]{\mkern 1.5mu\overline{\mkern-2mu #1 \mkern-2mu}\mkern 1.5mu}
\newcommand{\braceannotation}[4][160mu]{ % Multiline brace annotations for equations
  \smash{\mkern#1\raisebox{#2\baselineskip}{$\left.\rule{0pt}{#3\baselineskip}\right\}{\text{\small #4}}$}}%
}

\newcommand{\set}[1]{\mathcal{#1}}
\newcommand{\spacenodeset}{\Omega_{{S}}}
\newcommand{\timenodeset}{\Omega_{{T}}}
\newcommand{\adjmat}{\asmatrix{A}}

\newcommand{\asmatrix}[1]{\mathbf{\MakeUppercase{#1}}}
\newcommand{\asvector}[1]{\mathbf{\MakeLowercase{#1}}}
\newcommand{\tensor}[1]{\bm{\uppercase{\mathcal{#1}}}\xspace}
\newcommand{\gtensor}[1]{\boldsymbol{\mathit{#1}}}

\newcommand{\kten}[1]{\mathcal{K}^{#1}_{10}}
\newcommand{\vecop}{\text{Vec}\xspace}

\newcommand{\sendingset}{\Omega_{s}} % Set of sending nodes
\newcommand{\receivingset}{\Omega_{r}} % Set of receiving nodes

\newcommand{\hkmv}[1]{$\kappa^{#1}_{\text{KMV}}$}

\newcommand{\hrightrecur}{$\kappa_{R}^{\Delta}$} 
\newcommand{\hleftrecur}{$\kappa_{L}^{\Delta}$}

\newcommand{\depth}[1]{\langle#1\rangle}

\newcommand{\nodeset}{\mathcal{M}}
\newcommand{\node}{\mathit{m}}
\newcommand{\hi}{\uparrow}
\newcommand{\lo}{\downarrow}

\newcommand{\spaceset}{\mathcal{S}}
\newcommand{\timesset}{\mathcal{T}}
\newcommand{\stset}{\spaceset\timesset}

\newcommand{\spacenode}{\varsigma}
\newcommand{\timesnode}{\tau}

\newcommand{\attn}{\Theta}
\newcommand{\graph}{\mathcal{G}}

\newcommand{\spaceattn}{\boldsymbol{\Theta}_{\spaceset}}
\newcommand{\timesattn}{\boldsymbol{\Theta}_{\timesset}}

\newcommand{\dotkernel}{\varphi}

\newcommand{\dotumble}{\looparrowleft}

\newcommand{\attnspace}{\Theta_{\spaceset}}
\newcommand{\attntimes}{\Theta_{\timesset}}
\newcommand{\attnst}{\Theta_{\spaceset \timesset}}

\newcommand{\glu}{\text{GLU}}
\newcommand{\entmax}{\text{Entmax}}

\newcommand{\dropout}{\text{Dropout}}

\newcommand\midtilde[1]{%
  \mathord{\stackrel{\scalebox{1.1}[1]{\(\sim\)}}{\smash{#1}\rule{0pt}{0.85ex}}}%
}

\newcommand{\Langle}{\mathopen{\pmb{[}\kern-0.28em[}}
\newcommand{\Rangle}{\mathclose{]\kern-0.25em\pmb{]}}}

\newcommand{\LLangle}{\mathopen{\pmb{\big[}\kern-0.32em\big[}}
\newcommand{\RRangle}{\mathclose{\big]\kern-0.28em\pmb{\big]}}}

\newcommand{\LLLangle}{\mathopen{\pmb{\Big[}\kern-0.38em\Big[}}
\newcommand{\RRRangle}{\mathclose{\Big]\kern-0.35em\pmb{\Big]}}}

\newcommand{\perror}[1]{\mathit{(\pm #1)}}
\newcommand{\unperror}[2]{\underset{\perror{#2}}{#1}}
\newcommand{\subperror}[2]{{#1}_{\perror{#2}}}

\tikzset{
    block/.style = {
        rectangle,
        draw,
        minimum height=2em,
        minimum width=3em,
        align=center
    },
    oval/.style = {
        ellipse,
        draw,
        minimum height=2em,
        minimum width=3em,
        align=center
    },
    box/.style = {
        rectangle,
        draw,
        text width=4em,
        text centered,
        minimum height=3em
    },
    arr/.style = {
        -Stealth,
        thick
    },
    rounded/.style = {
        rectangle,
        draw,
        rounded corners,
        minimum height=2em,
        minimum width=3em,
        align=center
    },
    circle node/.style = {
        circle,
        draw,
        minimum size=#1,
        inner sep=0pt % ensures that the circle size closely matches the specified minimum size
    },
    circle node/.default = 0.5cm,
  mydiamond/.style={
    draw,
    diamond,
    minimum width=#1,
    minimum height=#1,
    inner sep=0pt
  },
}

\newcommand{\drawThreeDNodeLineWithPlane}[7]{
    \def\M{#1}
    \def\N{#2}
    \def\T{#3}
    \def\scale{#4}
    \def\separation{#5}
    \def\curveheight{#6}
    \def\planedepth{#7}
    \begin{tikzpicture}[tdplot_main_coords, scale=\scale]
    \foreach \i in {1,...,\M} {
        \coordinate (N\i) at (0, \i * \separation, 0);
    }
    \foreach \i in {1,...,\N} {
        \foreach \j in {1,...,\T} {
            \coordinate (P\i\j) at (\j * \separation, \i * \separation, -\planedepth);
        }
    }
    
    \foreach \i in {1,...,\N} {
        \foreach \j in {1,...,\T} {
            \foreach \k in {1,...,\N} {
                \foreach \l in {1,...,\T} {
                    \ifnum\i\j<\k\l
                        \pgfmathsetmacro{\midx}{(\j+\l)/2*\separation}
                        \pgfmathsetmacro{\midy}{(\i+\k)/2*\separation}
                        \pgfmathsetmacro{\midz}{-\planedepth-0.8*\separation}
                        \draw[magenta, opacity=0.95, line width=0.6pt] 
                            (P\i\j) .. controls (\midx, \midy, \midz) .. (P\k\l);
                    \fi
                }
            }
        }
    }
    
    \foreach \i in {1,...,\M} {
        \foreach \j in {\i,...,\M} {
            \ifnum\i<\j
                \pgfmathsetmacro{\midpoint}{(\i+\j)/2}
                \draw[orange, line width=0.8pt] 
                    (N\i) .. controls (0, \midpoint * \separation, \curveheight * \separation) 
                    .. (N\j);
            \fi
        }
    }
    
    \foreach \i in {1,...,\N} {
        \foreach \j in {1,...,\T} {
            \filldraw[blue, draw=black] (P\i\j) circle (2pt);
        }
    }
    
    \foreach \i in {1,...,\M} {
        \filldraw[red] (N\i) circle (2pt);
    }
    
    \foreach \i in {1,...,\M} {
        \foreach \j in {1,...,\N} {
            \foreach \k in {1,...,\T} {
                \draw[green!60!black, opacity=1, line width=0.8] (N\i) -- (P\j\k);
            }
        }
    }
    
    \draw[->, >=stealth, thick] (0, 0.5 * \separation, \curveheight * \separation + 0.2) 
        -- (0, \M * \separation + 0.5 * \separation, \curveheight * \separation + 0.2) 
        node[midway, above, sloped] {time flow};

    \draw[->, >=stealth, thick, black] 
        (1.5, -1.8+0.5 * \M * \separation, 0.5 * \separation) 
        -- 
        (1 * \T * \separation, -0.7+0.5 * \N * \separation, -\planedepth + 0.5 * \separation);
    
    \end{tikzpicture}
}

\newcommand{\drawBlueNodesWithMagentaArcs}[6]{
    \def\N{#1}
    \def\T{#2}
    \def\scale{#3}
    \def\separation{#4}
    \def\curveheight{#5}
    \def\planedepth{#6}
    \begin{tikzpicture}[tdplot_main_coords, scale=\scale]
    \foreach \i in {1,...,\N} {
        \foreach \j in {1,...,\T} {
            \coordinate (P\i\j) at (\j * \separation, \i * \separation, -\planedepth);
        }
    }
    
    \foreach \i in {1,...,\N} {
        \foreach \j in {1,...,\T} {
            \foreach \k in {1,...,\N} {
                \foreach \l in {1,...,\T} {
                    \ifnum\i\j<\k\l
                        \pgfmathsetmacro{\midx}{(\j+\l)/2*\separation}
                        \pgfmathsetmacro{\midy}{(\i+\k)/2*\separation}
                        \pgfmathsetmacro{\midz}{-\planedepth-0.8*\separation}
                        \draw[magenta, opacity=0.95, line width=0.6pt] 
                            (P\i\j) .. controls (\midx, \midy, \midz) .. (P\k\l);
                    \fi
                }
            }
        }
    }
    
    \foreach \i in {1,...,\N} {
        \foreach \j in {1,...,\T} {
            \filldraw[blue, draw=black] (P\i\j) circle (2pt);
        }
    }
    \end{tikzpicture}
}

\newcommand{\drawRedNodesWithOrangeArcs}[5]{
    \def\M{#1}
    \def\scale{#2}
    \def\separation{#3}
    \def\curveheight{#4}
    \def\planedepth{#5}
    \begin{tikzpicture}[tdplot_main_coords, scale=\scale]
    \foreach \i in {1,...,\M} {
        \coordinate (N\i) at (0, \i * \separation, 0);
    }
    
    \foreach \i in {1,...,\M} {
        \foreach \j in {\i,...,\M} {
            \ifnum\i<\j
                \pgfmathsetmacro{\midpoint}{(\i+\j)/2}
                \draw[orange, line width=0.8pt] 
                    (N\i) .. controls (0, \midpoint * \separation, \curveheight * \separation) 
                    .. (N\j);
            \fi
        }
    }
    
    \foreach \i in {1,...,\M} {
        \filldraw[red] (N\i) circle (2pt);
    }
    
    \draw[->, >=stealth, thick] (0, 0.5 * \separation, \curveheight * \separation + 0.2) 
        -- (0, \M * \separation + 0.5 * \separation, \curveheight * \separation + 0.2) 
        node[midway, above, sloped] {time flow};
    
    \end{tikzpicture}
}

\newcommand{\drawAllToAllPlane}[5]{
    \def\T{#1}
    \def\N{#2}
    \def\scale{#3}
    \def\separation{#4}
    \def\nodecolor{#5}
    \begin{tikzpicture}[scale=\scale]
    \foreach \i in {1,...,\T} {
        \foreach \j in {1,...,\N} {
            \coordinate (N\i\j) at (\i * \separation, \j * \separation);
        }
    }
    
    \foreach \i in {1,...,\T} {
        \foreach \j in {1,...,\N} {
            \foreach \ii in {1,...,\T} {
                \foreach \jj in {1,...,\N} {
                    \ifnum\i=\ii
                        \ifnum\j=\jj
                        \else
                            \draw[gray, line width=0.1pt] (N\i\j) -- (N\ii\jj);
                        \fi
                    \else
                        \draw[gray, line width=0.1pt] (N\i\j) -- (N\ii\jj);
                    \fi
                }
            }
        }
    }
    
    \foreach \i in {1,...,\T} {
        \foreach \j in {1,...,\N} {
            \filldraw[\nodecolor] (N\i\j) circle (2pt);
        }
    }
    
    \foreach \i in {1,...,\T} {
        \ifnum\i=1
            \node[above=5pt] at (N\i\N) {t=0};
        \else
            \node[above=5pt] at (N\i\N) {\the\numexpr\i-1\relax};
        \fi
    }
    
    \foreach \j in {1,...,\N} {
        \node[right=5pt] at (N\T\j) {node \the\numexpr\N-\j\relax};
    }
    
    \draw[<->, line width=0.5pt] (0.75*\separation, 0.75*\separation) -- (\T*\separation + 0.25*\separation, 0.75*\separation);
    \node[font=\footnotesize] at ({\T*\separation/2 + 0.5*\separation}, 0.25*\separation) {Time steps};
    
    \draw[<->, line width=0.5pt] (0.75*\separation, 0.75*\separation) -- (0.75*\separation, \N*\separation + 0.25*\separation);
    \node[rotate=90, font=\footnotesize] at (0.25*\separation, {\N*\separation/2 + 0.5*\separation}) {Nodes};
    
    \end{tikzpicture}
}

\newcommand{\drawTwoRedNodesWithBlueNodePlane}[6]{
    \def\N{#1}
    \def\T{#2}
    \def\scale{#3}
    \def\separation{#4}
    \def\curveheight{#5}
    \def\planedepth{#6}
    \begin{tikzpicture}[tdplot_main_coords, scale=\scale]
    \coordinate (N1) at (0, \separation, 0);
    \coordinate (N2) at (0, 2 * \separation, 0);
    \foreach \i in {1,...,\N} {
        \foreach \j in {1,...,\T} {
            \coordinate (P\i\j) at (\j * \separation, \i * \separation, -\planedepth);
        }
    }
    
    \foreach \i in {1,...,\N} {
        \foreach \j in {1,...,\T} {
            \foreach \k in {1,...,\N} {
                \foreach \l in {1,...,\T} {
                    \ifnum\i\j<\k\l
                        \pgfmathsetmacro{\midx}{(\j+\l)/2*\separation}
                        \pgfmathsetmacro{\midy}{(\i+\k)/2*\separation}
                        \pgfmathsetmacro{\midz}{-\planedepth-0.8*\separation}
                        \draw[magenta, opacity=0.95, line width=0.6pt] 
                            (P\i\j) .. controls (\midx, \midy, \midz) .. (P\k\l);
                    \fi
                }
            }
        }
    }
    
    \foreach \i in {1,...,\N} {
        \foreach \j in {1,...,\T} {
            \draw[-{Stealth[length=2mm,width=1.5mm]}, black, opacity=1, line width=0.8, shorten >=3pt, dashed] (N2) -- (P\i\j);
        }
    }
    
    \foreach \i in {1,...,\N} {
        \foreach \j in {1,...,\T} {
            \filldraw[blue, draw=black] (P\i\j) circle (2pt);
        }
    }
    
    \filldraw[red] (N1) circle (2pt) node[left=3pt, black] {$\tau_i$};
    \filldraw[red] (N2) circle (2pt) node[left=3pt, black] {$\tau_j$};
    
    \draw[-{Stealth[length=3mm,width=2mm]}, black, line width=0.8pt] 
        (N1) .. controls (0, 1.5 * \separation, \curveheight * \separation) 
        .. (N2);
    
    \end{tikzpicture}
}

\newcommand{\drawkroneckerrepresentation}{
    \centering
    \begin{tikzpicture}
        \begin{scope}[local bounding box=kronecker_product]
            \draw (0,0) rectangle (8,8);
            
            \node at (1.15,7.7) {N};
            \node[rotate=90] at (0.25,6.8) {N};
            
            \fill[purple!60!blue] (0.5,6.2) rectangle +(1.3,1.3);  % First square solid green
            \foreach \x in {0,1,2}
                \foreach \y in {0,1,2}
                    \draw (0.5+\x*1.5,6.2-\y*1.5) rectangle +(1.3,1.3);
            \draw (6.2,6.2) rectangle +(1.3,1.3);
            \draw (0.5,0.5) rectangle +(1.3,1.3);
            \draw (6.2,0.5) rectangle +(1.3,1.3);
            
            \node at (5.6,7.1) {$\cdots$};  % Top horizontal
            \node at (4,0.8) {$\cdots$};  % Bottom horizontal
            \node at (1.1,2.4) {$\vdots$};  % Left vertical
            \node at (7.2,4) {$\vdots$};  % Right vertical
            \node at (5.5,2.5) {$\ddots$};  % Diagonal
            
            \node at (4,-0.5) {$P\cdot N$};
            \node[rotate=90] at (-0.5,4) {$P\cdot N$};
            \node at (4,8.5) {Kronecker Product};
        \end{scope}

        \node at (-7.3,5.5) {$\Theta_T$};
        \node at (-3.7,5.5) {$\Theta_S$};
        
        \draw (-6.5,2.915) rectangle +(-1.67,1.67);  % Left square outline
        \node[anchor=east] at (-8.3,3.75) {P};  % Left axis label
        \node[anchor=north] at (-7.335,2.815) {P};  % Bottom axis label
        
        \foreach \x in {0,...,5}
            \foreach \y in {0,...,5}
            {
                \ifnum\x=0
                    \ifnum\y=5
                        \fill[red] (-8.09+\x*0.26,2.99+\y*0.26) rectangle +(.24,.24);
                    \else
                        \fill[lightgray] (-8.09+\x*0.26,2.99+\y*0.26) rectangle +(.24,.24);
                    \fi
                \else
                    \fill[lightgray] (-8.09+\x*0.26,2.99+\y*0.26) rectangle +(.24,.24);
                \fi
            }
        
        \draw (-2.2,2.2) rectangle +(-2.9,2.9);  % Right square outline
        \node[anchor=east] at (-5.15,3.65) {N};  % Left axis label (adjusted position)
        \node[anchor=north] at (-3.65,2.1) {N};  % Bottom axis label
        
        \foreach \x in {0,...,20}
            \foreach \y in {0,...,20}
                \fill[blue] (-5.025+\x*0.13,2.325+\y*0.13) rectangle +(.11,.11);
        
        \node[scale=2] at (-6.07,3.75) {$\otimes$};

        \draw[dashed] (-8.9,1.5) rectangle (-1.9,5.8);
        
        \draw[->, thick] (-1.9,3.5) -- (1.2,6.8);
        
    \end{tikzpicture}
}

\begin{document}

\let\WriteBookmarks\relax
\def\floatpagepagefraction{1}
\def\textpagefraction{.001}
 
\shorttitle{Weaver: Kronecker Product Approximations of Spatiotemporal Attention for Traffic Network Forecasting} 
 
\shortauthors{C. Cheong, G. Davis, and S. Choi}
%\begin{frontmatter}

\title [mode = title]{Weaver: Kronecker Product Approximations of Spatiotemporal Attention for Traffic Network Forecasting}

% \tnotemark[1,2]

% \tnotetext[1]{This research was supported by Basic Science Research Program through the National Research Foundation of Korea(NRF) funded by the Ministry of  Science and ICT(NRF-2017R1A2B2002329)}

% \tnotetext[2]{The second title footnote which is a longer text matter
%   to fill through the whole text width and overflow into
%   another line in the footnotes area of the first page.}

\author[1]{Christopher Cheong}
\ead{cheon028@umn.edu}
\credit{Conceptualization of this study, Methodology, Formal analysis, Writing - original draft}

\author[1]{Gary Davis}
\ead{drtrips@umn.edu}
\credit{Methodology, Writing - original draft, Writing - review and editing}

\author[1]{Seongjin Choi}[orcid=0000-0001-7140-537X]
\ead{chois@umn.edu}
\cormark[1]
\credit{Conceptualization of this study, Methodology, Software, Validation, Formal analysis,  Writing - original draft, Writing - review and editing}

\address[1]{Department of Civil, Environmental, and Geo- Engineering, University of Minnesota, Twin Cities, 500 Pillsbury Drive S.E., Minneapolis, MN 55455-0116, United States of America}

\cortext[cor1]{Corresponding author}

\begin{abstract}
Spatiotemporal forecasting on transportation networks is a complex task that requires understanding how traffic nodes interact within a dynamic, evolving system dictated by traffic flow dynamics and social behavioral patterns. The importance of transportation networks and intelligent transportation systems (ITS) for modern mobility and commerce necessitates forecasting models that are not only accurate, but also interpretable, efficient, and robust under structural or temporal perturbations. Recent approaches, particularly Transformer-based architectures, have improved predictive performance but often at the cost of high computational overhead and diminished architectural interpretability.

In this work, we introduce Weaver, a novel attention-based model that applies \textbf{Kronecker product approximations (KPA)} to decompose the $PN\times PN$ spatiotemporal attention of $\Ocomplex(P^2N^2)$ complexity into local $P\times P$ temporal and $N\times N$ spatial attention maps. This \textbf{Kronecker attention map} enables our \textbf{Parallel-Kronecker Matrix-Vector product (P$^2$-KMV)} for efficient spatiotemporal message passing with $\mathcal{O}(P^2N + N^2P)$ complexity. To capture real-world traffic dynamics, we address the importance of negative edges in modeling traffic behavior (e.g., shockwaves, spillback) by introducing \textbf{Valence Attention} using the \textbf{continuous Tanimoto coefficient (CTC)}, which provides properties conducive to precise latent graph generation and training stability. To fully utilize the model's learning capacity, we introduce the \textbf{Traffic Phase Dictionary} for self-conditioning. This enables learning of universal traffic node representations on a compact, convex latent space, which augment the inputs through dynamic retrievals.

Weaver integrates principles from Graph Transformers, Interaction Networks, and Memory-Augmented Neural Networks into a modular architecture that separates spatial interactions, temporal evolution, and state representation, mirroring classical dynamical systems. It learns traffic dynamics without relying on exogenous metadata (e.g., timestamps) yet remains flexible to incorporate them when available. Evaluations on PEMS-BAY and METR-LA show that Weaver achieves competitive performance across model categories while training more efficiently.

\end{abstract}

\begin{keywords}
Traffic Forecasting\sep
Spatiotemporal forecasting\sep
Kronecker product approximation\sep
Deep Learning\sep
Time Expanded Network\sep
% Geometric Deep Learning\sep
\end{keywords}

\maketitle
\section{INTRODUCTION}

Urban transportation systems are central to daily mobility and economic activity, supporting diverse personal, commercial, and emergency travel. Effective planning and operation are essential for reducing congestion and delays, improving safety, minimizing environmental impact, and ensuring network reliability. Traffic speed is particularly pivotal, influencing travel time, fuel consumption, emissions, and Level-of-Service (LOS). Travel behavior reflects complex interplay between global factors such as time-of-day, weather, and seasonal cycles, and local conditions tied to infrastructure design, land use, and service disruptions \citep{LocalGlobal:pulugurtha2020exploring, Global:kumar1995temporal, Local:zhang2017impact}.

Recognizing the importance of short-term traffic forecasting, early systems like UTCS-2 incorporated predictive components for signal control \citep{Application:Head1995EventBased}. Recent deployments demonstrate operational viability in active traffic management. A model predictive controller in Chattanooga, Tennessee achieved over 30\% delay reduction and 5\% energy savings using real-time probe and sensor data \citep{Application:Wang2022MPCDeployment}. Forecasting models now support real-time incident detection by filtering participatory sensing alerts through predicted speed patterns \citep{Application:Hossain2025IncidentDetection} and proactive queue management in ramp metering \citep{Application:Taylor2000FuzzyRampMetering}. These applications reflect a broader shift toward embedding predictive analytics within traffic management systems (TMS) for dynamic decision-making \citep{Application:FHWA2024PredictiveAnalytics}.

Despite growing operational adoption, challenges remain in designing systems that generalize across spatial regions and adapt to traffic flow stochasticity. Deep learning architectures such as CNNs, RNNs, LSTMs, and hybrid models have shown strong performance in capturing spatiotemporal dependencies, yet often require large data volumes, careful tuning, and exhibit sensitivity to missing or noisy inputs \citep{Application:Kashyap2022Review}. Recent approaches incorporating Graph Neural Networks (GNN) and Transformer-based architectures offer greater modeling flexibility and long-range dependency capture, but raise concerns around computational cost, interpretability, and real-world scalability \citep{Application:Carianni2025Overview}. These challenges motivate research into architectures balancing predictive accuracy, robustness, and operational usability for intelligent transportation systems.

In this work, we propose \textit{Weaver} (Section~\ref{section:methodology}), a novel application of Kronecker product approximations to the Transformer architecture that enables system-level forecasting on traffic networks. 
Specifically, we construct a \textbf{Kronecker product approximation (KPA)} of the full $PN \times PN$ spatiotemporal attention map $\Theta_{\stset}$ (for $P$ temporal and $N$ spatial nodes) on a quasi-static Time-expanded Network (TEN)~\citep{TimeExpandedNetwork:ford1958}. 
The resulting \textit{Kronecker attention map} yields a novel \textbf{Kronecker-TEN graph} ($\kten{}$, see Figure~\ref{fig:graph_representations}C) representation that enables our \textbf{Parallel-Kronecker Matrix-Vector product (P$^2$-KMV)} for global message passing. 
The P$^2$-KMV operation weaves together the latent spatiotemporal embedding ($\tensor{U}_{\stset}$) with spatial ($\Theta_{\spaceset}$) and temporal ($\Theta_{\timesset}$) local attention maps, reducing the complexity of constructing $\attnst$ from $\Ocomplex(P^2N^2)$ to $\Ocomplex(P^2N + N^2P)$. 
We further derive an efficient formulation of P$^2$-KMV that leverages parallel computation patterns characteristic of modern GPU architectures, suggesting favorable scaling properties as hardware capabilities advance.

\begin{figure}[width=.99\linewidth,cols=4,pos=t]
    \centering
    \includegraphics[width=0.95\linewidth, trim=1.4cm 0.2cm 1.5cm 0.3cm, clip]{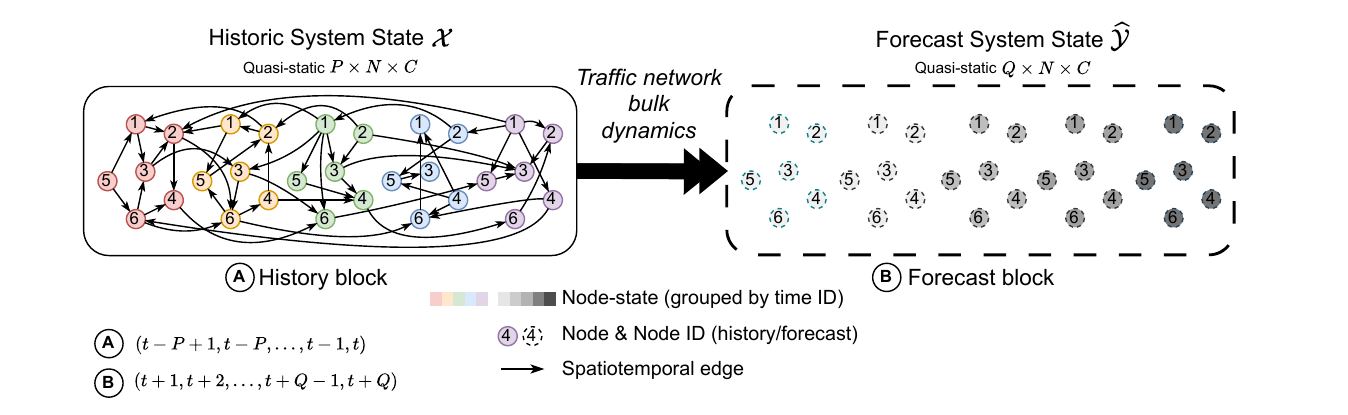} % [trim={left bottom right top},clip]
    \caption{Illustration of \textit{Weaver} performing spatiotemporal forecast as a bulk prediction problem in a quasi-static Time-expanded Network (TEN) setting. Weaver analyzes traffic patterns within the input system-state $\tensor{X} \in \mathbb{R}^{P \times N \times C}$ defined over the history block $(t-P+1, t-P+2, \ldots, t-1, t)$ using its \textbf{Kronecker Attention module} supported by the \textit{Traffic Phase Dictionary} and \textit{State Transition module} (see Figure~\ref{fig:weaver-superstructure}). These components learn and apply the traffic network's \textit{bulk dynamics} on the $PN \times PN$ spatiotemporal attention map to predict the block-level system trajectory, resulting in forecasted system-state $\widehat{\tensor{Y}}$ over the forecast block $(t+1, t+2, \ldots, t+Q-1, t+Q)$. }
    \label{fig:system-state-bulk-dynamics}
\end{figure}

We further demonstrate that applying typical head-mixing weights for multihead consolidation in Transformers results in a \textit{weighted implicit Kronecker product summation (W-iKPS)}, which relaxes the Kronecker product separability requirement of KPA.
Altogether, this enables \textit{Weaver} to operate concurrently over the entire \textit{quasi-static history window} (see Figure~\ref{fig:system-state-bulk-dynamics}), learning the traffic network's \textit{bulk dynamics} to predict the contiguous system-state in the succeeding \textit{forecast horizon} through the \textbf{state-transition module}.

Since Weaver systematically models higher-order relationships, local node relationships should be captured faithfully without superimposed dependencies. 
We introduce \textit{Valence Attention} through \textbf{Tanimoto Valence Attention (T-Valence)} in Section~\ref{section:problem2-signed-graphs}, addressing the misalignment between softmax-based attention and traffic node relationships. 
To fully utilize Kronecker attention capacity, we implement a \textbf{Traffic Phase Dictionary} that enables Weaver to self-condition by learning sparse latent parameter sets representing traffic phases. 
This shifts Weaver to a state-based regime, learning underlying network dynamics without exogenous cues from dynamic metadata such as timestamps.
 
Stated generally, \textit{Weaver} has \textit{three} major components (see Table~\ref{table:weaver-components-index} for corresponding modules):
\begin{alignat}{2}
    \tensor{U} &= \mathcal{F}_{\tensor{X}\rightarrow \tensor{U}}(\tensor{X}) &&\,:\, \mathbb{R}^{P \times N \times C} \rightarrow \mathbb{R}^{P \times N \times E} \,, \\
    \tensor{Z} &= \mathcal{F}_{\mathcal{G} \rightarrow \tensor{Z}}(\tensor{U}) + \tensor{U} &&\,:\, \mathbb{R}^{P \times N \times E} \rightarrow \mathbb{R}^{P \times N \times E} \,, \\
    \widehat{\tensor{Y}} &= \mathcal{F}_{\tensor{Z} \rightarrow \tensor{Y}}(\tensor{Z}) &&\,:\, \mathbb{R}^{P \times N \times E} \rightarrow \mathbb{R}^{Q \times N \times C} \,,
\end{alignat}
where 
(i) $\mathcal{F}_{\tensor{X}\rightarrow\tensor{U}}(\,\cdot\,)$, the point-of-entry, maps $\tensor{X}$ onto the feasible region of the model on its latent space $E$;
(ii) $\mathcal{F}_{\mathcal{G} \rightarrow \tensor{Z}}(\,\cdot\,)$ encodes relational information among nodes through graphical message passing, with attention modules as latent graph generators that are state-dependent representations of the traffic network;
and (iii) $\mathcal{F}_{\tensor{Z} \rightarrow \tensor{Y}}$ is the state-transition function that predicts the system trajectory by applying the learned bulk dynamics over the traffic network. Here, $P$ is the history window and $Q$ the forecast horizon. Important notations used in this paper is summarized in Appendix~\ref{appx:notations}.

\begin{table}[width=.99\linewidth,cols=4,pos=b]
\centering
\caption{Summary of core functions in \textit{Weaver} and their corresponding components, including the sections containing module blueprints (for specifics, refer to Section~\ref{section:problem-proposal-main}).  }
\begin{tabular}{l|l|l}
\toprule
\textbf{Function} & \textbf{Weaver Component} & \textbf{Section} \\
\midrule
$\mathcal{F}_{\tensor{X}\rightarrow\tensor{U}}$ 
  & Input module and Traffic Phase Dictionary (feasible region \& data augmentation)
  & Sec.~\ref{section:WeaverModule-input/augment} \\
$\mathcal{F}_{\mathcal{G} \rightarrow \tensor{Z}}$ 
  & Kronecker Attention and T-Valence (state-dependent signed latent graphs) 
  & Sec.~\ref{section:WeaverModule-kronecker-attention} \\
$\mathcal{F}_{\tensor{Z} \rightarrow \tensor{Y}}$ 
  & State-transition module (bulk dynamics and prediction) 
  & Sec.~\ref{section:WeaverModule-state-transition} \\
\bottomrule
\end{tabular} \label{table:weaver-components-index}
\end{table}

\section{LITERATURE REVIEW}

\subsection{Short-Term Traffic Forecasting: Exploring the `What, Why and How?'}

Short-term traffic forecasting aims to predict future traffic states including flow, speed, and occupancy using recent historical observations from roadway sensor networks. Given a multivariate time series $\{\mathbf{x}_{t}\}_{t=1}^{T}$ from $N$ sensors, where each $\mathbf{x}_{t} \in \mathbb{R}^{N}$ contains simultaneous observations at time $t$, the forecasting task uses a historical window of length $P$, $\{x_{t-P+1}, \, \ldots,x_t\}$ to estimate future values $\{\mathbf{x}_{t+1}, \dots, \mathbf{x}_{t+Q}\}$ over prediction horizon $Q$. For Intelligent Transportation System (ITS) applications, the lead time $Q$ is typically 60 minutes \citep{GraphWavenet:wu2019,KP:choi2023scalable,ASTGCN:guo2019astgcn,STTN:xu2021spatial}, as predictive reliability of short-history models degrades notably beyond this horizon \citep{wang2023anti}. Advances in mathematical modeling, data availability, and computational technology have collectively driven corresponding advancements in traffic forecasting techniques, some of which are detailed below.

\subsubsection{Classical foundations: Statistical methods and regression}

Early approaches often employ statistical and machine learning methods to capture localized temporal patterns in traffic flow. Kalman filtering enabled dynamic real-time traffic volume prediction \citep{HistKalman:okutani_stephan_1984}, while nonparametric regression using k-nearest neighbors demonstrated competitive performance against simple time-series predictors \citep{HistkNN:davis_nihan_1991}. Seasonal ARIMA models incorporated periodic trends and seasonality \citep{HistARIMA:akhtar2021traffic,HistARIMA:kumar2015short}, and moving average (MA) variants have provided lightweight yet competitive baselines for short-term traffic flow prediction \citep{HistMA:chowdhury2009development,HistMA:lv2015plane}. Linear regression leveraging local time-of-day windows provided interpretable short-term baselines \citep{HistLinear:sun2003use}, while Markov chains modeled transitions between discrete traffic states \citep{HistMarkov:yu2003short} with applications to real-time forecasting~\citep{HistMarkov:davis1995real}. Despite their interpretability, these classical models were constrained by linearity and short-memory assumptions, limiting their ability to capture nonlinear and network-dependent traffic dynamics.

\subsubsection{The era of deep learning: `And then... There was big data and computational power'}

Deep learning adoption accelerated with increased computational resources and data availability. Long short-term memory (LSTM) networks effectively modeled long-range temporal correlations, outperforming statistical baselines \citep{HistDL:zhao_et_al_2017_lstm}. Hybrid architectures emerged, including CLTFP \citep{HistDL:wu_tan_2016_cltfp}, which combined one-dimensional convolutions for spatial features with LSTMs for temporal modeling. Graph-based spatiotemporal models marked a significant advance, beginning with diffusion convolutional recurrent neural networks (DCRNN) \citep{DirectedGCN:li2018dcrnn}, which modeled traffic as diffusion over directed graphs integrated with gated recurrent units. The temporal graph convolutional network (T-GCN) \citep{HistDL:zhao_et_al_2019_tgcn} coupled graph convolution with GRUs for joint spatial-temporal modeling. STGCN \citep{HistDL:yu_et_al_2017_stgcn} replaced recurrent layers with temporal gated CNNs for improved efficiency, while Graph WaveNet \citep{GraphWavenet:wu2019} introduced adaptive adjacency learning and dilated causal convolutions to capture dynamic spatial structures without recurrence. These developments shifted from purely temporal models toward architectures capable of modeling complex spatiotemporal dynamics in large-scale networks.

\subsubsection{Transformers and multihead attention: `Everything, everywhere, all at once'}
More recently, Transformer architectures have been widely adopted for traffic forecasting, driven by their capacity to model long-range temporal dependencies and global spatial correlations through \textit{multihead dot-product attention}---a mechanism which exploits tensor operations and GPU parallelization to process multiple data projections concurrently. Transformer-based ST models, in contrast to traditional RNN or CNN approaches, handle sequences concurrently and assign dynamic weights to inputs across temporal and spatial dimensions, often in both directions, proving particularly advantageous in complex road networks. 

Representative examples include the Spatio-Temporal Transformer Network (STTN) \citep{STTN:xu2021spatial}, which applies separate spatial and temporal attention modules; the Spatial-Temporal Graph Sandwich Transformer (STGST) \citep{LitReviewTrf:Fan2023STGST}, which sandwiches spatial attention between temporal modules; GMAN \citep{LitReviewTrf:Zheng2020GMAN}, an encoder--decoder architecture with spatio-temporal attention blocks; and STTLM, which integrates a pre-trained language model to boost feature extraction \citep{LitReviewTrf:Ma2024STTLM}. Collectively, such works underscore the growing prominence of Transformer-based approaches in short-term traffic forecasting and spatiotemporal tasks, offering flexible and powerful alternatives to earlier architectures.

\subsection{Graph Representations in Deep Learning for Spatiotemporal Traffic Modeling}

Traffic networks exhibit complex spatiotemporal inter-dependencies---such as shared travel patterns, congestion propagation, and recurrent flow cycles---that cannot be captured by physical connectivity alone. Unlike conventional time-series forecasting, spatiotemporal modeling must consider interactions among spatially distinct nodes (e.g., traffic detectors or road segments) whose relationships are dictated by traffic dynamics rather than physical proximity.

For example, a highway overpass without connecting ramps may be completely independent from the highway below, despite their geographical overlap. Similarly, empirical studies show that traffic shockwaves propagate upstream across multiple roadway segments, affecting locations far from the original disturbance~\citep{shockwaveDistance:elfar2018traffic, shockwaveDistance:yang2024data}. Such cases reveal the inherently non-Euclidean structure of traffic networks and motivate the use of graph-based models that can represent and propagate information across latent relational structures.

\subsubsection{Graph representations of traffic networks}
Traffic networks can be represented as directed graphs $\mathcal{G} = (\Omega, \mathcal{E})$, where $\Omega = \{\omega_1, \dots, \omega_N\}$ is the set of nodes (e.g., sensors), and $\mathcal{E} \subseteq \Omega \times \Omega$ is the set of edges indicating pairwise connectivity or influence (e.g., road links or empirical correlations). Each node $\omega_r$ holds a feature vector $\mathbf{x}_r \in \mathbb{R}^C$ (e.g., speed, flow), and its neighborhood is defined as $\mathcal{N}(\omega_r) = \{\omega_s \mid (\omega_r, \omega_s) \in \mathcal{E}\}$, representing sender nodes that can provide information to the receiver node $\omega_r$ during \textbf{message passing (MP)}.

To model how information propagates over this structure, we adopt the \textbf{Message Passing Neural Network (MPNN)} framework~\citep{MPNN:gilmer2017neural}, which subsumes most convolutional graph neural networks and Transformer-based architectures. Classical approaches perform graph convolution over a predefined topology (Section~\ref{section:litreview_GNN-GCN}), whereas attention-based models infer edges dynamically through dot-product operators (Section~\ref{section:litreview_attention-MP}). The objective is to obtain expressive node representations admissible to traffic forecasting tasks.

\paragraph{Latent graph learning.}
Within geometric deep learning, \textit{latent graph learning} refers to discovering hidden relational structure directly from data rather than relying solely on a predefined physical topology. This is particularly relevant in traffic networks, where observed correlations often arise from phenomena not encoded in road geometry, such as shockwave propagation and spillback effects.

Early spatiotemporal models (e.g., Graph WaveNet~\citep{GraphWavenet:wu2019}, STTN~\citep{STTN:xu2021spatial}) introduced adaptive adjacency matrices learned from node embeddings, effectively adding new edges beyond the physical network. More recent architectures treat attention itself as the graph generator---e.g., Graphormer~\citep{TransformersAsGNNs:ying2021transformers}, Structure-Aware Transformer~\citep{TransformersAsGNNs:chen2022structure-main}, and GraphiT~\citep{TransformersAsGNNs:mialon2021graphit-main}---interpreting attention weights as dynamic edge affinities. This enables relational reasoning to emerge end-to-end during training, modeling data-driven node relationships without explicit graph design.

\subsubsection{Message passing (MP) neural networks} 
\label{section:litreview_GNN-GCN}

\textbf{Graph convolution (GC)}~\citep{GCN:kipf2017semi, DirectedGCN:zhang2019review} is a canonical message passing method that extends the idea of convolution to graphs. It operates over the \textbf{graph adjacency matrix (GAM)}: $\asmatrix{A} \in \mathbb{R}^{N_r \times N_s}$, which encodes pairwise connectivity among nodes ($A_{ij} = 0$ indicates no edge). Early spectral GC (SpecGC)~\citep{GCN:kipf2017semi} models were derived from spectral filtering on the graph Laplacian, requiring symmetrical GAMs and prior graph knowledge. Subsequently, spatial GC (SpatGC) models became dominant due to their simplicity, locality-based message passing, and strong generalization to unseen graph structures without enforcing GAM symmetry.

A simplified GC layer can be expressed as
\begin{align}
    \asmatrix{V}^{[\ell+1]}
    &=
    \underbrace{\midbar{\asmatrix{A}}\, \asmatrix{V}^{[\ell]}}_{\text{Aggregation}}
    \underbrace{\asmatrix{W}^{[\ell]}}_{\text{Update}}\, ,
    \label{eqn:graph-convolution-basic}
\end{align}
where $\asmatrix{V}^{[\ell]} \in \mathbb{R}^{N \times E}$ are node representations at layer $\ell=1,\ldots,L$, and
$\midbar{\asmatrix{A}} = \operatorname{GNorm}(\asmatrix{A} + \asmatrix{I})$ is the normalized adjacency with self-loops $\asmatrix{I} \in \mathbb{R}^{N \times N}$ to retain self-information. Nodes first \textit{aggregate} messages from neighbors, then \textit{update} their states using $\asmatrix{W}^{[\ell]} \in \mathbb{R}^{E \times E}$ (or an MLP, $\mathcal{F}(\cdot)$, in modern variants).

A key limitation of classical GC is that $\asmatrix{A}$ is \textit{fixed} and does not change with system state. Even when time-indexed (e.g., graph snapshots $\asmatrix{A}[t]$, see Figure~\ref{fig:graph_representations}A), the adjacency remains predefined and cannot adapt to dynamic traffic conditions. To overcome this rigidity, we employ attention-based graph generation, allowing node relationships to be inferred in latent space rather than imposed a priori.

\subsubsection{Attention as dynamic graph MPNN} 
\label{section:litreview_attention-MP}

The \textbf{scaled dot-product attention (SDPA)} mechanism, originally introduced in the Transformer architecture for sequential modeling~\citep{vaswani2017attention} in NLP, has since been recognized as a SpatGC operation that performs message passing over dynamically constructed graphs.
This development eventually led to \textit{Graph Transformers}~\citep{TransformersAsGNNs:joshi2025,TransformersAsGNNs:rampavsek2022recipe,TransformersAsGNNs:chen2022structure-main}, where attention defines edges dynamically, while structural encodings inject persistent graph priors.

\paragraph{Graph Transformers.}
For a graph with $N_{\receivingset}$ receiver nodes ($\receivingset=\{\omega_{r1}, \ldots, \omega_{rN_{\Omega_r}}\}$) and $N_{\sendingset}$ sender nodes ($\sendingset=\{\omega_{s1}, \ldots, \omega_{sN_{\Omega_s}}\}$), message passing in Graph Transformers follows equations~\eqref{eqn:multihead_attn}--\eqref{eqn:multihead_mlp}. When $\receivingset=\sendingset$, equation~\eqref{eqn:multihead_attn} performs \textit{self-attention}; when $\Omega_r \neq \Omega_s$, it performs \textit{cross-attention}. Equation~\eqref{eqn:multihead_attn} carries out \textit{message aggregation} using \textbf{multihead SDPA} to compute $H$ latent graph variants in parallel on GPUs, equations~\eqref{eqn:multihead_rearrange}--\eqref{eqn:multihead_mixhead} consolidate the heads, and equation~\eqref{eqn:multihead_mlp} updates receiver node representations using an MLP with LayerNorm~\citep{LayerNorm:ba2016layer}:
\begin{align}
    \tensor{U}^{[\ell]} &= \operatorname{Softmax}\left( \frac{\tensor{Q}^{[\ell]}(\tensor{K}^{[\ell]})^\top}{\sqrt{d_K}} \right) \tensor{V}^{[\ell]} = \attn^{[\ell]} \tensor{V}^{[\ell]} \,,
    \braceannotation[80mu]{0.0}{1.05}{\text{Aggregation}}
    \label{eqn:multihead_attn} \\
    \asmatrix{U}^{[\ell]} &= \LLangle \tensor{U}^{[\ell]}_{[ H \times N_{\receivingset} \times d_{\text{head}}\,\rightarrow\,N_{\receivingset} \times HD_{\text{head}}]} \,\RRangle ,
    \braceannotation[160mu]{-0.4}{1.25}{\text{Head-consolidation and self-loop}}
    \label{eqn:multihead_rearrange} \\
    \asmatrix{Z}^{[\ell]} &= \asmatrix{U}^{[\ell]} \asmatrix{W}^{[\ell]}_O + \asmatrix{H}_{\receivingset}^{[\ell]} ,
    \label{eqn:multihead_mixhead} \\
    \asmatrix{H}_{\receivingset}^{[\ell+1]} &= \sigma(\operatorname{LayerNorm}(\asmatrix{Z}^{[\ell]}) \asmatrix{W}_{\text{up}}^{[\ell]} + \asvector{b}_{\text{up}}^{[\ell]} ) \asmatrix{W}_{\text{dn}}^{[\ell]} + \asvector{b}_{\text{dn}}^{[\ell]} + \asmatrix{Z}^{[\ell]} \,.
    \braceannotation[22mu]{0.0}{1.05}{\text{Update}}
    \label{eqn:multihead_mlp}
\end{align}
Hereafter, we drop $[\ell]$ for legibility. The tensors $\tensor{Q}, \tensor{V} \in \mathbb{R}^{H \times N_{\receivingset} \times d_{\text{head}}}$ and $\tensor{K} \in \mathbb{R}^{H \times N_{\sendingset} \times d_{\text{head}}}$ are the multihead Query, Value, and Key tensors, where $d_K = d_{\text{head}}$ and $E = H d_{\text{head}}$. The head-mixing weight is $\asmatrix{W}_O \in \mathbb{R}^{E \times E}$, while the MLP weights and biases are $\asmatrix{W}_{\text{up}}\in \mathbb{R}^{E\times M_{\text{up}}E}, \asvector{b}_{\text{up}}\in\mathbb{R}^{M_{\text{up}}E}, \asmatrix{W}_{\text{dn}}\in \mathbb{R}^{M_{\text{up}}E\times E}, \asvector{b}_{\text{dn}}\in\mathbb{R}^{E}$, where $M_{\text{up}} \in \{2,3,4\}$. The terms $\asmatrix{H}_{\receivingset}$ and $\asmatrix{Z}$ in equations~\eqref{eqn:multihead_mixhead} and~\eqref{eqn:multihead_mlp} correspond to residual connections~\citep{ResNet:he2016residual}, analogous to self-loops in GC.

SDPA applies $\operatorname{Softmax}(\cdot)$ as the row-normalization scheme for $\operatorname{GNorm}(\cdot)$: 
\begin{align}
    \attn_{hij}=\operatorname{Softmax}(\midbar{\attn})_{hij} &= 
    \frac{\exp(\midbar{\attn}_{hij})}{\sum_{j'=1}^{N_{\sendingset}} \exp(\midbar{\attn}_{hij'})}\,, \label{eqn:attention-softmax-definition}
\end{align}
where $\midbar{\attn}_{hij}$ denotes the raw attention logits indexed by $h,i,j$ corresponding to dimensions: $H,N_{\Omega_r},N_{\Omega_s}$.

The computations for \textbf{multihead Query, Key and Values} are as follows:
\begin{align}
    \asmatrix{Q} &= \mathring{\mathcal{F}}_Q (\asmatrix{H}_{\receivingset} \asmatrix{W}_Q , \mathring{\bm{B}}_{\receivingset} ) ,
    \quad 
    \asmatrix{K}  = \mathring{\mathcal{F}}_K (\asmatrix{H}_{\sendingset} \asmatrix{W}_K , \mathring{\bm{B}}_{\sendingset} ), 
    \quad 
    \asmatrix{V}  = \mathring{\mathcal{F}}_V (\asmatrix{H}_{\receivingset} \asmatrix{W}_V , \mathring{\bm{B}}_{\receivingset} ),
\end{align}
followed by multihead stratification:
\begin{align}
    \tensor{Q} &= \LLangle \asmatrix{Q}_{[N_{\receivingset} \times HD_{\text{head}} \,\rightarrow\, H \times N_{\receivingset} \times d_{\text{head}}]} \RRangle, \nonumber \\
    \tensor{K} &= \LLangle \asmatrix{K}_{[N_{\sendingset} \times HD_{\text{head}} \,\rightarrow\, H \times N_{\sendingset} \times d_{\text{head}}]} \RRangle, 
    \braceannotation[22mu]{0.0}{1.80}{\text{Multihead stratification}} \\
    \tensor{V} &= \LLangle \asmatrix{V}_{[N_{\receivingset} \times HD_{\text{head}} \,\rightarrow\, H \times N_{\receivingset} \times d_{\text{head}}]} \RRangle \nonumber .
\end{align}
where $\asmatrix{Q},\asmatrix{V} \in \mathbb{R}^{N_{\receivingset} \times E}$, $\asmatrix{K} \in \mathbb{R}^{N_{\sendingset} \times E}$, and $\asmatrix{W}_{Q},\asmatrix{W}_{K},\asmatrix{W}_{V}\in\mathbb{R}^{E \times E}$.
Here, \textbf{structural encodings} $\mathring{\asmatrix{B}}_{\receivingset}\in \mathbb{R}^{N_{\receivingset} \times M_{B}}$ and $\mathring{\asmatrix{B}}_{\sendingset}\in \mathbb{R}^{N_{\sendingset} \times M_{B}}$ are learned or handcrafted graphical priors, whereas the \textbf{structural encoder function} $\mathring{\mathcal{F}}_{Q/K/V}(\cdot)$ (e.g., an MLP or linear function) converts the inputs to inductive biases for edge formations, enabling attention mechanisms to construct GAMs dynamically. $\mathring{\mathcal{F}}_{Q/K/V}(\cdot)$ can be a straightforward summation (with $M_B=E$) or include concatenations, e.g., $\mathring{\mathcal{F}}: \mathbb{R}^{N\times (E + M_B)} \rightarrow \mathbb{R}^{N\times E}$. Typically, $\mathring{\asmatrix{B}}_{\receivingset}, \mathring{\asmatrix{B}}_{\sendingset}$ and $\mathring{\mathcal{F}}_{Q/K/V}(\cdot)$ are applied once at $\ell=1$. Positional encodings are special cases of structural encodings that encode linear positions during sequence processing~\citep{vaswani2017attention, PositionalEmbeds:Dufter2022}, and are often pre-computed.

\subsection{Kronecker Product Approximations (KPA): Computational Complexity Reduction}\label{subsec:kpa}

\paragraph{Kronecker product approximation (KPA).} The Kronecker product~\citep{Kronecker:vanloan2000kronecker} has been widely applied to reduce computational complexity in machine learning. For $\asmatrix{A} \in \mathbb{R}^{I_{A_1} \times I_{A_2}}$ and $\asmatrix{B} \in \mathbb{R}^{I_{B_1} \times I_{B_2}}$, the KPA of $\asmatrix{C} \in \mathbb{R}^{I_{A_1} I_{B_1} \times I_{A_2} I_{B_2}}$ is:
\begin{align}
   \asmatrix{C} \approx \asmatrix{A} \otimes \asmatrix{B} =
   \begin{bmatrix}
       a_{11}\asmatrix{B} & a_{12}\asmatrix{B} & \cdots & a_{1I_{A_2}}\asmatrix{B} \\
       a_{21}\asmatrix{B} & a_{22}\asmatrix{B} & \cdots & a_{2I_{A_2}}\asmatrix{B} \\
       \vdots & \vdots & \ddots & \vdots \\
       a_{I_{A_1}1}\asmatrix{B} & a_{I_{A_1}2}\asmatrix{B} & \cdots & a_{I_{A_1}I_{A_2}}\asmatrix{B}
   \end{bmatrix}.
   \label{eqn:kronecker-decomp-basic}
\end{align}
This factorization enables substantial computational savings across diverse applications. In image processing, \citet{KronSum:moravitz2000svd} demonstrated that KPA reduces the cost of SVD-based restoration while preserving quality, \citet{KronSum:zhang2003mkps} extended this framework with multiple Kronecker products for compact hierarchical encodings, and \citet{KronSum:duarte2012kronecker} exploited separable operators for efficient multidimensional recovery through Kronecker compressive sensing. In deep learning, \citet{KronSum:jose2018kru} showed that Kronecker Recurrent Units reduce parameter count by three orders of magnitude while retaining long-horizon stability, and \citet{KronSum:benzing2019optimal} developed optimal Kronecker-sum approximations to Real-Time Recurrent Learning with unbiased gradient estimates. For spatiotemporal traffic modeling, \citet{KP:choi2023scalable} demonstrated that Kronecker decomposition of covariance matrices reduces complexity while capturing separable space-time correlations in probabilistic forecasting.

\paragraph{Kronecker product summation (KPS).} The KPS approximation: $\asmatrix{C}_R \approx \sum_{r=1}^{R} \asmatrix{A}_r \otimes \asmatrix{B}_r$ generalizes KPA by relaxing the strict separability requirement in target matrices. Empirical evidence consistently shows that \textbf{leading KPS ranks capture most structured variance relevant to model inference}. 

\citet{KronSum:greenewald2013kronecker} show that rank-1 Kronecker sum approximations capture 75--90\% of spatiotemporal covariance energy in motion-capture data, indicating most structure lies in the first term.
\citet{KronSum:genton2007separable} find that separable Kronecker models yield near-lossless space--time representations, with covariance errors below 2\% and prediction errors under 1\% in the Irish wind study.
\citet{KronSum:Luttinen2012} demonstrate that low-rank sums of separable Kronecker products enable Gaussian-process inference on million-point datasets with linear scaling and competitive accuracy.
\citet{KronSum:tsiligkaridis2013covariance} show that Kronecker PCA produces statistically efficient covariance estimates and better prediction, while \citet{KronSum:tsiligkaridis2014robust} extend it with sparse corrections that preserve structure under heavy-tailed noise and outperform standard estimators.

Altogether, these applications and observations motivate our use of KPA and KPS for modeling tractable spatiotemporal attention maps, enabling compact representations of large-scale traffic dynamics that retain expressiveness.

% Our model takes historical \textbf{spatiotemporal traffic network states} $\mathbf{X} \in \mathbb{R}^{P \times N \times C}$ (e.g., speed, volume, occupancy) over a history window of length $P$ and generates \textbf{forecasts} $\widehat{\mathbf{Y}} \in \mathbb{R}^{Q \times N \times C}$ over a forecast horizon of length $Q$ using message passing (MP) operations over a \textbf{quasi-static spatiotemporal attention map} $\attnst$ within a time-expanded network (TEN) graph.

\section{PROBLEM STATEMENTS and PROPOSALS} \label{section:problem-proposal-main}

\begin{figure}[width=.99\linewidth,cols=4,pos=t]
    \centering
    \includegraphics[width=0.80\linewidth,
        trim=85 15 30 15,  % left bottom right top (adjust as needed)
        clip
    ]{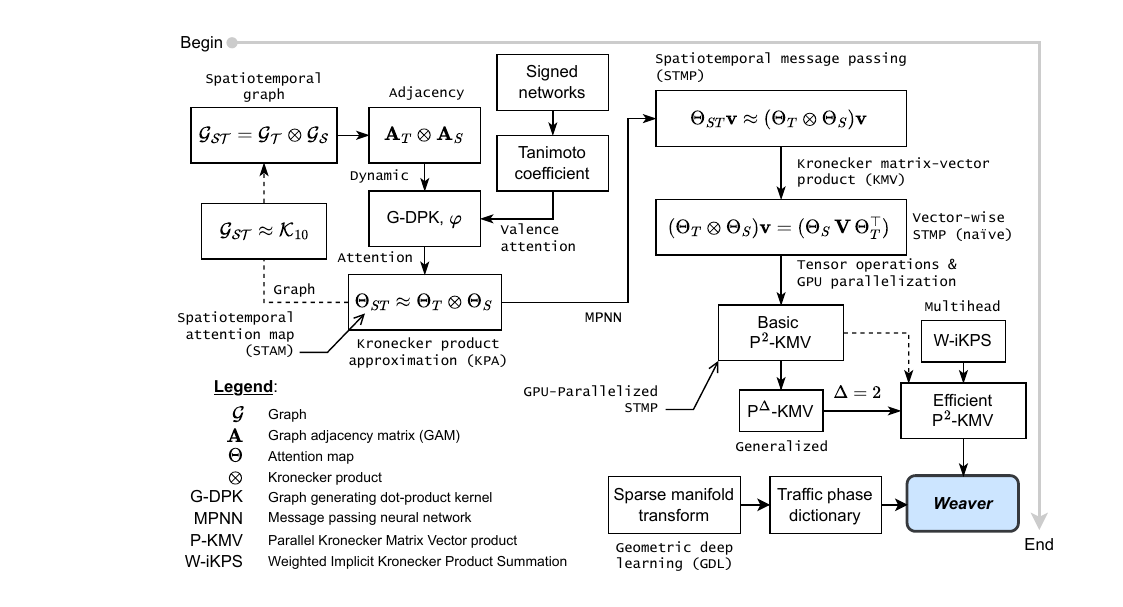}
    \caption{Flowchart of concept development in \textit{Weaver}. For STMP, the P$^2$-KMV is discussed in Section~\ref{section:problem1-kten} and P$^\Delta$-KMV in Appendix~\ref{appx:generalized-pkmv-main}; Signed networks, valence attention and Tanimoto coefficient in Section~\ref{section:problem2-signed-graphs}; for model self-conditioning, the sparse manifold transform and traffic phase dictionary in Section~\ref{section:problem3-self-conditioning}; for KPS, see W-iKPS in Section~\ref{section:kronecker-attention-module} and Appendix~\ref{appx:weighted-KPS}.}
    \label{fig:weaver-concept-map}
\end{figure}

\paragraph{General problem statement.} Given a traffic network with $N$ spatially distinct nodes representing traffic detectors over $P$ observed time steps, the goal is to predict $Q$ time steps ahead for all nodes:
\begin{align}
    \widehat{\tensor{Y}} &= \widehat{\mathcal{F}}(\tensor{X})\,,
\end{align}
where $\tensor{X}\in \mathbb{R}^{P \times N \times C}$ represents the observed traffic states with feature dimension $C$ (typically $C=1$), $\widehat{\tensor{Y}}\in \mathbb{R}^{Q\times N \times C}$ denotes the predictions, and $\widehat{\mathcal{F}}(\cdot)$ is the predictive model to be specified and learned. We propose modeling $\widehat{\mathcal{F}}(\cdot)$ as an MPNN which directly applies \textbf{spatiotemporal message passing (STMP)} over a fully-connected spatiotemporal graph, $\graph_{\mathcal{ST}}(\Omega_{\stset}, \attn_{\stset})$, where $\Omega_{\stset}, \vert \Omega_{\stset} \vert = PN$ is the spatiotemporal node-set and $\attn_{\stset}$ is the $PN \times PN$ \textbf{spatiotemporal attention map (STAM)} defining all feasible edges.

The primary contribution in this paper is approximating the STAM using \textbf{Kronecker product approximation (KPA)} (Section~\ref{subsec:kpa}): $\attnst \approx (\attntimes \otimes \attnspace)$. Consequently, STMP can be parformed directly on the spatiotemporal graph induced via the corresponding \textit{Kronecker graph product}: $\graph_{\mathcal{ST}} \approx \kten{} = \graph_{\timesset} \otimes \graph_{\spaceset}$ (Appendix~\ref{appx:Kronecker-graph-spatiotemporal}). To this end, we approach short‐term spatiotemporal traffic forecasting from a graph-theoretic and information-centric perspective, investigating computationally tractable spatiotemporal graph constructions and node representation learning.

\paragraph{Outline.} This section examines the graph topology characteristics of $\attnst$ under KPA and addresses three major challenges in applying this framework to traffic networks, with a high-level summary provided in Figure~\ref{fig:weaver-concept-map}.

\textbf{First}, Section~\ref{section:problem1-kten} analyzes the modeling mechanics under KPA and the resulting spatiotemporal message passing (STMP) over the TEN graph, revealing a computational bottleneck in the Kronecker matrix-vector product (KMV, Equations~\eqref{eqn:KMV_basic}). 

\textbf{Second}, Section~\ref{section:problem2-signed-graphs} identifies and addresses fundamental misalignment between traffic network dynamics---which exhibit both positive and negative feedback---and classical positive-only row-normalized attention maps, limiting the representational capacity of spatiotemporal attention. 

\textbf{Third}, Section~\ref{section:problem3-self-conditioning} observes that traffic datasets $\tensor{X} \in \mathbb{R}^{P \times N \times C}$ typically exhibit low feature dimensionality ($C=1$), creating a mismatch between the rich modeling capacity of TEN spatiotemporal attention and the limited input representation, necessitating strategies to enrich the input space through self-conditioning.

\subsection{Problem 1: The Kronecker-TEN ($\mathcal{K}_{10}$) Spatiotemporal Network} \label{section:problem1-kten}

\subsubsection{Spatiotemporal coherence of contemporary methods}

\begin{figure}[width=.99\linewidth,cols=4,pos=t]
    \centering
    \includegraphics[width=0.92\linewidth,
    trim=4cm 0.2cm 2.2cm 0.2cm, 
    clip]{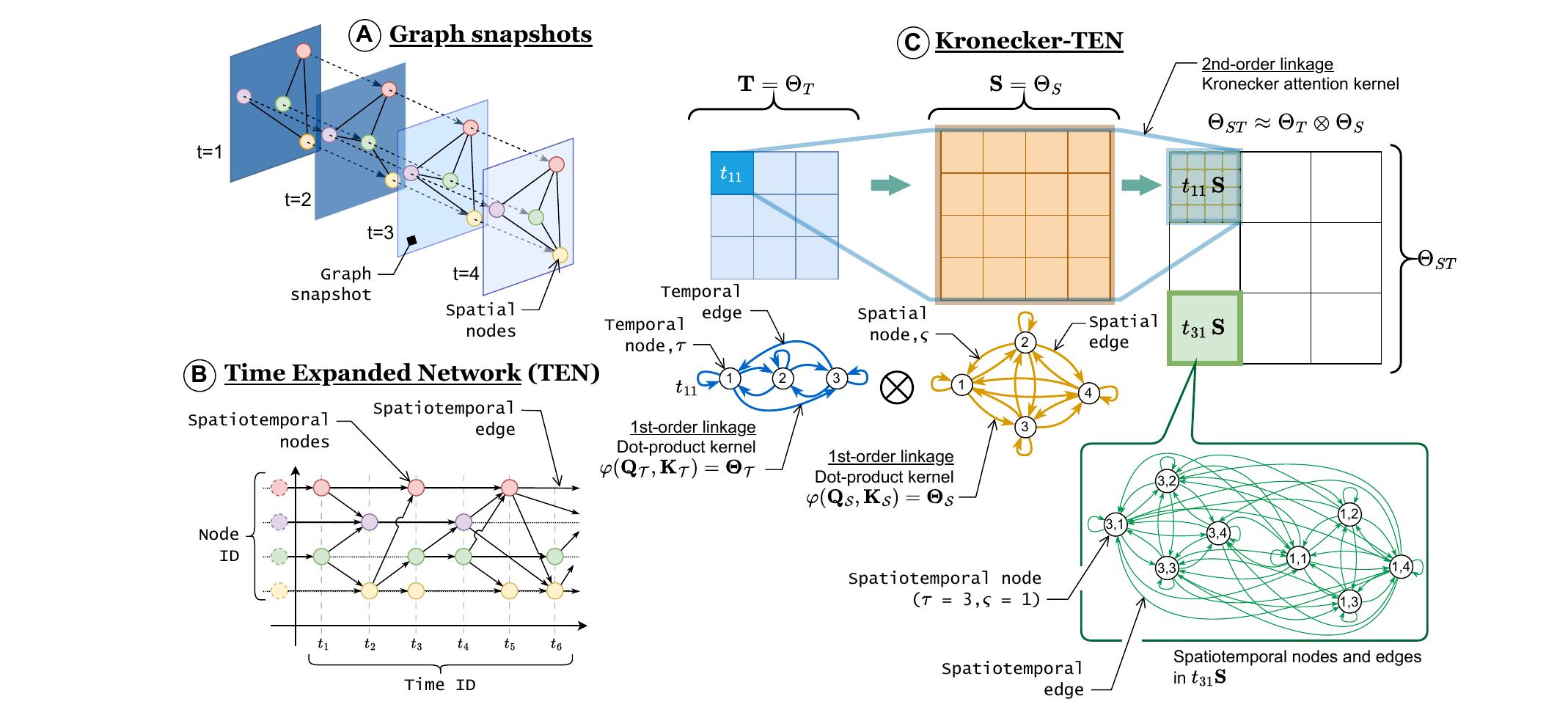} % [trim={left bottom right top},clip]
    \caption{Common spatiotemporal graph representations and our \textbf{Kronecker Time Expanded Network} (Kronecker-TEN, $\kten{}$). 
    (A) \textbf{Graph snapshots}: Each snapshot represents the network at a single time point with edges connecting nodes only within that time slice. 
    (B) \textbf{Time Expanded Networks (TEN)}: Nodes are identified by both location and time, with edges representing displacement across space and time. 
    (C) \textbf{Kronecker-TEN}: Applying KPA to the STAM: $\attn_{\mathcal{ST}} = \attntimes \otimes \attnspace$ induces a valid spatiotemporal graph via the Kronecker graph product: $\graph_{\mathcal{ST}} \approx \kten{} = \graph_{\mathcal{T}} \otimes \graph_{\mathcal{S}}$. First-order linkage (MP1) uses pairwise dot-product attention, while second-order linkage (MP2) applies the Kronecker attention kernel.
    }
    \label{fig:graph_representations}
\end{figure}

\paragraph{Current approaches.} One classical \textit{machine learning (ML)} approach in spatiotemporal traffic modeling represents the dynamic network graph as a sequence of \textit{graph snapshots} (Figure~\ref{fig:graph_representations}A), $\mathcal{G}_T=\{\mathcal{G}_1, \mathcal{G}_2, \ldots, \mathcal{G}_P\}$, where each $\mathcal{G}_t = (\mathcal{V}, \mathcal{E}_t)$ contains time-varying, prespecified edges $\mathcal{E}_t$ at time $t$. However, spatial and temporal dependencies are modeled on separate architectures, typically spatial graph convolutions on $\mathcal{G}_t$ and temporal dependencies on recurrent units or temporal convolutions. 

\begin{figure}[width=.99\linewidth,cols=4,pos=t]
    \centering
    \includegraphics[width=0.50\linewidth,
        trim=18 21 20 0,  % left bottom right top (adjust as needed)
        clip,
    ]{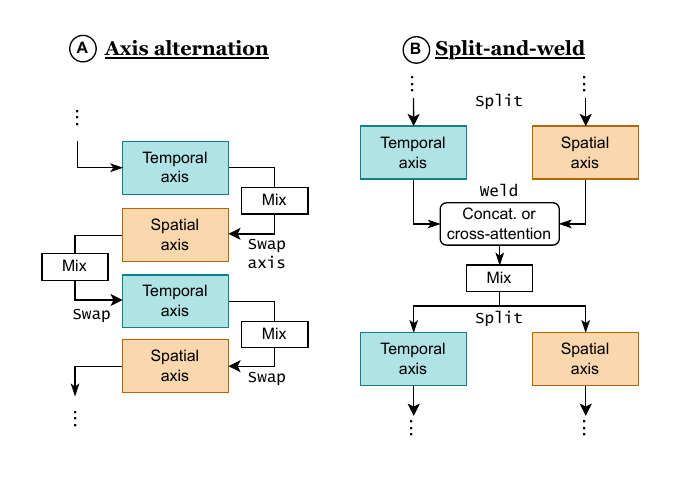}
    \caption{Conventional heuristic spatiotemporal processing in classical and contemporary models. (A) Axis alternation swaps processing between temporal and spatial axes, requiring multiple iterations to emulate spatiotemporal graph diffusion. (B) Split-and-weld performs separate spatial and temporal analysis, then welds them via concatenation or cross-attention prior to parametric mixing, similarly requiring multiple iterations to emulate spatiotemporal graph diffusion.}
    \label{fig:heuristic-spatiotemporal-fusion}
\end{figure}

To our knowledge, this separate treatment of spatial and temporal axes has remained pervasive in contemporary attentional deep learning methods, which typically employ heuristic fusion schemes (Figure~\ref{fig:heuristic-spatiotemporal-fusion}): 
\begin{itemize}
    \item \textbf{Axis alternation} (Figure~\ref{fig:heuristic-spatiotemporal-fusion}A):  
    Spatial ($N$) and temporal ($P$) dimensions are processed in alternating order—for example, applying spatial graph convolution then temporal recurrence (or the reverse).  
    This pattern appears in models such as~\citep{STAE:liu2023staeformer, STTN:xu2021spatial, AxisAlternation:chai2024spatiotemporal,AxisAlternation:bian2024traffic}.

    \item \textbf{Split-and-weld} (Figure~\ref{fig:heuristic-spatiotemporal-fusion}B):  
    Spatial and temporal representations are computed independently and later fused, e.g., by concatenation with MLP/gating or via cross-attention between spatial–temporal and temporal–spatial embeddings.  
    Models following this pattern include~\citep{SAW:li2024dynamic,SAW:lablack2023spatio,LitReviewTrf:Zheng2020GMAN}.
\end{itemize}
Consequently, this decoupling requires iterative processing to emulate complex spatiotemporal dependencies, introduces a bias toward local temporal coherence, and produces opaque graph-diffusion dynamics through heuristic fusion mechanisms, ultimately lacking true spatiotemporal cohesion since space and time are never modeled jointly.

% For \textbf{dynamic graphs}, a classical \textit{machine learning (ML)} approach in spatiotemporal traffic modeling (Figure~\ref{fig:graph_representations}A) represents the network as a sequence of \textit{graph snapshots} $\mathcal{G}_T=\{\mathcal{G}_1, \mathcal{G}_2, \ldots, \mathcal{G}_P\}$, where each $\mathcal{G}_t = (\mathcal{V}, \mathcal{E}_t)$ encodes the spatial graph structure with nodes $\mathcal{V} = \{v_s\}_{s \in [N]}$ denoting $N$ sensor locations and time-varying edges $\mathcal{E}_t$ at time $t$. The node features are represented separately as $\mathbf{X}_t \in \mathbb{R}^{N \times C}$, where $C$ denotes the number of traffic features, yielding a tensor $\mathbf{X} \in \mathbb{R}^{P \times N \times C}$ over a history window of length $P$. Spatial dependencies are modeled through graph convolutions on $\mathcal{G}_t$, while temporal dependencies are captured by recurrent units or temporal convolutions along the time axis. The forecasting task learns a mapping from historical observations $\mathbf{X} = (\mathbf{X}_{t-P+1}, \ldots, \mathbf{X}_t)$ and graph structure $\mathcal{G}_T$ to future predictions $\widehat{\mathbf{Y}} = (\widehat{\mathbf{Y}}_{t+1}, \ldots, \widehat{\mathbf{Y}}_{t+Q})$:
% \begin{align}
% \widehat{\mathbf{Y}} = \mathcal{F}(\mathbf{X}, \mathcal{G}_T)\,.
% \end{align}
% Because spatial and temporal dependencies are modeled separately, such approaches inherit a bias toward local temporal coherence. Time-expanded Networks resolve this by integrating time directly within the graph topology.

\paragraph{Spatiotemporal attention on Time-expanded networks (TEN).} Holistic spatiotemporal modeling can be achieved using TEN frameworks~\citep{TimeExpandedNetwork:ford1958}, which preserve spatiotemporal cohesion by creating nodes and edges indexed by both spatial (e.g., sensor IDs) and temporal (e.g., timestamps) identifiers (see Figure~\ref{fig:graph_representations}B). This produces \textbf{spatiotemporal nodes} ($\Omega_{\stset}$) \textbf{and edges} ($\mathcal{E}_{\stset}$) that evolve concurrently in space and time on the spatiotemporal graph $\graph_{\stset}:=\graph\,(\Omega_{\stset}, \mathcal{E}_{\stset})$. 

Previously, TENs have found numerous applications in network optimization problems, such as transit operations~\citep{TimeExpandedNetwork:kohler2002time, TimeExpandedNetwork:transitFare2022} and operational research~\citep{TimeExpandedNetwork:TDN2007, TimeExpandedNetwork:ho2014dynamic, TimeExpandedNetwork:liu2023multi}, enabling conventional static network optimization techniques by converting dynamic spatial graphs into \textit{quasi-static} spatiotemporal graphs. However, TENs have not seen widespread application in machine learning due to computational feasibility limitations. Specifically for dot-product attention, modeling attention over $PN$ spatiotemporal nodes, i.e., $\tensor{X}'\in \mathbb{R}^{PN\times C}$, incurs impractical $\Ocomplex(P^2N^2)$ complexity due to the $PN \times PN$ \textbf{spatiotemporal attention map (STAM)} $\Theta_{ST}\in \mathbb{R}^{PN \times PN}$ computation.

\subsubsection{Proposal: Kronecker-TEN representation of spatiotemporal Kronecker attention maps} \label{section:K10-main}

% \paragraph{Kronecker-TEN (Figure~\ref{fig:graph_representations}C).} Under the assumption that the TEN spatiotemporal attention map $\attnst \in \mathbb{R}^{PN \times PN}$ admits a KPA, a two-dimensional KMV characterizes spatial and temporal interactions on two levels: (i) local interactions (space-space, time-time), and (ii) global interactions (time-space). 
% The \textbf{Kronecker-TEN} reformulates the TEN graph as $\mathcal{K}_{10} = \{\mathcal{V}_{\kten{}}, \mathcal{E}_{\kten{}}\}$, where the node set $\mathcal{V}_{\kten{}} = (\spaceset \cup \timesset)$ is partitioned into spatial nodes $\spaceset = \{\spacenode_1, \ldots, \spacenode_N\}$ and temporal nodes $\timesset = \{\timesnode_1, \ldots, \timesnode_P\}$. 
% This bipartite structure requires \textit{two} orders of message passing: \textbf{(i) first-order MP1} (local: space-space, time-time) models pairwise node edges, and \textbf{(ii) second-order MP2} (global: time-space) models intermodal spatiotemporal edges.

\paragraph{Kronecker product approximation (KPA) of STAM.}  
To address the computational intractability of STAM, we propose applying KPA to $\attn_{ST}$, where $\otimes$ is the Kronecker product (Appendix~\ref{appx:Kronecker-product-and-operations}):
\begin{align}
    \attn_{ST} \approx \attn_{T} \otimes \attn_{S}. \label{eqn:kronecker-attention-kernel}
\end{align} 
This defines the \textbf{Kronecker attention kernel (Kronecker kernel)}, a non-parametric spatiotemporal fusion technique with systematic graph connectivity patterns---we call the resulting structure the \textbf{Kronecker TEN} (\textbf{K-TEN}, $\kten{}$), illustrated in Figure~\ref{fig:graph_representations}C. Appendix~\ref{appx:Kronecker-graph-spatiotemporal} shows Equation~\eqref{eqn:kronecker-attention-kernel} induces valid spatiotemporal graphs via the Kronecker graph product: $\graph_{\stset} \approx \kten{} = \graph_{\timesset} \otimes \graph_{\spaceset}$. Structurally, this reformulates the TEN graph as $\mathcal{K}_{10} = \{\Omega_{\kten{}}, \mathcal{E}_{\kten{}}\}$, with bipartite node-set $\Omega_{\kten{}} = (\Omega_{\spaceset} \cup \Omega_{\timesset})$ where $\Omega_{\spaceset} = \{\spacenode_1, \ldots, \spacenode_N\}$ and $\Omega_{\timesset} = \{\timesnode_1, \ldots, \timesnode_P\}$ are spatial and temporal node-sets.

\begin{figure}[width=.99\linewidth,cols=4,pos=t]
    \centering
    \includegraphics[width=0.94\linewidth,trim=1.0cm 0.2cm 0.77cm 0.2cm, clip]{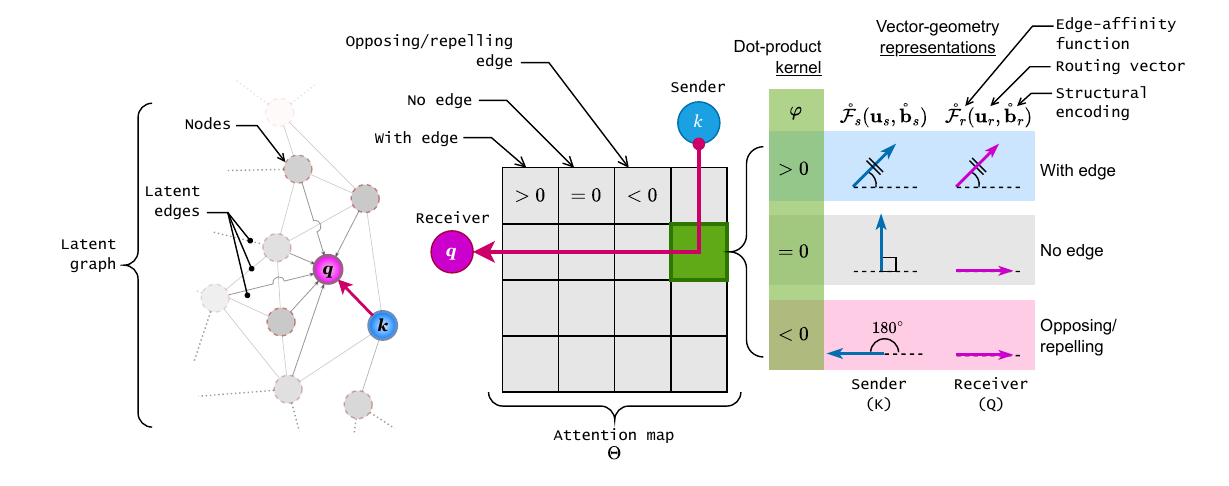} 
    % [trim={left bottom right top},clip]
    \caption{\textbf{Latent graph representations} induced by dot-product kernels $\varphi$ (e.g., dot-product attention, cosine similarity) between receiver (\textit{r}, query) and sender (\textit{s}, key) nodes.
    In typical attention, node routing vectors $\asvector{u}_r$ and $\asvector{u}_s$ (typically node features) are projected by routing weights $\asmatrix{W}_{q}$ and $\asmatrix{W}_{k}$, parameterizing a family of possible node-to-node relations.
    For each node-pair, the kernel $\varphi(\asvector{q},\asvector{k})$ instantiates a \emph{latent edge} by producing an edge affinity (strength and polarity); stacking all affinities over node pairs, i.e., $\varphi(\asmatrix{Q}, \asmatrix{K})$ yields attention maps $\attn$.
    The latent graph can be further shaped by augmenting the routing vectors, for example by appending structural encodings $\mathring{\asvector{B} \in \mathbb{R}^{N \times M_B}}$.
    } \label{fig:latent-graph-representation}
\end{figure}

The Kronecker kernel enables the \textbf{Kronecker matrix-vector (KMV)} product (Appendix~\ref{appx:Kronecker-matrix-vector-definition-KMV}) to perform STMP directly on the induced $\kten{}$ spatiotemporal graph. Its bipartite structure requires \textit{two} message passing orders: \textbf{(i) first-order MP1} (local: space-space, time-time) models pairwise node edges, and \textbf{(ii) second-order MP2} (global: time-space) models intermodal spatiotemporal edges.

\paragraph{First-order MP (MP1).} \textbf{MP1} constructs \textbf{latent graphs} using \textbf{graph-generating dot-product kernels (G-DPK)} $\varphi(\cdot)$ which generates adaptive edges between receiver (r) and sender (s) nodes dependent on each node's \textbf{routing vectors}: $\asvector{u}_r, \asvector{u}_s \in \mathbb{R}^{E}$ (Figure~\ref{fig:latent-graph-representation}). A similar framework was previously proposed in \citet{TransformersAsGNNs:chen2022structure-main}, but for softmax-normalized exponential kernels (e.g., Equation~\eqref{eqn:multihead_attn}). In other words, the G-DPK dynamically generates pairwise node edges by decoding vector-geometry, where $\varphi > 0$ indicates a positive edge, $\varphi=0$ no edge, and $\varphi < 0$ opposing or repelling edges~\citep{DynSignedNetworks:Shi2018DynamicsOS}. Common choices include the scaled dot-product (Equation~\eqref{eqn:multihead_attn}) and cosine coefficient (Equation~\eqref{eqn:sdp-cosine-definitions}). 

The G-DPK enables departure from classical softmax-based SDPA formulation (Equations~\eqref{eqn:multihead_attn}, \eqref{eqn:attention-softmax-definition}) and provides a unified vocabulary for graph theory and attention mechanisms. The G-DPK between a receiver $r$ and sender $s$ is:
\begin{align}
\dotkernel\big((\asvector{u}_r,\mathring{\asvector{b}}_r),(\asvector{u}_s,\mathring{\asvector{b}}_s);\mathring{\mathcal{F}}_r,\mathring{\mathcal{F}}_s\big)
    = \dotkernel\Big(
        \langle \mathring{\mathcal{F}}_r(\asvector{u}_r,\mathring{\asvector{b}}_r),
                \mathring{\mathcal{F}}_s(\asvector{u}_s,\mathring{\asvector{b}}_s) \rangle,
        \mathring{\mathcal{F}}_r(\asvector{u}_r,\mathring{\asvector{b}}_r),
        \mathring{\mathcal{F}}_s(\asvector{u}_s,\mathring{\asvector{b}}_s)
    \Big). \label{eqn:graph-generating-dot-G-DPK}
\end{align}
Here, $\asvector{u}_r,\asvector{u}_s \in \mathbb{R}^{1 \times E}$ are routing vectors;  
$\mathring{\asvector{b}}_r,\mathring{\asvector{b}}_s$ are structural encodings; and  
$\mathring{\mathcal{F}}_r(\cdot)$, $\mathring{\mathcal{F}}_s(\cdot)$ are \textbf{edge-affinity functions} (e.g., MLPs) that map node states to $E$-dimensional vectors, encoding latent edges via their lengths and angles. The G-DPK produces signed affinities, where $\varphi>0$ indicates an attractive edge, $\varphi=0$ no edge, and $\varphi<0$ a repelling edge~\citep{DynSignedNetworks:Shi2018DynamicsOS}.

For matrices of routing vectors, $\asmatrix{U}_r \in \mathbb{R}^{N_r \times E}$ and $\asmatrix{U}_s \in \mathbb{R}^{N_s \times E}$, Equation~\eqref{eqn:graph-generating-dot-G-DPK} yields $N_r \times N_s$ latent adjacencies (attention maps) with elements $\Theta_{rs}$:
\begin{align}
    \attn_{rs} = \dotkernel\big((\asmatrix{u}_r, \mathring{\asmatrix{B}}_r), (\asmatrix{u}_s, \mathring{\asmatrix{B}}_s); \mathring{\mathcal{F}}_r, \mathring{\mathcal{F}}_s  \big)\,.
\end{align}
In standard attention form, the Query and Key are linear projections of the routing vectors. Stated without structural encodings for notational clarity:
\begin{align}
    \asmatrix{Q} &= \mathring{\mathcal{F}}_r(\asmatrix{U}_r) = \asmatrix{U}_r  \asmatrix{W}_Q \,,\quad
    \asmatrix{K} = \mathring{\mathcal{F}}_s(\asmatrix{U}_s) = \asmatrix{U}_s  \asmatrix{W}_K \,.
\end{align}

\paragraph{Second-order MP (MP2).} Conversely, \textbf{MP2} is modeled by the Kronecker kernel (Equation~\eqref{eqn:kronecker-attention-kernel}), and applies STMP through KMV for $\mathbf{v}_e \in \mathbb{R}^{PN \times 1}$ through the $\kten{}$ graph, i.e., the \textbf{KMV-MP}:
\begin{align}
\mathbf{z}_e = \attnst \mathbf{v}_e^{\depth{0}} \approx (\Theta_{\timesset} \otimes \attnspace) \mathbf{v}_e^{\depth{0}} = \vecop(\attnspace \mathbf{V}_e^{\depth{1}} \Theta_{\timesset}^{\top})\,, \label{eqn:KMV_basic}
\end{align}
where $e \in [E]$ and $\mathbf{V}_e^{\depth{\delta}} = \vecop^{-1}(\mathbf{v}^{\depth{\delta-1}}_e) \in \mathbb{R}^{N \times P}$ denotes the vector folding (VF) of $\mathbf{v}_e$. Here, $\vecop(\attnspace \mathbf{V}_e^{\depth{\delta}} \Theta_{\timesset}^{\top})$ is the \textbf{\textit{KMV-vectorization}} operation, and $\depth{\delta} \in \{ \depth{0}, \depth{1}, \ldots, \depth{\Delta} \}$ the \textbf{\textit{KMV-vectorization depth}}.

\paragraph{The KMV-vectorization bottleneck.} To demonstrate the computational bottleneck in Equation~\eqref{eqn:KMV_basic}, we apply the \textbf{column-wise picture of matrix multiplication (CMM)} (Equation~\eqref{eqn:column-picture-matmul}, Appendix~\ref{appx:matrix-operations}) to compactly express Equation~\eqref{eqn:KMV_basic} for $e=1,\ldots, E$:
\begin{align}
   (\attntimes\otimes \attnspace)\,\asmatrix{V}^{\depth{0}} = 
   \begin{bmatrix} \, \vecop(\attnspace\, \asmatrix{V}^{\depth{1}}_1 \, \attntimes^\top) & \cdots & \vecop(\attnspace\, \asmatrix{V}^{\depth{1}}_{E} \, \attntimes^\top) \, \end{bmatrix}. 
   \label{eqn:column_KMV}
\end{align}
Here, the $\vecop(\cdot)$ is the computational bottleneck, since each $e=1,\ldots, E$ requires its own KMV-vectorization, which becomes costly when computed na\"ively using for-loops and concatenation. 

\paragraph{Basic P$^2$-KMV.} Using BMM (Equation~\eqref{eqn:definition-BMM-basic}, Appendix~\ref{appx:tensor-operations}) we may instead use an $E$-stratified tensor to parallelize Equation~\eqref{eqn:column_KMV} on GPU-accelerated frameworks. Put simply, the basic \textbf{P$^2$-KMV} (\textbf{Parallel 2D Kronecker Matrix-Vector} product) can be summarized in two equations:
\begin{align}
   \tensor{U}^{\depth{1}} &= \gtensor\Theta^!_{\spaceset} \tensor{V}^{\depth{1}} ( \gtensor\Theta_{\timesset}^!)^\top \in \mathbb{R}^{E \times N_1 \times P_1}, \label{eqn:P2KMV_basic1} \\ 
   \tensor{Z} 
   = (\timesattn\otimes \spaceattn)\,\tensor{V}^{\depth{0}} 
   &= \Langle \tensor{U}^{\depth{1}}_{[E \times N_1 \times P_1 \rightarrow P_1 N_1 \times E]} \Rangle \in \mathbb{R}^{P_1  N_1 \times E}, \label{eqn:P2KMV_basic2}
\end{align}
where $\tensor{V}^{\depth{1}} \in \mathbb{R}^{E \times P_2 \times N_2}$, with $\gtensor\Theta^!_{\spaceset} \in \mathbb{R}^{1_E^! \times N_1 \times N_2}$, and $\gtensor\Theta^!_{\timesset} \in \mathbb{R}^{1_E^! \times P_1 \times P_2}$ using tensor broadcasting (see Appendix~\ref{appx:notations}).

\paragraph{The (R)-P$^2$-KMV.} Efficiency on Equation~\eqref{eqn:P2KMV_basic1} can be further improved on GPU-accelerated frameworks using the general (R)-P$^{\Delta}$-KMV (Right-handed P$^{\Delta}$-KMV) form derived in Appendix~\ref{section:PKMV3} with $\Delta=2$, allowing us to exploit the invariance of $\attnspace$ and $\Theta_{\timesset}$ attention maps across $E$. This replaces tensor broadcasting with tensor rearrangements:
\begin{alignat}{2}
    \tensor{U}^{\depth{2}} &= \tensor{V}^{\depth{3}} (\attnspace)^\top &&\in \mathbb{R}^{E P_2 \times N_1}, \\
    \tensor{U}^{\depth{1}} &= \tensor{U}^{\depth{2}}_{\dotumble} (\Theta_{\timesset})^\top &&\in \mathbb{R}^{E N_1 \times P_1}, \\
    \tensor{Z} &= \tensor{U}^{\depth{1}}_{\dotumble} && \in \mathbb{R}^{P_1 N_1 \times E},
\end{alignat}
where $\tensor{V}^{\depth{3}}\in \mathbb{R}^{EP_2 \times N_2}$ and $\tensor{U}^{\depth{2}}_{\dotumble} \in \mathbb{R}^{E N_1 \times P_2}$. The $\dotumble$ (Kronecker-Tumble) tensor rearrangement comes from the general P$^{\Delta}$-KMV (see Appendix~\ref{section:PKMV3}). For brevity, we shall refer to this as P$^2$-KMV moving forward.

\begin{figure}[width=.99\linewidth,cols=4,pos=t]
    \centering
    \includegraphics[width=0.85\linewidth,
    trim=0.1cm 0.2cm 0.3cm 0cm, 
    clip
    ] % [trim={left bottom right top},clip]
    {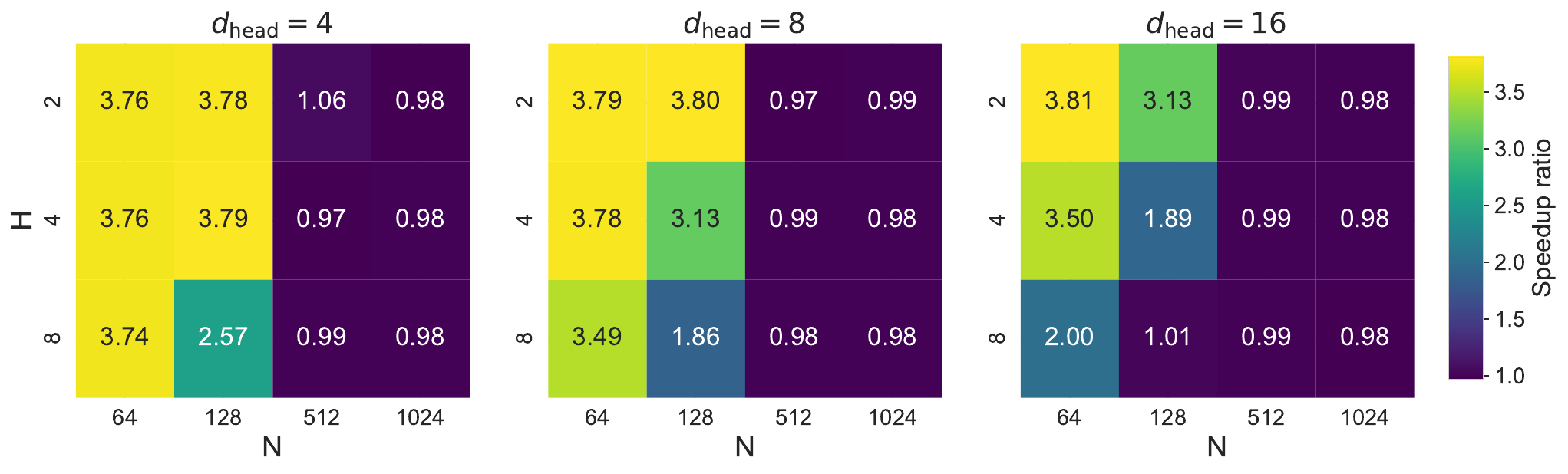}
    \caption{Relative computational time comparison of P$^2$-KMV on the NVidia A100 GPU using the Basic P$^2$-KMV (Basic) and (R)-P$^2$-KMV formulations for multihead tensors: $\tensor{U}\in \mathbb{R}^{B\times H\times d_{\text{head}} \times N \times P }, \attnspace \in \mathbb{R}^{B \times H \times N \times N}, \attntimes \in \mathbb{R}^{B \times H \times P \times P} $. For simplicity, $B=32$ and $P=12$. Heatmap values show the ratio of execution times: $\text{Speedup} = t_{\text{Basic}}/t_{\text{Efficient}}$. Values $>1.0$ indicate that the (R)-P$^2$-KMV is faster. Execution times $t_{\text{Basic}},\,t_{\text{Efficient}}$ are averaged over 10,000 trials. Numerical discrepancies are negligible, with absolute and relative differences of order $10^{-5}$ and $10^{-7}$, respectively.}
    \label{fig:pkmv_speedup_heatmaps}
\end{figure}

\paragraph{Multihead (R)-P$^2$-KMV.} The \textit{multihead} (R)-P$^2$-KMV is simply:
\begin{alignat}{2}
    \tensor{U}_H^{\depth{2}} &= \tensor{V}_H^{\depth{3}} (\Theta_{H,\spaceset})^\top &&\in \mathbb{R}^{H \times d_{\text{head}} P_2 \times N_1}, \label{eqn:PKMV-efficient1} \\ 
    \tensor{U}_H^{\depth{1}} &= \tensor{U}^{\depth{2}}_{H,\dotumble} (\Theta_{H,\timesset})^\top &&\in \mathbb{R}^{H \times d_{\text{head}} N_1 \times P_1}, \label{eqn:PKMV-efficient2} \\
    \tensor{Z}_H &= \tensor{U}^{\depth{1}}_{H,\dotumble} && \in \mathbb{R}^{H \times P_1 N_1 \times d_{\text{head}}}. \label{eqn:PKMV-efficient3}
\end{alignat}
Here, $\tensor{V}^{\depth{3}}_{H} \in \mathbb{R}^{H\times d_{\text{head}}P_2 \times N_2}$ and $\tensor{U}^{\depth{2}}_{H,\dotumble} \in \mathbb{R}^{H \times d_{\text{head}} N_1 \times P_2}$ with $E = H\cdot d_{\text{head}}$, where $H$ is the number of heads and $d_{\text{head}}$ is the head dimension. 

Computational speedup, expressed as the execution time ratio $t_{\text{Basic}} / t_{\text{Efficient}}$, is illustrated in Figure~\ref{fig:pkmv_speedup_heatmaps}, where $t_{\text{Basic}}$ denotes the runtime of the baseline implementation (Equations~\eqref{eqn:P2KMV_basic1}–\eqref{eqn:P2KMV_basic2}) and $t_{\text{Efficient}}$ corresponds to the optimized formulation (Equations~\eqref{eqn:PKMV-efficient1}–\eqref{eqn:PKMV-efficient3}). For moderately sized configurations (e.g., $N=64, H=2$), the proposed method achieves up to a 3.80× speedup. In larger-scale settings, the gain converges toward $\sim$1.0×, but likely due to memory bandwidth saturation on the A100 GPU~\citep{NVIDIA2023CUDA}, although the advantage is expected to scale favorably with future architectures offering higher memory throughput and improved tensor core performance; additional results comparing different GPUs are provided in Appendix~\ref{appx:P-KMV-extra-comparisons}.

\subsection{Problem 2: Traffic Network Dynamics, Why Signed Graph Representations Make Sense} \label{section:problem2-signed-graphs}
Since higher-order relationships in \textit{Weaver} are explicitly captured in $\attnst=\attntimes \otimes \attnspace$, the role of $\attnspace$ and $\attntimes$ is to portray node relationships accurately without redundant coupling. In this section, we discuss the relevance of signed graphs in traffic network representation, the downsides of traditional softmax attention, and propose Valence Attention to model signed edges using the continuous Tanimoto coefficient.

\subsubsection{The case for signed graphs}
\paragraph{Empirical evidence.} Empirical analyses in Minneapolis--St.\ Paul reveal that a substantial fraction of traffic edge-pairs exhibit negative correlations during congestion~\citep{negcorr:ermagun2017}, particularly among competing roads where increased flow on one coincides with decreased flow on another. Similarly, \citet{trafficstudy:ermagun2017using} demonstrate that, after controlling for temporal demand effects, correlations are not uniformly positive: although positive correlations dominate along consecutive links, negative correlations are also prevalent but generally weaker in magnitude under normal conditions, with stronger surges occurring across parallel or upstream segments during congestion. Real-world observations of Braess's Paradox~\citep{negcorr:youn2008} and simulation studies~\citep{negcorr:bazzan2005case} provide additional confirmation of this phenomenon.

\paragraph{Theoretical characteristics of softmax attention.} \label{section:network-dynamics_theoretical}
Softmax attention cannot efficiently capture systematic inverse relationships because the softmax function $p_i = e^{z_i}/\sum_j e^{z_j}$ produces strictly positive normalized weights ($p_i > 0$, $\sum_i p_i = 1$). This prevents modeling negative feedback from drivers' route adaptations and network equilibration. Beyond sign constraints, softmax attention (SDPA, Equation~\eqref{eqn:multihead_attn}) introduces spurious structural dependencies through its Jacobian:
\begin{align}
    \frac{\partial \alpha_{r,s}}{\partial z_{r,i}} 
    &= \alpha_{r,s} (\delta_{si} - \alpha_{r,i}), \quad 
    \delta_{ij} =
    \begin{cases}
        1 & i = j\quad, \\
        0 & i \neq j
    \end{cases}
    \label{eqn:softmax-attn-jacobian}
\end{align}
where $z_{r,i} = \mathbf{q}_r^\top \mathbf{k}_i$ is the attention logit and $\alpha_{r,s} = \frac{\exp(z_{r,s})}{\sum_j \exp(z_{r,j})}$ is the attention weight from query $r$ (receiver) to key $s$ (sender). Equation~\eqref{eqn:softmax-attn-jacobian} shows that increasing one key's score reduces all others' contributions, creating artificial dependencies \citep{AttnBehavior:correia2019,AttnBehavior:ye2021-sgat} despite traffic networks being sparse \citep{sparse:barthelemy2011spatial, sparse:tsiotas2025socioeconomic}. The normalization denominator $\sum_j \exp(z_{r,j})$ induces \textit{artificial higher-order coupling} \citep{AttnBehavior:koohpayegani2024} unrepresentative of genuine traffic interactions. This Jacobian is zero-sum\footnote{Zero-sum game: increasing one participant's reward necessarily decreases others' proportionally, resulting in zero net change.} \citep{AttnBehavior:yan2025,AttnBehavior:ramapuram2025}, contradicting traffic observations where multiple sensors simultaneously gain influence during widespread events like congestion or weather disruptions \citep{Global:kumar1995temporal,Global:liu2017weather,Global:loreti2024flood}. Although Sparsemax \citep{softmax:martins2016} and Entmax \citep{entmax:peters-etal-2019-sparse} address this, they still introduce unwanted higher-order dependencies through row-normalization.

\subsubsection{Proposal: Tanimoto Valence Attention}\label{sec:valattention}
We propose \textbf{Tanimoto Valence Attention (T-Valence)}, an attention mechanism that explicitly models polarity in signed networks. 
T-Valence builds on the \textbf{continuous Tanimoto coefficient (CTC)}~\citep{Tanimoto:Anastasiu2017,Tanimoto:Willett1998}:
\begin{align}
\varphi_{\text{ctc}}(\asvector{q},\asvector{k})
&= \frac{\asvector{q}^\top \asvector{k}}
{\Vert \asvector{q} \Vert^2 + \Vert \asvector{k} \Vert^2 - \asvector{q}^\top \asvector{k}}
\in \Big[-\tfrac{1}{3},\,1\Big], \label{eqn:definition-tanimoto-vector-qk}
\end{align}
whose asymmetric bound $[-\tfrac{1}{3},\,1]$ improves stability in signed networks. 

\paragraph{Signed network dynamics.} Prior work~\citep{DynSignedNetworks:Shi2018DynamicsOS} identifies \textit{two instability regimes} when negative edges encode \textbf{opposition}: 
(i) node-state collapse when the graph lacks \textit{structural balance}\footnote{
A signed graph is structurally balanced if nodes divide into two groups with positive intra-group and negative inter-group edges; equivalently, no cycle has an odd number of negative edges~\citep{DynSignedNetworks:Shi2018DynamicsOS}.
}, 
and (ii) divergence when update operators violate \textit{eventual positivity}, 
causing pathological growth of antagonistic influences over repeated updates 
(see also: Perron-Frobenius property~\citep{berman1994nonnegative,horn2012matrix}). 
The CTC's $3{:}1$ positive-to-negative range naturally regularizes message aggregation, preventing negative-edge dominance while permitting antagonistic relationships (sketch-of-proof in Appendix~\ref{appx:CTC-positive-dominance}). 
Its asymmetric saturation resembles activations like ELU~\citep{ELU:clevert2016,CELUs:klambauer2017snn,PELUs:trottier2017pelu}, potentially promoting self-normalizing gradients that improve training stability.

\begin{figure}[width=.99\linewidth,cols=4,pos=t]
    \centering
    \includegraphics[width=0.92\linewidth]{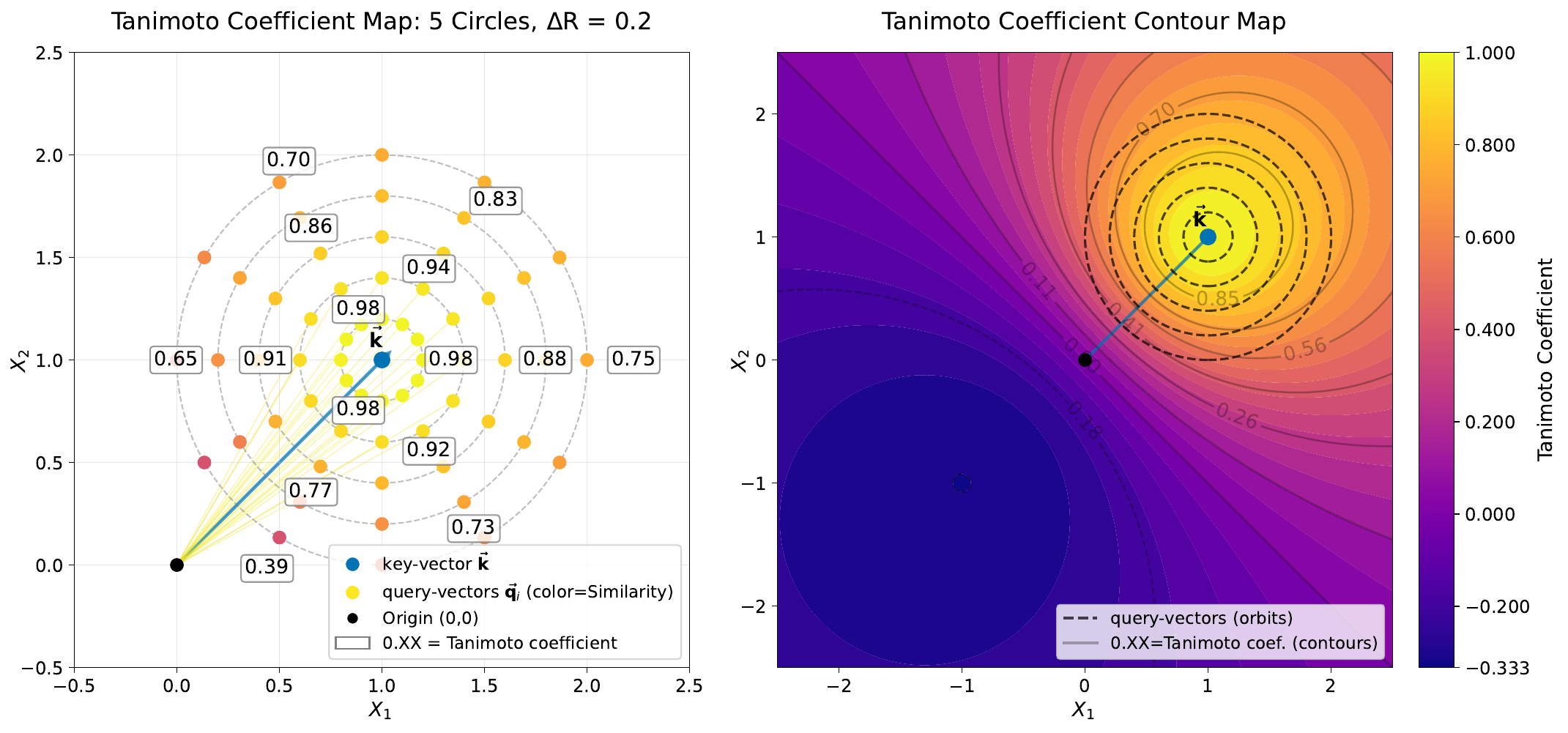}
    \caption{Visualization of 2D continuous Tanimoto coefficient (CTC) showing how CTC penalizes mismatches in both vector magnitude and angle through compact level sets. \textbf{Left:} Tanimoto coefficients (Equation~\eqref{eqn:definition-tanimoto-vector-qk}) computed between the 2D key vector $\vec{\asvector{k}} = ((0,0),(1,1))$ \textcolor{blue}{\textbf{(blue)}} and query vectors $\vec{\asvector{q}}_r = ((0,0), (x_1,x_2)_r)$ with endpoints $(x_1,x_2)$ placed on five concentric circles (radial spacing $\Delta R=0.2$). Point colors indicate similarity values (\textcolor{Goldenrod}{yellow} = high, \textcolor{Purple}{purple} = low). CTC values are labeled at staggered positions for readability. \textbf{Right:} Contour map of the continuous 2D Tanimoto topology with the query-vector orbits from the left panel superimposed. Unlike dot-product and cosine similarity, CTC produces compact, bounded level sets that provide numerical stability for signed network learning.}
    \label{fig:TaniContour}
\end{figure}

\begin{figure}[width=.99\linewidth,cols=4,pos=t]
    \centering
    \begin{subfigure}[b]{0.92\textwidth}
        \centering
        \includegraphics[width=0.92\linewidth]{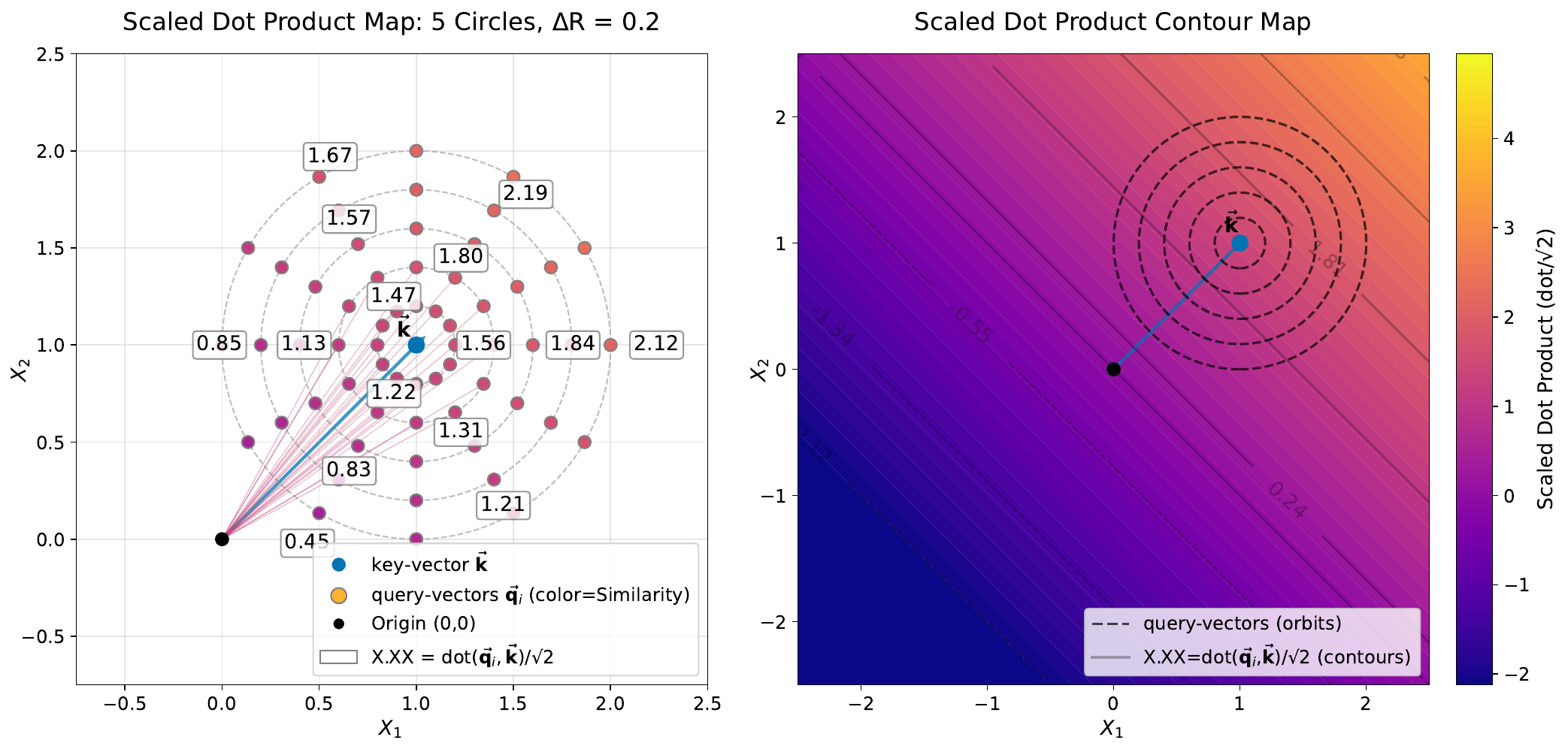}
        \caption{Scaled dot-product (SDP: $\asvector{q}^\top_r \asvector{k}/\sqrt{d_k}\,, \,d_k=2$) with non-compact level sets (unbounded longitudinal contours).}
        \label{fig:ScaledDotContour}
    \end{subfigure}
    
    \vspace{1em}
    
    \begin{subfigure}[b]{0.92\textwidth}
        \centering
        \includegraphics[width=0.92\linewidth]{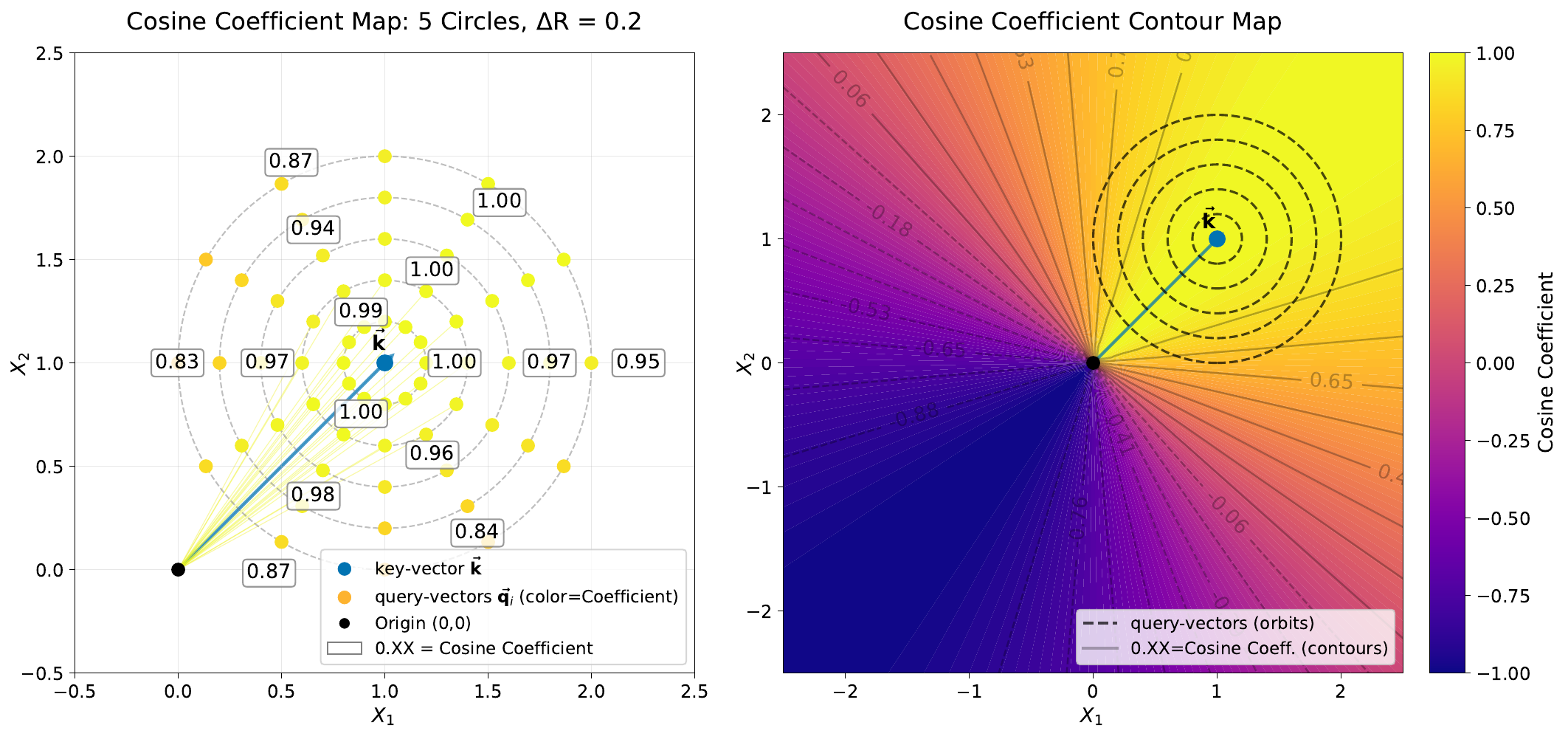}
        \caption{Cosine coefficient with radial non-compact level sets (unbounded radial contours).}
        \label{fig:CosineContour}
    \end{subfigure}
    \caption{Comparative visualization of vector-similarity topologies for (a) \textbf{scaled dot-product} (SDPA) and (b) \textbf{cosine coefficient}. Both metrics produce non-compact level sets, lacking the bounded structure of CTC (Figure~\ref{fig:TaniContour}) needed for stable signed network dynamics. Interpretation of plot elements in both figures match Figure~\ref{fig:TaniContour}.}
    \label{fig:Dot-Cosine-contours}
\end{figure}

\paragraph{Local characteristics.}
To understand CTC's local stability properties, we analyze the stationary points of $\varphi_{\text{ctc}}$ with respect to the query vector $\asvector{q}$ relative to the key vector $\asvector{k}$. Derivations are provided in Appendix~\ref{appx:tanimoto-critical-points}, and results are summarized in Table~\ref{tab:ctc_critical_points}.

\begin{table}[width=.99\linewidth,cols=4,pos=b]
\centering
\caption{Stationary points of the Continuous Tanimoto Coefficient (CTC) and their local stability.}
\begin{tabular}{l|c|c|cc}
\hline
\multirow{2}{*}{\textbf{Stationary Point}} & \multirow{2}{*}{\textbf{Type}} & \multirow{2}{*}{\textbf{CTC Value}} & \multicolumn{2}{c}{\textbf{Local Stability}} \\
\cline{4-5}
 & & & Gradient Ascent & Gradient Descent \\
\hline
$\asvector{q}_* = \asvector{k}$ & Local Maximum & $1$ & Stable & Unstable \\
$\asvector{q}_* = -\asvector{k}$ & Local Minimum & $-1/3$ & Unstable & Stable \\
\hline
\end{tabular}
\label{tab:ctc_critical_points}
\end{table}

Critical point analysis reveals that CTC admits two stationary states: a stable maximum at $\asvector{q}_*=\asvector{k}$ and a stable minimum at $\asvector{q}_*=-\asvector{k}$, depending on whether the objective is maximized or minimized. 
Under gradient ascent, the system converges toward full alignment ($\asvector{q} \to \asvector{k}$), while under gradient descent it converges toward full opposition ($\asvector{q} \to -\asvector{k}$). 
The origin $\asvector{q}=\boldsymbol{0}$ is not a stationary point—it lies on a sloped region where gradients align with $\asvector{k}$—thus avoiding collapse into uninformative null states. 
This complementary stability pattern ensures optimization moves decisively toward either alignment or opposition, rewarding clear relational polarity while penalizing ambiguity, which promotes well-defined signed graph structure in learned representations.

\paragraph{Qualitative comparisons to SDP and Cosine coefficient.}
Figure~\ref{fig:TaniContour} shows the CTC topology for 2D vectors, and Figure~\ref{fig:Dot-Cosine-contours} shows the corresponding topologies for SDP and Cosine coefficients. 
The left panels display orbital samples of coefficients computed on 2D query vectors $\asvector{q}_i$ around a fixed $\asvector{k}$, while the right panels show contour maps of all pairwise coefficient values, with orbital traces from the left panel superimposed. 

The \textbf{Scaled Dot-Product (SDP)} ($\varphi_{\text{sdp}}$) used in standard attention (without softmax) and the \textbf{Cosine coefficient} ($\varphi_{\text{cos}}$)---a potential alternative for Valence Attention with bounded range and support for negative similarity---are defined as:
\begin{align}
    \varphi_{\text{sdp}} &= \frac{ \asvector{q}^\top \asvector{k} }{ \sqrt{d_k} } \quad \in (-\infty, \infty), \quad
    \varphi_{\text{cos}} = \frac{ \asvector{q}^\top \asvector{k} }{ \Vert \asvector{q} \Vert \, \Vert \asvector{k} \Vert } \quad \in [-1, 1], \label{eqn:sdp-cosine-definitions}
\end{align}
where $d_k$ denotes the key dimension. CTC exhibits smooth, closed, and bipartite contours surrounding distinct maximum and minimum basins. 
In contrast, SDP shows open longitudinal contours without bounded extrema, while the Cosine coefficient forms radial contours with antipodal extrema but unbounded radial extent. 
The closed and bounded contour topology of CTC thus provides a well-defined optimization landscape with specific extrema, visually demonstrating the favorable training dynamics discussed above. 
Ablation studies in Section~\ref{section:problem2-signed-graphs} empirically confirm these properties.

\subsection{Problem 3: The Model–Data Capacity Gap (Self-Conditioning on Traffic Phases)} \label{section:problem3-self-conditioning}

Most traffic datasets provide only a single measurement channel ($C=1$), such as flow, speed, or occupancy, yielding $\tensor{X}\in\mathbb{R}^{P\times N\times1}$. This narrow feature space contrasts with the high representational capacity of $\textstyle{PN\times PN}$ attention mechanisms and richer multimodal inputs in domains like video or meteorology. Simple affine projections $\mathcal{F}(\tensor{X})$ merely reparameterize existing signals without introducing new information. To better utilize \textit{Weaver}'s modeling capacity, we introduce a \textit{self-conditioning} mechanism based on Memory Augmented Neural Networks (MANN)~\citep{NTM:graves2014ntm,NTM:DNC2016,MANN:ma2023memory}, where the model retrieves latent embeddings from a learned dictionary to condition downstream modules.

Classical traffic flow theory provides a natural foundation for such conditioning. The Greenshields and Cell Transmission models~\citep{greenshields1935study,daganzo1994cell} describe traffic dynamics through the \textit{fundamental diagram}—a relationship among flow, density, and speed that implicitly encodes distinct operating regimes (free flow, saturation, congestion) through parameterization. These regimes indicate that mesoscopic traffic behavior organizes around distinct \textit{phases} governing local operating conditions. Accordingly, self-conditioning in \textit{Weaver} retrieves phase-specific cofactors that explicitly capture these latent regimes, enabling the model to internalize shared, state-dependent structure underlying observed network dynamics.

\subsubsection{Proposal: Traffic Phase Dictionary (Retrieval-Based Self-Conditioning)}

Prior MANN-based Transformer models such as STTN~\citep{STTN:xu2021spatial} and STAEformer~\citep{STAE:liu2023staeformer} primarily learn network-specific embeddings rather than underlying dynamics. STTN encodes graph topology and temporal evolution directly within its embeddings, while STAEformer learns embeddings tied to fixed node positions and time indices. In both cases, node representations and graph structure become interdependent, making it unclear whether the learned embeddings reflect intrinsic node behavior or structural identity. This entanglement biases the model toward memorizing where and when measurements occur, rather than uncovering the underlying dynamics that generate those measurements.

\paragraph{Motivating work.}  
Our approach builds on the premise of \textit{Interaction Networks} (IN)~\citep{InteractionNet:battaglia2016interaction}, which learn universal object–object interaction functions that generalize across configurations. Since identical physical quantities are observed at every traffic node, they should obey a common set of fundamental principles independent of spatial or temporal context.

To design this dictionary, we require stable representations confined to bounded, continuous regions that avoid degeneracy while maintaining smooth, well-behaved mappings across downstream \textit{Weaver} modules. \citet{SparseManifoldTransform:chen2018,SparseManifoldTransform:chen2022minimalistic} solved a related problem through the \textit{Sparse Manifold Transform} (SMT), a geometric framework in which dictionary atoms act as discrete \textit{landmarks} approximating a smooth underlying manifold. Each sample is expressed as a sparse convex combination of nearby landmarks, enabling interpolation within the learned manifold approximation.

SMT enforces continuity, stability, and approximate invertibility by constraining activations to the convex hull of the manifold and by learning a linear, similarity-preserving embedding from local neighborhoods, originally temporal and later generalized to spatial and co-occurrence structures. These properties align with the goals of our dictionary design, promoting sparsity, preserving structural distinctions, and mitigating degeneracy among learned landmarks.

\paragraph{Traffic Phase Dictionary.}
While SMT provides desirable theoretical guarantees, its two-step procedure is not directly compatible with end-to-end deep network training. For practical implementation, we therefore approximate SMT characteristics using differentiable components that preserve similar local smoothness and convex-combination behavior.

\begin{figure}[width=.99\linewidth,cols=4,pos=t]
    \centering
    \includegraphics[width=0.64\linewidth,trim=0.0cm 0.4cm 0.7cm 0.2cm, clip]{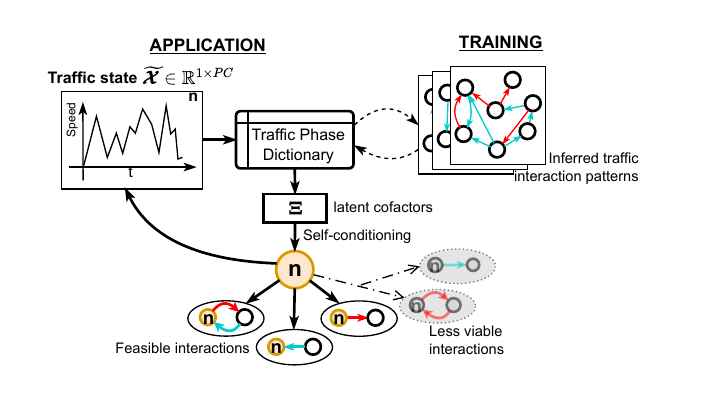}
    % [trim={left bottom right top},clip]
    \caption{Traffic Phase Dictionary retrieval. Nodes query a learned dictionary $\tensor{C}_{\Xi}$ with traffic state $\widetilde{\asmatrix{X}}$ to retrieve phase-specific cofactors $\boldsymbol{\Xi}$ that condition downstream interactions.}
    \label{fig:traffic_phase_dictionary}
\end{figure}

We define the \textbf{Traffic Phase Dictionary} as a shared sparse retrieval mechanism that conditions each spatiotemporal node in the $P\times N$ graph (corresponding to time steps $P$ and spatial nodes $N$) on latent cofactors $\boldsymbol{\Xi}$ representing the local traffic phase:
\begin{alignat}{3}
    S_{\Xi} (\,\midtilde{\asmatrix{X}} \,) &= \entmax_{1.5}\big(\, \mathcal{F}_{\Xi}\,(\widetilde{\asmatrix{X}})\,/\, (\boldsymbol{\tau}_{\Xi})_+\,\big) & \quad & \in \mathbb{R}^{N \times M_{\Xi}}  \,\, &&\text{(Sparse retrieval weights)}\,, \label{eqn:engram-sparse-map}  \\
    \midtilde{ \boldsymbol{\Xi} } &= \, S_{\Xi} (\,\midtilde{\tensor{X}} \,)  \,\, \tensor{C}_{\Xi} && \in \mathbb{R}^{N \times P K_{\Xi} } \,\, &&\text{(Latent parameter retrieval)}\,, \\
    \boldsymbol{\Xi} &= \Langle \, \midtilde{\tensor{E}}_{[N \times PK_{\Xi} \rightarrow P \times N \times K_{\Xi}]} \, \Rangle && \in \mathbb{R}^{P\times N \times K_{\Xi}} \,\, &&\text{(Latent cofactors)}\,,  
\end{alignat}
where $\widetilde{\asmatrix{X}} \in \mathbb{R}^{N \times PC}$ denotes the node traffic-state matrix, $\mathcal{F}_{\Xi}:\mathbb{R}^{N\times PC} \rightarrow \mathbb{R}^{N \times M_{\Xi}}$ is an MLP that produces retrieval queries, and $\tensor{C}_{\Xi} \in \mathbb{R}^{M_{\Xi} \times P K_{\Xi}}$ is is the learned dictionary of phase landmarks, representing a closed, continuous set of latent parameters $\midtilde{\boldsymbol{\Xi}}$ characterizing traffic phases. The resulting outputs, $\boldsymbol{\Xi}$, are the node's \textbf{latent cofactors}\footnote{In biochemistry \citep{cofactors:kara2014recent}, a ``cofactor'' is a helper molecule (non-protein compound or metallic ion) that assists enzymes in catalyzing reactions during biotransformations.}: phase-specific parameters retrieved from the dictionary that modulate each spatiotemporal node's participation in downstream processes such as latent edge formation and state transitions (Figure~\ref{fig:traffic_phase_dictionary}). 

\begin{figure}[width=.99\linewidth,cols=4,pos=t]
    \centering
    \includegraphics[width=0.90\linewidth,trim=0.0cm 0.1cm 0.0cm 0.0cm, clip]{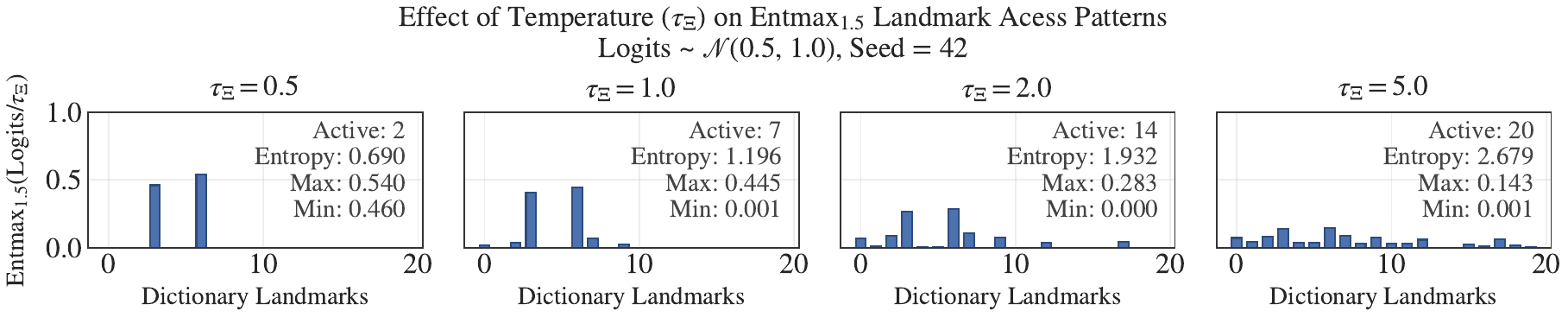} % [trim={left bottom right top},clip]
    \caption{Effect of temperature parameter $\tau_{\Xi}$ on sparsity within the Traffic Phase Dictionary manifold. Curves show retrieval weights for 20 logits sampled from $\mathcal{N}(0.5,1.0)$. Higher $\tau_{\Xi}$ produces broader activations blending multiple phase landmarks, whereas lower values yield sharper, more selective associations. Learned node-specific temperatures $\tau_{\Xi} \in \mathbb{R}^{N\times1}$ enable adaptive retrieval sharpness across traffic nodes, adjusting for local context while maintaining a shared dictionary manifold.}
    \label{fig:entmax_temperature_effects}
\end{figure}

To account for node-specific variability, learnable temperatures $\boldsymbol{\tau}_{\Xi}\in \mathbb{R}^{N\times1}$ modulate the dictionary access range (Figure~\ref{fig:entmax_temperature_effects}) through a positivity constraint $(\,\cdot\,)_+$, e.g., $\operatorname{Softplus}(\tau)=\log(1+e^\tau)$, preventing sign inversion. These lightweight parameters enable localized adaptation without over-specializing the shared dictionary $\tensor{C}_{\Xi}$ to specific locales. In practice, the learned temperatures act as \textit{node-local calibration factors} that implicitly adjust for operational conditions such as speed limits, ramp proximity, and driver behavior, serving as statistical proxies for locality context.

In Equation~\eqref{eqn:engram-sparse-map}, the SMT is realized through the sparse retrieval mechanism $S_{\Xi}(\,\widetilde{\asmatrix{X}}\,)\,\tensor{C}_{\Xi}$, which employs \textit{Entmax}~\citep{entmax:peters-etal-2019-sparse}. \textit{Entmax} interpolates between Softmax and Sparsemax~\citep{softmax:martins2016}, providing a differentiable means of enforcing convex-hull feasibility, continuity, and sparsity. The \textit{Entmax}$_{1.5}$ variant admits the closed form:
\begin{align}
    p_i = \operatorname{Entmax}_{1.5}(x)_i 
    = \operatorname{ReLU}\Bigl(\tfrac{x_i}{2} - \gamma \Bigr)^{2}, 
    \quad \text{s.t. } \sum_i p_i = 1.0,
\end{align}
where the threshold $\gamma$ is solved locally to ensure that the retrieval weights lie on the standard simplex. An exact $\mathcal{O}(d \log d)$ algorithm with near-Softmax runtime is provided in \citet{entmax:peters-etal-2019-sparse}. Unlike its typical probabilistic interpretation, \textit{Entmax} here serves a geometric role—producing sparse, simplex-constrained attention over dictionary landmarks that approximates SMT behavior within a fully differentiable framework.

\clearpage 
\section{METHODOLOGY} \label{section:methodology}
\begin{figure}[width=.99\linewidth,cols=4,pos=t]
    \centering
    \includegraphics[width=0.87\linewidth,trim=1.0cm 1.05cm 2.1cm 0.68cm, clip]{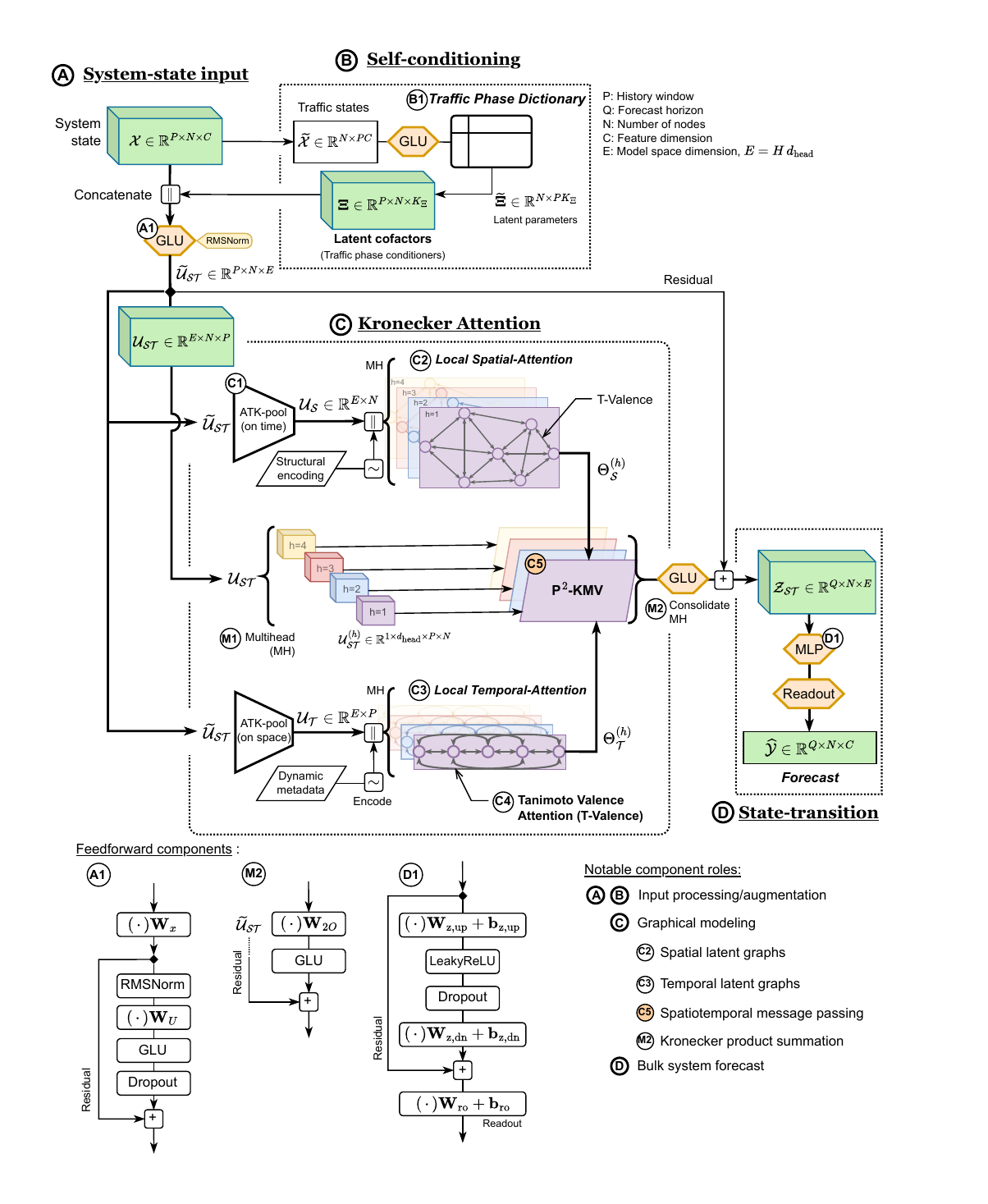} 
    % [trim={left bottom right top},clip]
    \caption{Schematic diagram of the \textbf{Weaver} pipeline: (A,B) The model self-conditions through the \textbf{Traffic Phase Dictionary} by mapping traffic states $\midtilde{\tensor{X}}$ to \textit{latent cofactors} ${\boldsymbol{\Xi}}$, then conditions the input space on traffic phases via concatenation: $(\tensor{X}\Vert\boldsymbol{\Xi})$. (C) The \textbf{Kronecker Attention module} generate local node representations with ATk-pooling, computes attention maps ($\Theta_\spaceset, \Theta_\timesset$), then performs \textit{P$^{\,2}$-KMV} for spatiotemporal message passing by \textit{weaving} multihead $\Theta_\spaceset^H, \Theta_\timesset^H$ with $\textstyle{\tensor{U}}_{\stset}^H$. (D) The \textbf{State-transition module} predicts $\widehat{\mathbf{Y}}$ via MLP and linear readout. Module configurations for (A1), (M2) and (D1) are along the bottom, where $(\,\cdot\,)$ are inputs from prior modules. Full details in Section~\ref{section:methodology} and~\ref{section:model-specification}.}
    \label{fig:weaver-superstructure}
\end{figure}

In this section, we provide a high-level overview of the \textit{Weaver} architecture (Figure~\ref{fig:weaver-superstructure}) and its three primary modules: (A/B) Input processing and self-conditioning, (C) Kronecker Attention, and (C) State-transition. For the specific implementation of Weaver in this work, refer to Section~\ref{section:model-specification}; neural network components, see Appendix~\ref{appx:neural-network-components}.

\subsection{Main Components} 
\subsubsection{Input and Conditioning (Figure~\ref{fig:weaver-superstructure}, A/B)}
The \textbf{Traffic Phase Dictionary} (Figure~\ref{fig:weaver-superstructure}, \textit{B2} (Equations~\eqref{eqn:dict-reshape-input}--\eqref{eqn:dict-reshape-output}), enables model self-conditioning by mapping node-wise traffic states $\midtilde{\tensor{X}}\in \mathbb{R}^{N \times PC}$ to latent parameters $\midtilde{\boldsymbol{\Xi}} \in \mathbb{R}^{N \times PK_{\Xi}}$ that characterize the traffic phase at each node. These parameters are rearranged to latent cofactors $\boldsymbol{\Xi} \in \mathbb{R}^{P \times N \times K_{\Xi}}$, which dictate the admissible spatiotemporal node feature transformations and latent edge formation in subsequent modules. 

The conditioned representation is formed by concatenating the raw input and cofactors $(\tensor{X} \Vert \boldsymbol{\Xi})$, then projecting into model space (Equations~\eqref{eqn:proj-concat}--\eqref{eqn:proj-reshape}) via gated transformations with residual connections (Figure~\ref{fig:weaver-superstructure}, \textit{A1}). This yields $\midtilde{\tensor{U}}_{\stset} \in \mathbb{R}^{E \times N \times P}$, which is rearranged to the P$^2$-KMV compatible form: $\tensor{U}_{\stset} \in \mathbb{R}^{E \times P \times N}$.

\subsubsection{Kronecker Attention module (Figure~\ref{fig:weaver-superstructure}, C)} \label{section:kronecker-attention-module}
The \textbf{Kronecker Attention module} handles graph learning through a four-stage process. %Implementation details are provided in Section~\ref{section:WeaverModule-kronecker-attention}.
\paragraph{Stage 1: ATk-pooling C1.} 
\textbf{ATk-pooling} (Equations~\eqref{eqn:atk-spatial-aggregate}--\eqref{eqn:atk-spatial-scores}) aggregates spatiotemporal features of $\midtilde{\tensor{U}}_{\stset}$ into local representations of spatial nodes $(\mathbf{U}_{\mathcal{S}}\in\mathbb{R}^{N \times E})$ and temporal nodes $(\mathbf{U}_{\mathcal{T}}\in\mathbb{R}^{P \times E})$ with adaptive top-$k$ selection. The selection uses a pooling informativeness metric (PIM) which employs a heuristic to quantify node diversity among several linear top-$k$-scorers ($\boldsymbol{\Gamma}$) to mitigate degenerate scoring, preferring the scorer that yields greater node variety.
This approach builds upon top-k pooling methods \citep{TopK:GraphUnets,TopK:SparseHierarchical,TopK:PoolingGCNNTAGCN,TopK:DifferentiableSparseTopk} originally designed for node selection in static graph coarsening without feature aggregation. 
ATk-pooling extends this with top-k feature aggregation and selecting from multiple scorers to accommodate dynamic nodes.

\paragraph{Stage 2: Local attention, C2/3.} 
Local attention layers (Equations~\eqref{eqn:local-attn-weights}--\eqref{eqn:temporal-cyclical}) construct signed latent graphs independently for spatial and temporal nodes using multihead (MH) CTC kernels with domain-specific encodings.

For \textbf{spatial attention}~(C2), $\Theta_{\spaceset}^{H}$, we use a data-driven structural encoding $\mathring{\mathbf{B}}_{\text{kern}}\in\mathbb{R}^{N\times D_{\text{kern}}}$ representing persistent baseline spatial affinities. To enable state dependency, we leverage how ReLU-based MLPs inherently generate feature segmentations in latent subspace (i.e., polytopes)~\citep{Hyperplane-DL:locallyLinearReLU2023,Hyperplane-DL:piecewiseLinearNNs2022,Hyperplane-DL:tropicalGeometryDNN2018}. These contextualized structural encodings induce dynamic node groupings in latent space, allowing dot-product kernels to form state-dependent latent graphs that capture spatial dynamics.

For \textbf{temporal attention}~(C3), $\Theta_{\timesset}^{H}$, we apply cyclical sine/cosine encoding to time-of-day (TOD) and day-of-week (DOW) as dynamic temporal metadata to encode the inherent periodicity of daily and weekly cycles. We then apply a simple linear projection to map these encodings to the model space $E$. While this dynamic encoding is optional---since Weaver operates in a traffic state-based regime---it can help improve performance, as we demonstrate in Section~\ref{section:experiments-main}.

\paragraph{Stage 3: P$^2$-KMV message passing, C5.} \textbf{Spatiotemporal Kronecker attention} (Equations~\eqref{eqn:kmv-spatial-mp}--\eqref{eqn:kmv-output}) weaves $\Theta_{\spaceset}^{H}, \Theta_{\spaceset}^{H}$ and $\tensor{U}^{H}_{\stset} \in \mathbb{R}^{H \times d_{\text{head}} \times P \times N}$ into a multihead spatiotemporal representation via \textbf{P$^2$-KMV message passing}.

\paragraph{Stage 4: Multihead consolidation, M2.} 
\textbf{MH consolidation} (Equation~\eqref{eqn:kmv-head-mixing}) applies head-mixing weights, analogous to output projection $\mathbf{W}_O$ in the original Transformer (Equation~\eqref{eqn:multihead_mixhead}). We show in Appendix~\ref{appx:weighted-KPS} that this operation is equivalent to a weighted implicit KPS (W-iKPS), which relaxes the Kronecker product separability requirement in KPA. At this strategic junction, GLU enhancement increases W-iKPS representational capacity by learning subsets of latent graph activations through multiplicative gating, followed by a residual self-loop connection from $\midtilde{\tensor{U}}_{\stset}$, yielding the consolidated spatiotemporal representation $\tensor{Z}_{\stset}\in \mathbb{R}^{Q \times N \times E}$ after rearrangement.

\subsubsection{Forecast (Figure~\ref{fig:weaver-superstructure}, D)}
The \textbf{State Transition module} predicts subsequent states using a shared residual MLP with LeakyReLU activation (D1, Equation~\eqref{eqn:forecast-mlp}) that learns the bulk system dynamics which integrate prior states and graphical interactions, followed by a linear readout layer (Equation~\eqref{eqn:forecast-readout}) that projects to the target output dimension, $C$: $\widehat{\tensor{Y}}\in \mathbb{R}^{Q\times N \times C}$.

%===== Experimental formatting=================
\section{MODEL SPECIFICATION} \label{section:model-specification}
This section provides implementation details for the \textit{Weaver} model components illustrated in Figure~\ref{fig:weaver-superstructure}, including the Traffic Phase Dictionary, Kronecker Attention module, and State Transition module. 
For clarity and organization, we refer readers to: \textit{Entmax}~\citep{entmax:peters-etal-2019-sparse}, \textit{RMSNorm}~\citep{RMSNorm:zhang2019root}, and \textit{Dropout}~\citep{Dropout:srivastava2014dropout} for implementation details.

\subsection{Input and dictionary augmentation} \label{section:WeaverModule-input/augment}

\subsubsection{Dictionary augmentation} \label{section:dictionary-augmentation}
Compute latent cofactors $\boldsymbol{\Xi} \in \mathbb{R}^{P\times N \times K_{\Xi}}$ via sparse dictionary lookup:
\begin{alignat}{3}
    \widetilde{\tensor{X}} &= \Langle\, \tensor{X}_{[P \times N \times C] \rightarrow [N \times P\,C\,]} \,\Rangle\,, \label{eqn:dict-reshape-input} \\
    S_{\Xi} (\,\midtilde{\tensor{X}} \,) &= \entmax_{1.5}\big(\operatorname{Dropout}( \glu(\widetilde{\tensor{X}}\,\asmatrix{W}_{\tilde{x}} + \asvector{b}_{\tilde{x}}))\,/\, \text{Softplus}(\boldsymbol{\tau}_{\Xi})\,\big) \in \mathbb{R}^{N \times M_{\Xi}} \,, \label{eqn:dict-sparse-coefficients} \\
    \midtilde{ \boldsymbol{\Xi} } &= \, S_{\Xi} (\,\midtilde{\tensor{X}} \,)  \,\, \tensor{C}_{\Xi} \in \mathbb{R}^{N \times P K_{\Xi} } \,, \label{eqn:dict-lookup} \\
    \boldsymbol{\Xi} &= \Langle \, \midtilde{\boldsymbol{\Xi}}_{[N \times PK_{\Xi} \rightarrow P \times N \times K_{\Xi}]} \, \Rangle \in \mathbb{R}^{P\times N \times K_{\Xi}} \,, \label{eqn:dict-reshape-output}
\end{alignat}

\subsubsection{Model space projection} \label{section:model-space-projection}

Project augmented input to model space $\tensor{U}_{\spaceset\timesset} \in \mathbb{R}^{E \times N \times P}$:
\begin{alignat}{2}
    \midtilde{\tensor{U}}_{\Xi} &= (\tensor{X} \Vert \boldsymbol{\Xi} ) \,\asmatrix{W}_{x} \, &\quad& \in\, \mathbb{R}^{P\times N \times E} \,, \label{eqn:proj-concat} \\
    \midtilde{\tensor{U}}_{\spaceset\timesset} &= \dropout \big( \glu ( \text{RMSNorm} (\midtilde{\tensor{U}}_{\Xi})\,\asmatrix{W}_{U} )\, \big) + (\midtilde{\tensor{U}}_{\Xi}\,)_{\text{res}} \, &\quad& \in\, \mathbb{R}^{P\times N \times E} \,, \label{eqn:proj-transform} \\
    \tensor{U}_{\spaceset\timesset} &=\, \Langle\, \midtilde{\tensor{U}}_{\spaceset\timesset\,[\,P\times N \times E \rightarrow E \times P \times N\,]} \,\Rangle && \in \,\mathbb{R}^{E \times P \times N} \,, \label{eqn:proj-reshape}
\end{alignat}

\subsection{Kronecker Attention Module} \label{section:WeaverModule-kronecker-attention}

\subsubsection{ATk-pooling (local node representations)} \label{section:atk-pooling}

For brevity, only the spatial ($\spaceset$) version is presented, pooling on dimension $P$ (mode-1) for $\midtilde{\tensor{U}}_{\stset}\in \mathbb{R}^{P\times N \times E}$ and yields spatial node representations $\asmatrix{U}_{\spaceset} \in \mathbb{R}^{N \times E}$. The temporal ($\timesset$) version are similar but operates along dimension $N$ and \textit{inverts} mode-1 with mode-2, yielding ${\asmatrix{U}}_{\timesset} \in \mathbb{R}^{P \times E}$. For tensor operations, see Appendix~\ref{appx:tensor-operations}; tensor mode summation is defined in Equation~\eqref{eqn:definition-tensor-mode-summation}, tensor mode variance in Equation~\eqref{eqn:tensor-mode-variance}, and top-k selection in Equation~\eqref{eqn:top-k-selection}.

\paragraph{Local spatial node representations.}
We obtain ${\asmatrix{U}}_{\spaceset}$ by aggregating features over the top-$k$ temporal nodes:
\begin{alignat}{2}
    {\asmatrix{U}}_{\spaceset} &= \sum_{i=1}^{k} \LLangle \, \midtilde{\tensor{U}}_{\stset}^{\text{Top-k}(P,\boldsymbol{\Gamma}_{\gamma^*, \spaceset})} \odot \tensor{H}_{\gamma}^{\spaceset} \, \RRangle_{\text{mode-1}}\,\,\,,
    \label{eqn:atk-spatial-aggregate}
\end{alignat}
where $k = \lceil \rho_{\timesset} \cdot P \rceil$ with pooling ratio $\rho_{\timesset}\in (0,1]$, and the supporting components are defined as:
\begin{alignat}{3}
    \tensor{H}_{\gamma}^{\spaceset} 
    &= \mathrm{Softmax}\left( \big(\boldsymbol{\Gamma} ^ {\spaceset}_{\gamma^*} \big)^{\text{Top-k}(P,\,\boldsymbol{\Gamma}_{\gamma^*}^{\spaceset})} \right) 
    &\quad &\in \mathbb{R}^{k \times N \times 1}, \label{eqn:atk-spatial-weights} \\
    \boldsymbol{\Gamma}_{\gamma^*}^{\spaceset} 
    &= \mathrm{argmax} \Big\{ \mathcal{I}_{\gamma}^{\spaceset} \big(\boldsymbol{\Gamma}_{\gamma}^{\spaceset} \big) \Big\}
    &\quad &\in \mathbb{R}^{P \times N \times 1}, \label{eqn:atk-spatial-best} \\
    \mathcal{I}_{\gamma}^{\spaceset} (\boldsymbol{\Gamma}_{\gamma}^{\spaceset} ) 
    &= \sum_{n=1}^N \LLangle\, \mathrm{Var}_{\text{mode-1}}\big( \boldsymbol{\Gamma}_{\gamma}^{\spaceset} \big)\, \RRangle_{\text{mode-2}}
    &\quad &\in \mathbb{R}^{M_{\gamma}^{\spaceset}}, \label{eqn:atk-spatial-informativeness} \\
    \boldsymbol{\Gamma}_{\gamma}^{\spaceset} 
    &= \midtilde{\tensor{U}}_{\stset} \, \asmatrix{W}_{\gamma}^{\spaceset}
    &\quad &\in \mathbb{R}^{P \times N \times M_{\gamma}^{\spaceset}}. \label{eqn:atk-spatial-scores}
\end{alignat}
Here, $\tensor{H}_{\gamma}^{\spaceset}$ are the aggregation weights computed via top-$k$ selection and softmax normalization; $\boldsymbol{\Gamma}_{\gamma^*}^{\spaceset}$ is the scorer that maximizes the pooling informativeness metric (PIM), $\mathcal{I}_{\gamma}^{\spaceset}$, which heuristically summarizes scoring variety by computing mode-1 variance summed over mode-2; and $\boldsymbol{\Gamma}_{\gamma}^{\spaceset}$ are the pooling scores computed using vector-normalized weights $\asmatrix{W}_{\gamma}^{\spaceset} = \big\Vert_{i=1}^{M_{\gamma}^{\spaceset}} \frac{ \asvector{w}_{\gamma,\, i}^{\spaceset} } {\Vert \asvector{w}_{\gamma,\, i}^{\spaceset} \Vert} \in \mathbb{R}^{E \times M_{\gamma}^{\spaceset}}$, where $M_{\gamma}^{\spaceset}$ is the number of scorers.

\subsubsection{Local attention layers (latent graphs, MP1)} \label{section:local-attention}
Both local spatial- and temporal-attention layers follow a common architecture with domain-specific encodings, where spatial attention operates over nodes ($\spaceset$, dimension $N$) and temporal attention operates over time steps ($\timesset$, dimension $P$), with $E = H d_{\text{head}}$ for multihead~(MH). The Valence Attention weights are computed using MH continuous Tanimoto coefficient (CTC) kernels (Equation~\eqref{eqn:definition-tanimoto-vector-qk}) in tensor form (omitting subscripts $\spaceset$ or $\timesset$ for legibility):
\begin{align}
    \Theta^H &= \, \frac{\tensor{Q}\tensor{K}^\top}{\Vert\tensor{Q}\Vert^2_{\text{vec}} 
    + (\,\Vert\tensor{K}\Vert^2_{\text{vec}}\,)^\top - \tensor{Q}\tensor{K}^\top + \epsilon_0}\,, \label{eqn:local-attn-weights}
\end{align}
with $\Theta_{\spaceset} \in \mathbb{R}^{H \times N \times N}$ for spatial and $\Theta_{\timesset} \in \mathbb{R}^{H \times P \times P}$ for temporal attention. Vector norms are:
\begin{align}
\lVert\tensor{Q}\rVert_{\text{vec}}^2 &= \sum_{d} \tensor{Q}_{h\,\ldots\, d}^2, \quad
\lVert\tensor{K}\rVert_{\text{vec}}^2 = \sum_{d} \tensor{K}_{h \,\ldots\, d}^2. \label{eqn:local-attn-norms}
\end{align}
Multihead queries and keys are constructed via:
\begin{alignat}{2}
    \tensor{Q} &= \Langle \asmatrix{Q}_{\,[\cdots\, \times HD_{\text{head}}\, \rightarrow H\times \cdots\, \times d_{\text{head}}\,]} \Rangle\,, &\quad \tensor{K} &= \Langle \asmatrix{K}_{\,[\cdots\, \times HD_{\text{head}}\, \rightarrow H\times \cdots\, \times d_{\text{head}}\,]} \Rangle \quad \text{(Figure~\ref{fig:weaver-superstructure}, M1)}, \label{eqn:local-attn-reshape} \\
    \asmatrix{Q} &= \asmatrix{G}\,\asmatrix{W}_{\text{Q}}\,, &\quad \asmatrix{K} &= \asmatrix{G}\,\asmatrix{W}_{\text{K}}\,, \label{eqn:local-attn-projections}
\end{alignat}
where $\asmatrix{W}_{\text{Q}}, \asmatrix{W}_{\text{K}} \in \mathbb{R}^{E\times E}$, and $\asmatrix{G}$ represents the structurally-encoded representations that incorporate domain-specific inductive biases (spatial structure or dynamic metadata) into the query-key computations.
\paragraph{Spatial structural encoding.} The spatial representations $\asmatrix{G}_{\spaceset} \in \mathbb{R}^{N \times E}$ learns node relational information through a multi-layer network with $\operatorname{ReLU}(\,\cdot\,)$:
\begin{align}
    \asmatrix{G}_{\spaceset} &= 
    \text{ReLU}_{(\kappa-1)}\big(\ldots\,\text{ReLU}_{(0)} \big( ( \asmatrix{U}_{\spaceset} \Vert \mathring{\asmatrix{B}}_{\text{kern}} ) \, \asmatrix{W}^{\spaceset}_{0} \, \big)\asmatrix{W}^{\spaceset}_{1} \,\ldots\big) \asmatrix{W}^{\spaceset}_{\kappa} \,, \label{eqn:spatial-representations}
\end{align}
where $\mathring{\asmatrix{B}}_{\text{kern}} \in \mathbb{R}^{N \times D_{\text{kern}}}$ learns the graph kernel structure; the projection weights are:
\begin{align*}
\asmatrix{W}^{\spaceset}_{0} &\in \mathbb{R}^{(E+D_{\text{kern}}) \,\times\, D_1}, \quad
\asmatrix{W}^{\spaceset}_{\ell} \in \mathbb{R}^{D_{\ell} \times D_{\ell+1}} \text{ for } \ell\in [\kappa-1], \quad
\asmatrix{W}^{\spaceset}_{\kappa} \in\mathbb{R}^{D_{\kappa}\times E}.
\end{align*}
\paragraph{Temporal dynamic encoding.} This can be omitted if temporal metadata is unavailable. The temporal representations $\asmatrix{G}_{\timesset} \in \mathbb{R}^{P \times E}$ incorporate cyclical time features:
\begin{align}
    \asmatrix{G}_{\timesset} &=  ( \asmatrix{U}_{\timesset} \Vert \mathring{\asmatrix{B}}_{\text{dyn}} ) \, \asmatrix{W}^{\timesset} \,, \label{eqn:temporal-representations}
\end{align}
where $\mathring{\asmatrix{B}}_{\text{dyn}} \in \mathbb{R}^{P \times 4}$ concatenates cyclical encodings of time-of-day (TOD) and day-of-week (DOW):
\begin{align}
    \mathring{\asmatrix{B}}_{\text{dyn}} &= \big[\, \mathring{\asvector{b}}_{\text{cyc}}(t_{\text{TOD}}, 1440) \,\Vert\, \mathring{\asvector{b}}_{\text{cyc}}(t_{\text{DOW}}, 7) \,\big], \label{eqn:temporal-dynamic} \\
    \mathring{\asvector{b}}_{\text{cyc}}(t, T_{\text{period}}) &= \Big[\, \sin\,\big(\frac{2\pi\, t}{T_{\text{period}}} \big) \,, \cos\,\big(\frac{2\pi\, t}{T_{\text{period}}} \big)   \, \Big] \in \mathbb{R}^{2}\,, \label{eqn:temporal-cyclical}
\end{align}
with $t_{\text{TOD}} \in [0, 1439]$ (minutes), $t_{\text{DOW}} \in [1, 7]$ (days), and $\asmatrix{W}^{\timesset} \in \mathbb{R}^{(E+4) \times E}$. 

\subsubsection{Spatiotemporal Kronecker attention (MP2)} \label{section:st-kronecker-attention}
\paragraph{Spatiotemporal message passing.} Apply multihead spatiotemporal MP using P$^2$-KMV with $\tensor{U}_{\stset}$ from Equation~\eqref{eqn:proj-reshape}:
\begin{alignat}{2}
    \tensor{U}_H^{\langle 2 \rangle} &= \tensor{V}_{H, \stset}^{\langle 3 \rangle} (\Theta^{H}_{\spaceset})^\top &&\in \mathbb{R}^{H \times d_{\text{head}} P \times N}, \label{eqn:kmv-spatial-mp} \\
    \tensor{U}_H^{\langle 1 \rangle} &= \tensor{U}^{\langle 2 \rangle}_{H,\dotumble} (\Theta^{H}_{\timesset})^\top &&\in \mathbb{R}^{H \times d_{\text{head}}  N \times P}, \label{eqn:kmv-temporal-mp} \\
    \tensor{Z}_{H,\stset} &= \tensor{U}^{\langle 1 \rangle}_{H,\dotumble} && \in \mathbb{R}^{H \times P N \times d_{\text{head}} }\,. \label{eqn:kmv-output}
\end{alignat}
where $\tensor{V}_{H,\stset}^{\langle 3 \rangle} = \LLangle \, \tensor{U}_{\stset\,[H\,d_{\text{head}} \times P \times N \,\rightarrow\, H \times d_{\text{head}} P \times N ]} \, \RRangle$ and $\tensor{U}_{H,\dotumble}^{\depth{2}} \in \mathbb{R}^{H \times d_{\text{head}} N \times P}$.

\paragraph{Multihead consolidation.} Head-mixing weights with GLU enhancement and residual connection:
\begin{align}
    \tensor{Z}_{\stset} &= \glu(\midtilde{\tensor{Z}}_{\stset} \asmatrix{W}_{2O}) + (\midtilde{\tensor{U}}_{\stset})_{\text{res}}\,, \label{eqn:kmv-head-mixing}
\end{align}
where $\asmatrix{W}_{2O} \in \mathbb{R}^{E \times 2E}$ and $\midtilde{\tensor{Z}}_{\stset} = \LLangle\, \tensor{Z}_{H,\stset \,[H \times PN \times d_{\text{head}} \rightarrow P \times N \times HD_{\text{head}}]} \,\RRangle\,$.

\subsection{Forecast module} \label{section:WeaverModule-state-transition}
Apply residual MLP with expansion factor $M_z=2$:
\begin{align}
    \tensor{Z}^{[1]}_{\stset} 
    &= \Big( \text{Dropout}\big( \text{LeakyReLU} (\tensor{Z}^{}_{\stset}\,\asmatrix{W}_{z,\text{up}} + \asvector{b}_{z,\text{up}} ) \big)\,\asmatrix{W}_{z,\text{dn}} + \asvector{b}_{z,\text{dn}} \Big) + \tensor{Z}^{}_{\stset}\,, \label{eqn:forecast-mlp}
\end{align}
followed by linear readout to forecast horizon $Q$:
\begin{align}
    \widehat{\tensor{Y}} 
    &= \tensor{Z}^{[1]}_{\stset}\,\asmatrix{W}_{\text{ro}} + \asvector{b}_{\text{ro}}\,, \label{eqn:forecast-readout}
\end{align}
where
\begin{align*}
\asmatrix{W}_{z,\text{up}} &\in \mathbb{R}^{E \times M_z E}, \quad \asvector{b}_{z,\text{up}} \in \mathbb{R}^{M_z E}, \\
\asmatrix{W}_{z,\text{dn}} &\in \mathbb{R}^{M_z E \times E}, \quad \asvector{b}_{z,\text{dn}} \in \mathbb{R}^{E}, \\
\asmatrix{W}_{\text{ro}} &\in \mathbb{R}^{E \times QC}, \quad \asvector{b}_{\text{ro}} \in \mathbb{R}^{QC}.
\end{align*}

\subsection{Training objective} \label{section:training-objective}
The model uses Mean Absolute Error (MAE) training loss on original scale, consistent with prior spatiotemporal models in this domain:
\begin{align}
    \tensor{L}_{\text{MAE}} &= \frac{1}{Q \, N \, C} 
    \sum_{q=1}^{Q} \sum_{n=1}^{N} \sum_{c=1}^{C} 
    \big| \tensor{Y}_{qnc} - \widehat{\tensor{Y}}_{qnc} \big| \,, \label{eqn:mae-loss}
\end{align}
where $\tensor{Y} \in \mathbb{R}^{Q \times N \times C}$ is the ground truth and $\widehat{\tensor{Y}} \in \mathbb{R}^{Q \times N \times C}$ is the predicted output.
%===== Experimental formatting=================

\section{EXPERIMENTS and RESULTS} \label{section:experiments-main}

\subsection{Experiment Setup} \label{sec:experimental-setup}

\begin{figure}[width=.99\linewidth,cols=4,pos=t]
    \centering
    \begin{subfigure}[t]{0.45\textwidth}
        \centering
        \includegraphics[width=\linewidth]{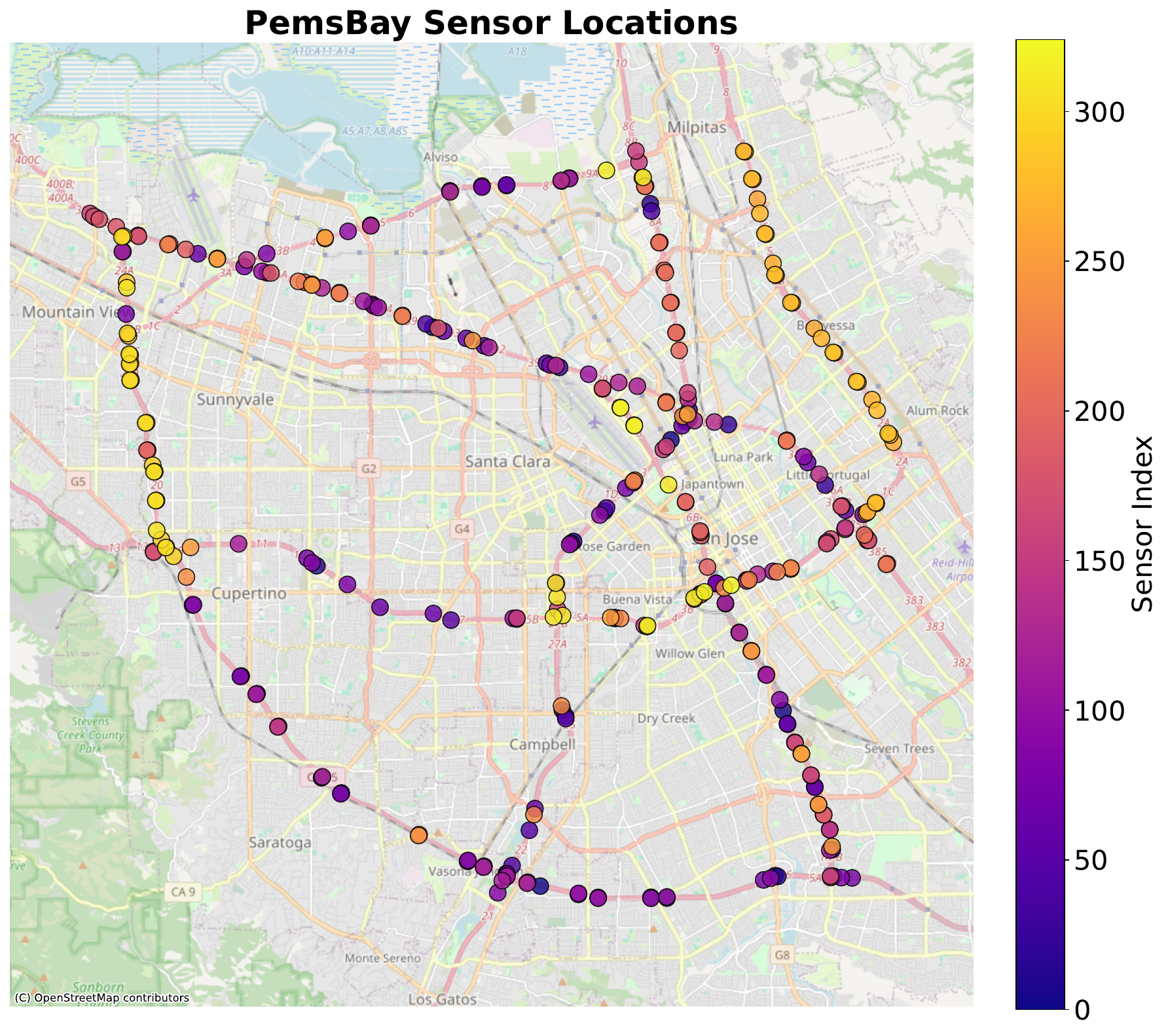}
        \caption{PemsBay sensor network in San Francisco Bay.}
        \label{fig:pemsbay_sensors}
    \end{subfigure}
    \hfill
    \begin{subfigure}[t]{0.45\textwidth}
        \centering
        \includegraphics[width=\linewidth]{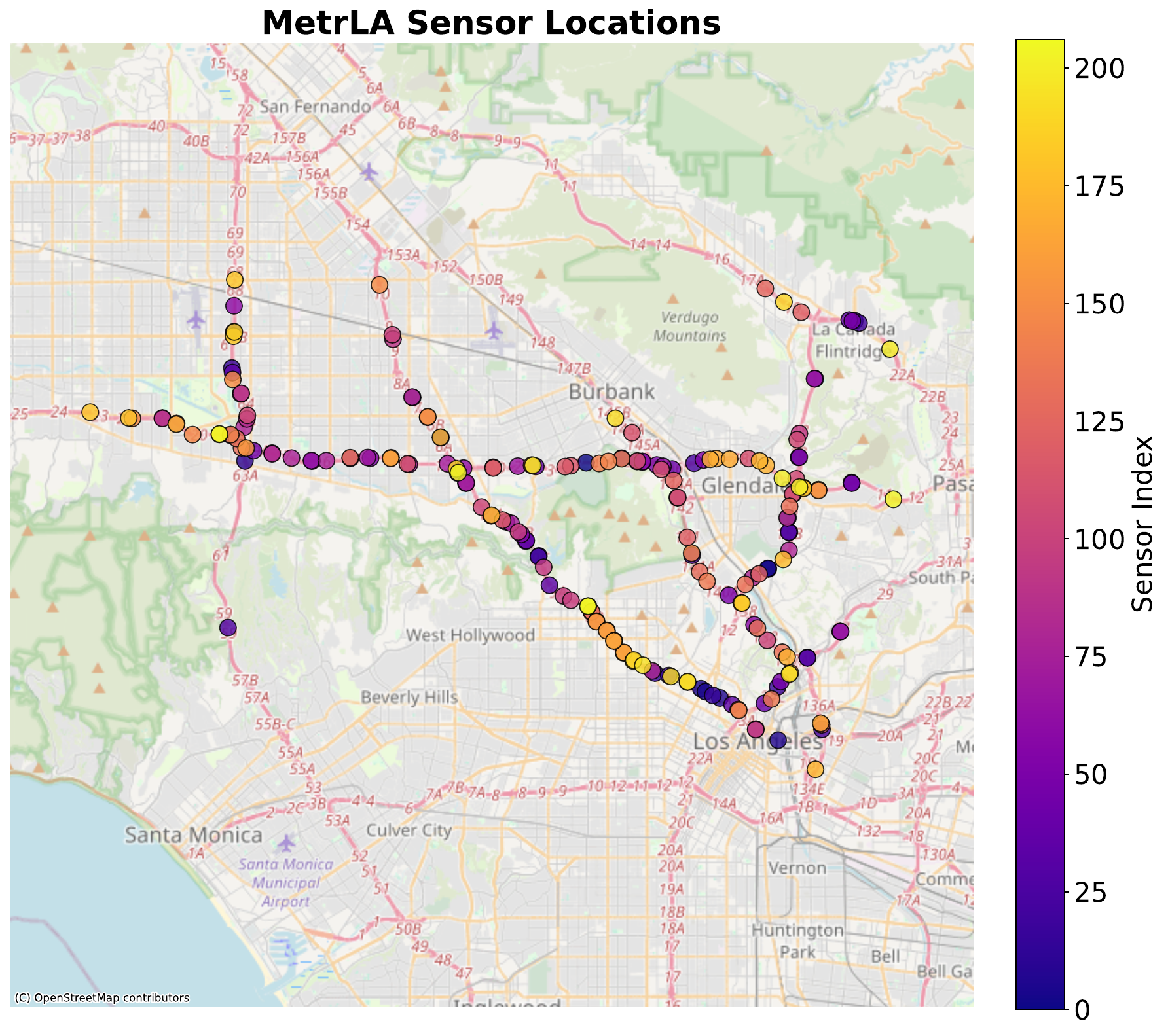}
        \caption{MetrLA sensor network in the Los Angeles metropolitan.}
        \label{fig:metrla_sensors}
    \end{subfigure}
    \caption{Geographic distribution of traffic sensors across two major California highway networks. Color indicates sensor index (different from sensor ID) corresponding to the order of appearance in the original dataset from TSL.}
    \label{fig:sensor_locations}
\end{figure}

\begin{figure}[width=.99\linewidth,cols=4,pos=t]
    \centering
    \includegraphics[width=0.86\linewidth]{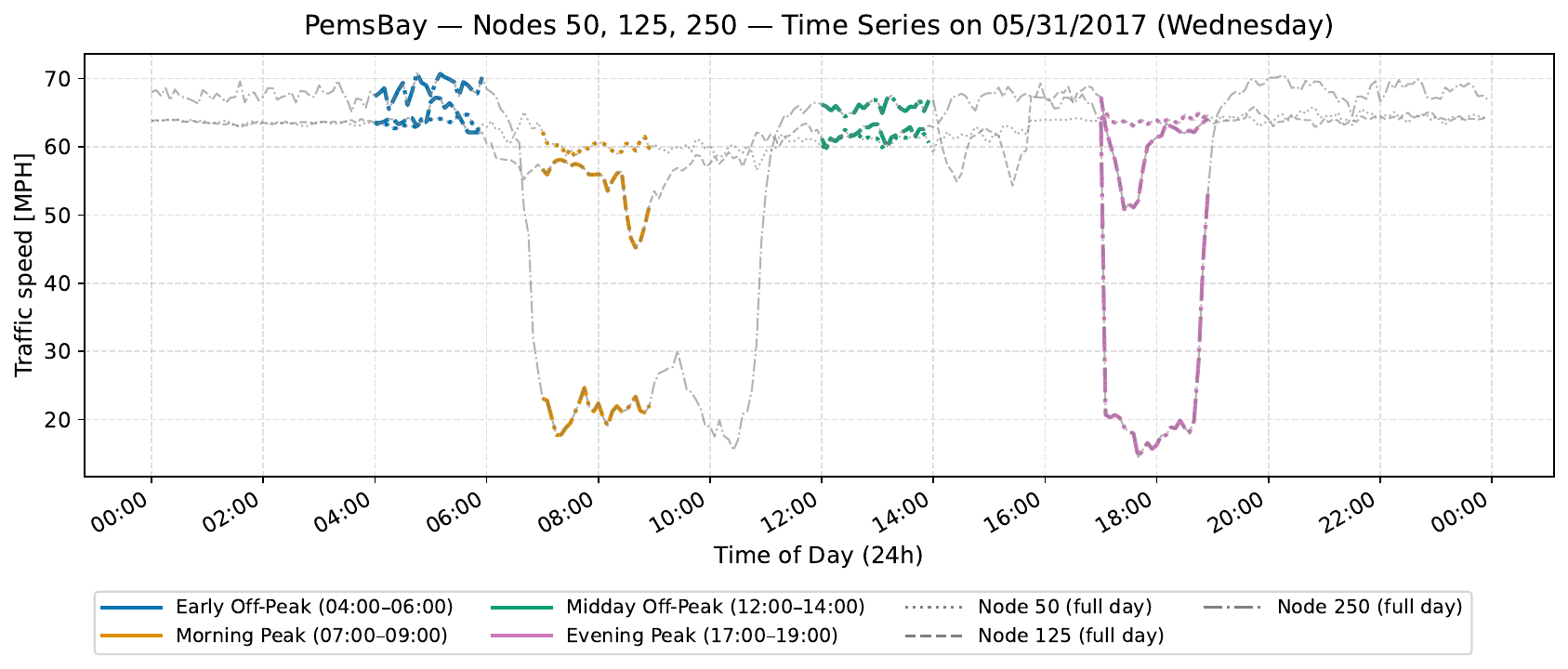}
    \caption{Example vehicle speeds [MPH] time series sampled from the PemsBay dataset on 05/31/2017 (Wednesday) from three nodes (node index: 50, 125, 250, in order of appearance in the TSL database), four distinct traffic phases are indicated: early off-peak, morning peak, midday off-peak, evening peak.}
    \label{fig:pemsbay-example-series}
\end{figure}

Training was conducted on the \textbf{NVIDIA A100} GPU (Google Colab) using the \textbf{Torch Spatiotemporal Library (TSL)} \citep{TSL:Cini_Torch_Spatiotemporal_2022} with \textit{PyTorch 2.8.0+cu126}. Following standard procedures~\citep{DirectedGCN:li2018dcrnn}, data is split chronologically into training, validation, and testing sets (70:10:20). Two real-world traffic datasets were used \citep{DCRNN:li2018,TSL:Cini_Torch_Spatiotemporal_2022}: 
\begin{enumerate}
    \item \textbf{PemsBay} (Figure~\ref{fig:pemsbay_sensors}): 6 months of traffic speed readings [MPH] collected every 5 minutes by 325 sensors on highways in the San Francisco Bay Area from 01/01/2017 to 05/31/2017. The reported missing rate is $0.02\%$.
    \item \textbf{MetrLA} (Figure~\ref{fig:metrla_sensors}): 4 months of traffic speed readings [MPH] collected every 5 minutes by 207 sensors on highways in Los Angeles County from 03/01/2012 to 06/30/2012. The reported missing rate is $8.11\%$.
\end{enumerate}
For illustration, Figure~\ref{fig:pemsbay-example-series} shows a sample full-day time series from three nodes (50, 125, 250). The \textbf{task} is to forecast 12 time steps forward (60 minutes, 5-minute intervals) given 12 time steps of input; \textbf{sequence slicing} in TSL produces overlapping 2-hour windows with a stride of one time step.

\paragraph{Training configuration.} To minimize reliance on optimization-specific tuning, all models were trained under a generic a priori configuration: 32-sample batches (2 workers), Adam optimizer, and an exponential learning-rate scheduler ($\gamma=0.90$, $\alpha=0.001$), for up to 50 epochs with early stopping (patience = 10, minimum validation loss decrease: $\Delta_{\min}=10^{-3}$). Temporal metadata and graph connectivity were provided as required by specific architectures, with graph adjacency constructed using TSL defaults.

\paragraph{Data preprocessing.}
Default TSL masks $\tensor{M}\in\{0,1\}^{S\times N\times C}$ are applied ($S$: number of samples), and feature-wise standardization is performed \emph{before} sequence slicing (\texttt{StandardScaler(axis=(0,1))}). Mean ($\mu_k$) and standard deviation ($\sigma_k$) are computed using only training data to avoid information leakage:
\begin{align}
\midbar{X}_{ijk} &= M_{ijk}\,\frac{X_{ijk}-\mu_k}{\sigma_k + \epsilon_{\sigma}}, \\
\mu_k &= \frac{1}{n_k}\sum_{i,j} M_{ijk}X_{ijk}, \qquad
\sigma_k = \sqrt{\frac{1}{n_k}\sum_{i,j} M_{ijk}(X_{ijk}-\mu_k)^2},
\end{align}
where $n_k = \sum_{i,j} M_{ijk}$ is the count of observed entries for channel $k$, and $\epsilon_{\sigma}=10^{-8}$ avoids division by zero. This produces standardized inputs $\midbar{\tensor{X}}\in\mathbb{R}^{S\times N\times C}$.

\paragraph{Metric reporting.}
All metrics and losses are computed on masked TSL datasets using the original data scale to preserve real-world error magnitudes. 
Values are averaged across multiple trials with different random seeds, with 95\% confidence intervals calculated using Student's \emph{t}-distribution. 
Let $y_i$ and $\hat{y}_i$ denote the ground-truth and predicted values for $n$ valid data points within the forecast horizon. 
Table~\ref{tab:result} reports the following benchmark metrics:
\begin{align}
\text{MAE} &= \tfrac{1}{n}\sum_{i=1}^{n} |y_i - \hat{y}_i|\,, \quad \\ 
\text{MAPE} &= \tfrac{100\%}{n}\sum_{i=1}^{n} \Big|\tfrac{y_i - \hat{y}_i}{y_i}\Big|\,, \quad \\
\text{RMSE} &= \sqrt{\tfrac{1}{n}\sum_{i=1}^{n} (y_i - \hat{y}_i)^2}\,\,.
\end{align}
Here, \textbf{MAE} (Mean Absolute Error) measures the average magnitude of prediction errors, providing an interpretable value in the same unit as the data.  
\textbf{MAPE} (Mean Absolute Percentage Error) expresses the mean relative error as a percentage of the true values, reflecting proportional accuracy.  
\textbf{RMSE} (Root Mean Squared Error) emphasizes larger deviations by squaring residuals prior to averaging, thus being more sensitive to outliers and high-variance forecasts.

\subsubsection{Weaver hyperparameters}

The default values for Weaver are shown in Table~\ref{tab:WeaverHyperparams}, representing general-purpose configurations found to perform reliably across datasets without task-specific hyperparameter tuning. Ablation studies are provided in Section~\ref{section:ablations}.

\begin{table}[width=.99\linewidth,cols=4,pos=b]
\centering
\caption{Summary of Weaver hyperparameter values and configurations used in experiments. "Space metadata" describe network spatial configurations, e.g., sensor coordinates or graph connectivity. "Time metadata" are dynamic exogenous variables, i.e., cyclically encoded time-of-day (TOD) and day-of-week (DOW); "NT" (No Time) excludes time metadata. }
\resizebox{0.95\textwidth}{!}{
\begin{tabular}{l||c|c|c|c|c|cc|ccc|cc}
\toprule\toprule
Configuration 
 & \makecell{Heads\\$H$} %1
 & \makecell{Embed\\size\\$E$} %2
 & \makecell{Time\\metadata} %3
 & \makecell{Space\\metadata} %4
 & \makecell{Dropout} %5
 & \multicolumn{2}{c|}{Traffic Phase Dictionary} %6-7
 & \multicolumn{3}{c|}{ATk-Pooling} %8-10
 & \multicolumn{2}{c}{K-Bias} \\ %11-12
\cline{7-13}
& %0
& %1
& \makecell{\scriptsize (TOD/DOW)} %2
& %3
& %4
& \makecell{Count\\$M_{\Xi}$} %5
& \makecell{Width\\$K_{\Xi}$} %6
& \makecell{Num.\\Scorers\\$N_{\kappa}$} %7
& \makecell{pooling\\ratio, $\rho_{\spaceset}$\\(space)} %8
& \makecell{pooling\\ratio, $\rho_{\timesset}$\\(time)} %9
& \makecell{Count\\$D_{\text{kern}}$} %10
& \makecell{Num.\\Layers} %11
\\
\midrule
\textbf{Weaver} 
& - 
& - 
& - 
& - 
& - 
& - 
& - 
& - 
& - 
& - 
& - 
& - 
\\
%\midrule
$\triangleright$ Default(T)
& 8
& 128
& \checkmark
& \xmark
& 0.1 
& 64 
& 32 
& 5 
& 0.6 
& 0.6 
& 32
& 3
\\
$\triangleright$ Default(NT) 
& 8
& 128
& \xmark
& \xmark
& 0.1 
& 64 
& 32 
& 5 
& 0.6 
& 0.6 
& 32
& 3
\\
% $\triangleright$ Default(T)
% & 8
% & 128
% & \checkmark
% & \xmark
% & 0.1
% & 64 
% & 32 
% & 5 
% & 0.6 
% & 0.6 
% & 32
% & 3
% \\
% $\triangleright$ MetrLA(NT)
% & 8
% & 128
% & \xmark
% & \xmark
% & 0.1
% & 64 
% & 32 
% & 5 
% & 0.6 
% & 0.6 
% & 32
% & 3
% \\
% \midrule
% $\triangleright$ [Special Config] 
% & [ ] 
% & [ ] 
% & [ ] 
% & [ ] 
% & [ ] 
% & [ ] 
% & [ ] 
% & [ ] 
% & [ ] 
% & [ ] 
% & [ ] 
% & [ ] 
% \\
\bottomrule\bottomrule
\end{tabular}
}
\label{tab:WeaverHyperparams}
\end{table}

\begin{table}[width=.99\linewidth,cols=4,pos=b]
\caption{Comparison of time and space complexity of Weaver versus Transformer models, where \textit{L} denotes the number of stacked Transformer layers. "\textit{Graph connectivity}" refers to handcrafted network graph features; "\textit{dynamic metadata}" refers to time-varying exogenous features (e.g., time-of-day, day-of-week, weather). "N/A" indicates insufficient information.}
    \centering
    \resizebox{0.99\textwidth}{!}{%
        \begin{tabular}{c|c|c|c|c|c|cc}
            \toprule\toprule
            \multirow{2}{*}{Model} & \multirow{2}{*}{Time complexity} & \multirow{2}{*}{Space complexity} & \multirow{2}{*}{\makecell{Learnable\\ embedding}} & \multirow{2}{*}{\makecell{Fixed\\positional\\encoding}} & \multirow{2}{*}{\makecell{Structural\\ encoding}} & \multicolumn{2}{c}{\textbf{Metadata}} \\
            \cmidrule(lr){7-8}
            & & & & & & \makecell{Graph\\ connectivity} & \makecell{Dynamic\\ metadata} \\
            \midrule\midrule
            Vanilla Transformer & $\Ocomplex\left(L P^2 N^2\right)$ & $\Ocomplex\left(L P^2 N^2\right)$ & N/A & Yes & N/A & N/A & N/A \\
            \midrule
            STTN & $\Ocomplex\left(L(P^2 + N^2)\right)$ & $\Ocomplex\left(L(P^2 + N^2)\right)$ & Yes & N/A & Yes, learned & Yes & Optional \\
            \midrule
            STAEformer & $\Ocomplex\left(LPN(P + N)\right)$ & $\Ocomplex\left(LPN(P + N)\right)$ & Yes & Yes & N/A & No & Yes \\
            \midrule
            \textbf{Weaver (proposed)} & $\Ocomplex\left(PN(P + N)\right)$ & $\Ocomplex\left(P^2 + N^2\right)$ & Yes & No & Yes, learned & Optional & Optional \\
            \bottomrule\bottomrule
        \end{tabular}
    }
    \label{tab:complexity_comps}
\end{table}

\subsubsection{Benchmark models}

To ensure consistency and fairness across all benchmarks, \textbf{every model was uniformly provided with a 1-hour historical input window} (12 time steps of 5 minutes), reflecting practical data availability and storage constraints. Representative models were selected from two architectural families, with hyperparameters reproduced from their original sources to the best of our knowledge (Table~\ref{tab:appendix_model_manifest}, Appendix~\ref{appx:benchmarks-setup}): 
\begin{enumerate}
    \item Transformer-based (computational complexities compared in Table~\ref{tab:complexity_comps}):
    \begin{enumerate}
        \item STTN \citep{STTN:xu2021spatial}: 
        Learns separate spatial ($\mathbb{R}^{N\times N}$) and temporal ($\mathbb{R}^{P\times P}$) embeddings to encode road topology and time-step information. These embeddings are concatenated with node features and passed through stacked spatial-temporal blocks, where multi-head attention alternates between spatial and temporal transformers. Moreover, a fixed Chebyshev graph convolution layer encodes stationary spatial structure alongside spatial attention.
        \item STAEformer \citep{STAE:liu2023staeformer}:
        Learns lookup tables $T_w \in \mathbb{R}^{N_w \times d_f}$ and $T_d \in \mathbb{R}^{N_d \times d_f}$, indexed respectively by day-of-week ($0$–$6$. $N_w=7$) and time-of-day ($0$–$287$, corresponding to 5-minute intervals over 24 hours, $N_d=288$). The retrieved embeddings ($\asmatrix{E}_w^t \in \mathbb{R}^{T \times d_f}$, $\asmatrix{E}_d^t \in \mathbb{R}^{T \times d_f}$) are concatenated and broadcast across all nodes to form the periodicity embedding $\tensor{E}_p \in \mathbb{R}^{T \times N \times 2d_f}$. This is further concatenated with a learned spatiotemporal embedding ($\mathcal{E}_a \in \mathbb{R}^{T \times N \times d_a}$) and node-wise feature inputs ($\tensor{X} \in \mathbb{R}^{P \times N \times 1}$). The resulting joint representation is processed by sequential temporal and spatial transformer layers, where multi-head attention alternates along each axis ($\mathbb{R}^{H \times P \times N \times N} \leftrightarrow \mathbb{R}^{H \times N \times P \times P}$) to capture spatiotemporal dependencies.
    \end{enumerate}
    \item GCN-based:
    \begin{enumerate}
        \item Graph Wavenet \citep{GraphWavenet:wu2019}: 
        Employs stacked dilated causal convolutional layers (directed, forward-time) to model long-range temporal dependencies, combined with diffusion-based graph convolutions that separately process forward and backward directional flows. The model uses a fixed, distance-derived adjacency matrix, augmented by a learnable adjacency matrix trained end-to-end to refine graph connectivity.
        \item ASTGCN \citep{ASTGCN:guo2019astgcn}: 
        Analyzes traffic dynamics in three timescales across separate model blocks: recent (hourly), daily-periodic, and weekly-periodic sequences. Within each block, spatial and temporal attention modules adaptively weight correlations, while Chebyshev graph convolutions and 1D temporal convolutions extract spatial and temporal features. The outputs are fused through concatenation and learnable weighting to produce the final forecast. In our benchmark, only the one-hour history window is provided for uniformity.
    \end{enumerate}
\end{enumerate}

\subsection{Results and Discussion}
Table~\ref{tab:result} reports benchmarking results on both PemsBay and MetrLA using a standardized training configuration (Section~\ref{sec:experimental-setup}). Baseline implementations occasionally yield different values from their original publications because all models are trained under identical optimizer settings and hyperparameters, ensuring that performance differences arise primarily from model design rather than tuning effort.

On \textbf{PemsBay} (minimal missing data, 0.02\%), Graph WaveNet achieves the best performance at the 15-minute horizon across all metrics. However, Weaver (Default(T)) surpasses all baselines at longer horizons (30 and 60 minutes), and achieves the best overall average performance (`All'). This indicates that Weaver distributes prediction error differently across time horizons---with particular strength in long-range prediction. The advantage is consistent with the design of Kronecker Attention, whose TEN formulation enables higher-order interactions between node, time, and feature dimensions.

\begin{table}[width=.99\linewidth,cols=4,pos=b]
\captionsetup{width=\textwidth} % or a smaller width
\caption{Performance metrics (RMSE, MAPE, MAE) averaged on the number of trials (Num. trials) across \textbf{prediction horizons (15, 30, 60 minutes)} and \textbf{overall metrics (All)} across all horizons (15, 30, 45, 60 minutes). The corresponding \textbf{95\% error margins} are included in subscript: $\textstyle(\pm X.XX)$. Results are averaged over multiple random seeds, potentially differing from the original papers. Lower values are better, the best results per dataset are marked with \textbf{bold} and asterisk (*).}
\centering
\resizebox{\textwidth}{!}{
\begin{tabular}{c|l|cccc|cccc|cccc|c}
\toprule\toprule
\multirow{2}{*}{\textbf{Data} [Units]}  &  \multirow{2}{*}{Model} & \multicolumn{4}{c|}{$\text{Avg. RMSE}_{(95\% Err.)}$}  & \multicolumn{4}{c|}{$\text{Avg. MAPE}_{(95\% Err.)}$}  &  \multicolumn{4}{c|}{$\text{Avg. MAE}_{(95\% Err.)}$} & \multirow{2}{*}{Num.}
\\\cmidrule(lr){3-6}\cmidrule(lr){7-10}\cmidrule(lr){11-14}
(Missing rate)
 & (Num. trials) & 15~min & 30~min &  60~min & All & 15~min & 30~min  & 60~min & All & 15~min & 30~min  & 60~min & All &  Params.
\\\midrule\midrule
\multirow{7}{*}{\shortstack{\textbf{PemsBay}\\ 2017\\ Speed [MPH] \\ (0.02\%)}} 
& STTN (10) & $\subperror{2.86}{0.01}$ & $\subperror{3.81}{0.01}$ & $\subperror{4.62}{0.02}$ & $\subperror{3.73}{0.02}$ & $\subperror{2.91}{0.01}$ & $\subperror{3.85}{0.02}$ & $\subperror{4.81}{0.03}$ & $\subperror{3.72}{0.02}$ & $\subperror{1.37}{0.01}$ & $\subperror{1.71}{0.01}$ & $\subperror{2.05}{0.01}$ & $\subperror{1.65}{0.01}$ & 0.59M \\
& STAEformer (9)${}^{1}$ & $\subperror{2.92}{0.03}$ & $\subperror{3.90}{0.04}$ & $\subperror{4.56}{0.04}$ & $\subperror{3.76}{0.03}$ & $\subperror{2.99}{0.05}$ & $\subperror{3.93}{0.06}$ & $\subperror{4.73}{0.07}$ & $\subperror{3.75}{0.05}$ & $\subperror{1.36}{0.01}$ & $\subperror{1.68}{0.01}$ & $\subperror{1.95}{0.02}$ & $\subperror{1.61}{0.01}$ & 1.40M \\
& ASTGCN (10) & $\subperror{3.10}{0.03}$ & $\subperror{4.35}{0.06}$ & $\subperror{5.70}{0.14}$ & $\subperror{4.38}{0.08}$ & $\subperror{3.09}{0.01}$ & $\subperror{4.49}{0.06}$ & $\subperror{6.07}{0.16}$ & $\subperror{4.37}{0.07}$ & $\subperror{1.47}{0.01}$ & $\subperror{1.95}{0.02}$ & $\subperror{2.51}{0.04}$ & $\subperror{1.90}{0.02}$ & 0.50M \\
& Graph WaveNet (10) & $\mathbf{2.73}^*_{\mathit{(\pm0.01)}}$ & $\mathbf{3.67}^*_{\mathit{(\pm0.01)}}$ & $4.45_{\mathit{(\pm0.01)}}$ & $3.60_{\mathit{(\pm0.02)}}$ 
& $\mathbf{2.69}^*_{\mathit{(\pm0.01)}}$ & $\mathbf{3.63}^*_{\mathit{(\pm0.01)}}$ & $4.59_{\mathit{(\pm0.02)}}$ & $3.50_{\mathit{(\pm0.01)}}$ 
& $\mathbf{1.30}^*_{\mathit{(\pm0.00)}}$ & $1.63_{\mathit{(\pm0.01)}}$ & $1.96_{\mathit{(\pm0.01)}}$ & $1.58_{\mathit{(\pm0.04)}}$ & 0.30M \\
& \textbf{Weaver} (-) & - & - & - & - & - & - & - & - & - & - & - & - & - \\ 
& $\triangleright$ \textbf{Default(T)} (10) 
& $2.79_{\perror{0.01}}$ & $\mathbf{3.67}^*_{\perror{0.01}}$ & $\mathbf{4.27}^*_{\perror{0.01}}$ & $\mathbf{3.55}^*_{\perror{0.01}}$
& $2.77_{\perror{0.00}}$ & $\mathbf{3.63}^*_{\perror{0.01}}$ & $\mathbf{4.36}^*_{\perror{0.02}}$ & $\mathbf{3.46}^*_{\perror{0.01}}$
& ${1.32}_{\perror{0.00}}$ & $\mathbf{1.61}^*_{\perror{0.00}}$ & $\mathbf{1.85}^*_{\perror{0.00}}$ & $\mathbf{1.54}^*_{\perror{0.00}}$ & 0.34M \\
& $\triangleright$ \textbf{Default(NT)} (10) 
& $2.81_{\perror{0.01}}$ & $3.71_{\perror{0.01}}$ & $4.40_{\perror{0.01}}$ & $3.61_{\perror{0.01}}$ 
& ${2.76}_{\perror{0.01}}$ & $3.64_{\perror{0.01}}$ & $4.47_{\perror{0.02}}$ & $3.49_{\perror{0.01}}$
& ${1.32}_{\perror{0.00}}$ & $1.63_{\perror{0.00}}$ & $1.91_{\perror{0.00}}$ & $1.57_{\perror{0.00}}$ & 0.33M \\
% & $\triangleright$ \textbf{[Variant C]} ([--]) & [--] & [--] & [--] & [--] & [--] & [--] & [--] & [--] & [--] & [--] & [--] & [--] & [--] \\
% & $\triangleright$ \textbf{[Variant D]} ([--]) & [--] & [--] & [--] & [--] & [--] & [--] & [--] & [--] & [--] & [--] & [--] & [--] & [--] \\
% & $\triangleright$ \textbf{[Variant E]} ([--]) & [--] & [--] & [--] & [--] & [--] & [--] & [--] & [--] & [--] & [--] & [--] & [--] & [--] \\
% & $\triangleright$ \textbf{[Variant F]} ([--]) & [--] & [--] & [--] & [--] & [--] & [--] & [--] & [--] & [--] & [--] & [--] & [--] & [--] \\
\midrule\midrule
\multirow{7}{*}{\shortstack{\textbf{MetrLA}\\ 2012\\ Speed [MPH] \\(8.11\%)}} 
& STTN (10) 
& $\subperror{5.54}{0.15}$ & $\subperror{6.60}{0.29}$ & $\subperror{7.73}{0.42}$ & $\subperror{6.55}{0.29}$ 
& $\subperror{7.68}{0.33}$ & $\subperror{9.31}{0.59}$ & $\subperror{11.21}{0.96}$ & $\subperror{9.16}{0.60}$ 
& $\subperror{2.86}{0.05}$ & $\subperror{3.27}{0.10}$ & $\subperror{3.75}{0.16}$ & $\subperror{3.23}{0.10}$ & 0.57M \\
& STAEformer (10) & $\subperror{5.58}{0.06}$ & $\subperror{6.62}{0.08}$ & $\subperror{7.55}{0.04}$ & $\subperror{6.51}{0.05}$ & $\subperror{7.60}{0.18}$ & $\subperror{9.08}{0.31}$ & $\subperror{10.55}{0.27}$ & $\subperror{8.88}{0.24}$ & $\subperror{2.85}{0.02}$ & $\subperror{3.21}{0.03}$ & $\subperror{3.58}{0.02}$ & $\subperror{3.16}{0.02}$ & 1.26M \\
& ASTGCN (10) & $\subperror{5.89}{0.10}$ & $\subperror{7.22}{0.11}$ & $\subperror{8.96}{0.16}$ & $\subperror{7.27}{0.11}$ & $\subperror{8.08}{0.20}$ & $\subperror{10.41}{0.14}$ & $\subperror{13.73}{0.37}$ & $\subperror{10.42}{0.16}$ & $\subperror{3.05}{0.05}$ & $\subperror{3.63}{0.07}$ & $\subperror{4.46}{0.12}$ & $\subperror{3.62}{0.08}$ & 0.24M \\
& Graph WaveNet (10) & $\mathbf{5.19}^*_{\mathit{(\pm 0.02)}}$ & $6.22_{\mathit{(\pm 0.04)}}$ & $7.34_{\mathit{(\pm 0.07)}}$ & $6.18_{\mathit{(\pm 0.04)}}$ 
& $\mathbf{7.04}^*_{\mathit{(\pm 0.06)}}$ & $8.61_{\mathit{(\pm 0.12)}}$ & $10.50_{\mathit{(\pm 0.22)}}$ & $8.50_{\mathit{(\pm 0.12)}}$ 
& $2.73_{\mathit{(\pm 0.01)}}$ & $3.13_{\mathit{(\pm 0.02)}}$ & $3.61_{\mathit{(\pm 0.04)}}$ & $3.09_{\mathit{(\pm 0.02)}}$ & 0.30M \\
& \textbf{Weaver} (-) & - & - & - & - & - & - & - & - & - & - & - & - & - \\
& $\triangleright$ \textbf{Default(T)} (10) 
& ${5.27}_{\mathit{(\pm0.02)}}$ & $\mathbf{6.18}^*_{\mathit{(\pm0.03)}}$ & $\mathbf{7.09}^*_{\mathit{(\pm0.02)}}$ & $\mathbf{6.11}^*_{\mathit{(\pm0.02)}}$ 
& $7.10_{\mathit{(\pm0.03)}}$ & $\mathbf{8.37}^*_{\perror{0.04}}$ & $\mathbf{9.87}^*_{\perror{0.04}}$ & $\mathbf{8.25}^*_{\perror{0.03}}$ 
& $\mathbf{2.72}^*_{\perror{0.01}}$ & $\mathbf{3.04}^*_{\perror{0.01}}$ & $\mathbf{3.41}^*_{\perror{0.01}}$ & $\mathbf{3.00}^*_{\perror{0.01}}$ & 0.34M \\
& $\triangleright$ \textbf{Default(NT)} (10) 
& $5.30_{\perror{0.02}}$ & $6.24_{\perror{0.02}}$ & $7.24_{\perror{0.03}}$ & $6.19_{\perror{0.02}}$ 
& $7.08_{\perror{0.05}}$ & $8.39_{\perror{0.06}}$ & $10.00_{\perror{0.07}}$ & $8.28_{\perror{0.05}}$ 
& $2.74_{\perror{0.01}}$ & $3.08_{\perror{0.01}}$ & $3.50_{\perror{0.01}}$ & $3.05_{\perror{0.01}}$ 
& 0.32M \\
% & $\triangleright$ \textbf{[Variant I]} ([--]) & [--] & [--] & [--] & [--] & [--] & [--] & [--] & [--] & [--] & [--] & [--] & [--] & [--] \\
% & $\triangleright$ \textbf{[Variant J]} ([--]) & [--] & [--] & [--] & [--] & [--] & [--] & [--] & [--] & [--] & [--] & [--] & [--] & [--] \\
% & $\triangleright$ \textbf{[Variant K]} ([--]) & [--] & [--] & [--] & [--] & [--] & [--] & [--] & [--] & [--] & [--] & [--] & [--] & [--] \\
% & $\triangleright$ \textbf{[Variant L]} ([--]) & [--] & [--] & [--] & [--] & [--] & [--] & [--] & [--] & [--] & [--] & [--] & [--] & [--] \\
\bottomrule\bottomrule
\end{tabular}
}
\begin{tabular}{p{\textwidth}}
\scriptsize${}^{1}$ One trial excluded due to bad initialization.
\end{tabular}
\label{tab:result}
\end{table}

On \textbf{MetrLA} (high missing data, 8.11\%), Weaver’s advantage becomes more pronounced. While Graph WaveNet remains strong at the 15-minute horizon, Weaver (Default(T)) achieves the best performance at 30 and 60 minutes and on overall metrics. Importantly, Weaver exhibits tight 95\% confidence intervals across all metrics (RMSE: $\sim\pm0.01$--$0.03$, MAPE: $\sim\pm0.03$--$0.07$, MAE: $\sim\pm0.01$), showing high stability even under imperfect and noisy data. Other baselines show substantially higher variance, especially on MetrLA: ASTGCN (RMSE: $\sim\pm0.03$--$0.16$, MAPE: $\sim\pm0.01$--$0.37$, MAE: $\sim\pm0.01$--$0.12$); STTN (RMSE: $\sim\pm0.15$--$0.42$, MAPE: $\sim\pm0.33$--$0.96$, MAE: $\sim\pm0.05$--$0.16$); and even Graph WaveNet experiences increased variability (MAPE: $\sim\pm0.06$--$0.22$ on MetrLA vs. $\sim\pm0.01$--$0.02$ on PemsBay). This contrast suggests that several baselines are sensitive to missing data or initialization, whereas Weaver remains robust without specialized tuning.

The time-metadata variant (T) consistently outperforms the non-time-metadata variant (NT), confirming the utility of incorporating temporal context into Weaver's forecasting process. Even without temporal metadata, Default(NT) remains competitive or superior to baselines at longer horizons, indicating that Weaver effectively learns spatiotemporal structure directly from the network dynamics. Furthermore, Weaver maintains competitive parameter efficiency (0.32--0.34M), comparable to Graph WaveNet (0.30M), despite supporting richer representational capacity via Kronecker Attention, the Traffic Phase Dictionary, and Tanimoto Valence Attention.

\paragraph{Cost-performance analysis.} 
\begin{figure}[width=.99\linewidth,cols=4,pos=t]
    \centering
    \includegraphics[width=0.60\linewidth]{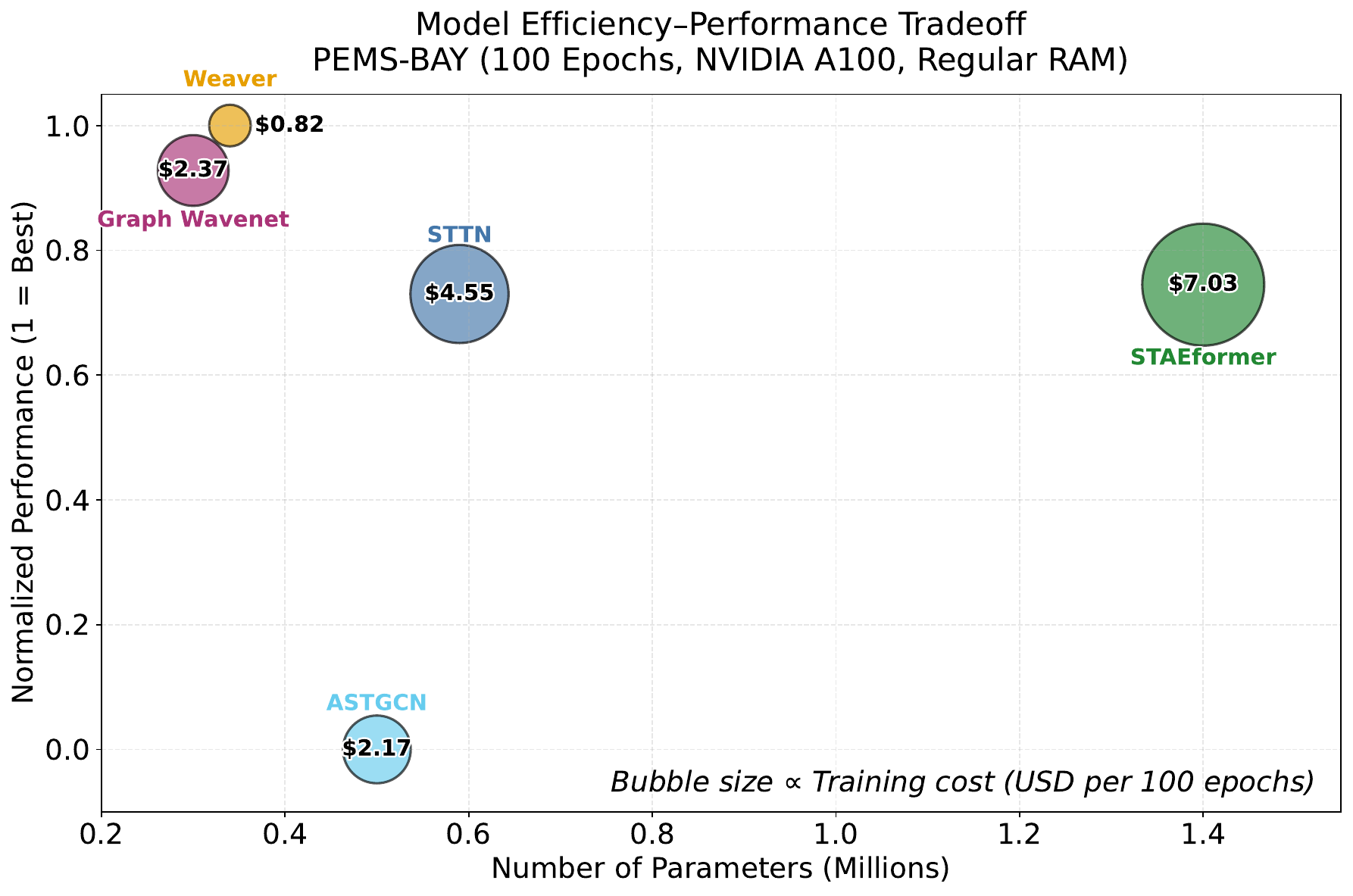}
    \caption{Model efficiency trade-off analysis on PemsBay dataset. The x-axis shows parameters in millions (\textbf{\textit{lower}} is better), y-axis shows normalized performance score where 1 = best, computed as average min-max normalized score across RMSE, MAPE, and MAE (\textbf{\textit{higher}} is better), and bubble areas are proportional to training cost for 100 epochs on an A100 GPU (\textbf{\textit{smaller}} is better). Training costs based on Google Colab pricing as of October 2025 (5.37 compute units per hour, \$49.99 per 500 compute units, equivalent to \$0.537/hour). Weaver achieves the best performance (1.000) with lowest training cost (\$0.82) and compact model size (0.34M parameters).}
    \label{fig:model_efficiency_tradeoff}
\end{figure}

Beyond predictive accuracy, computational efficiency determines whether a model is deployable at scale. Figure~\ref{fig:model_efficiency_tradeoff} shows the cost–performance trade-off on PemsBay, where the bubble area reflects the compute cost of training each model for 100 epochs on an A100 GPU (Google Colab, 5.37 compute units per hour at \$49.99 per 500 compute units). The summary performance score on the y-axis is computed as follows: for each metric $\omega \in \{\text{RMSE}_{\text{All}}, \text{MAPE}_{\text{All}}, \text{MAE}_{\text{All}}\}$ across the `All' horizon, we apply min–max normalization,
\begin{align}
    \omega_{\text{norm}} = \frac{\omega - \omega_{\min}}{\omega_{\max} - \omega_{\min}}\,,
\end{align}
where $\omega_{\min}$ and $\omega_{\max}$ denote the minimum and maximum values across models. To compute 1.0 as the highest score:
\begin{align}
    \text{Normalized Performance} = 1 - \frac{1}{3}\sum_{\omega} \omega_{\text{norm}}\,,
\end{align}

\begin{figure}[width=.99\linewidth,cols=4,pos=t]
    \centering
    % Crop margins: adjust trim={<left> <bottom> <right> <top>}
    \includegraphics[width=0.50\linewidth,trim={10 14 10 7},
    clip]{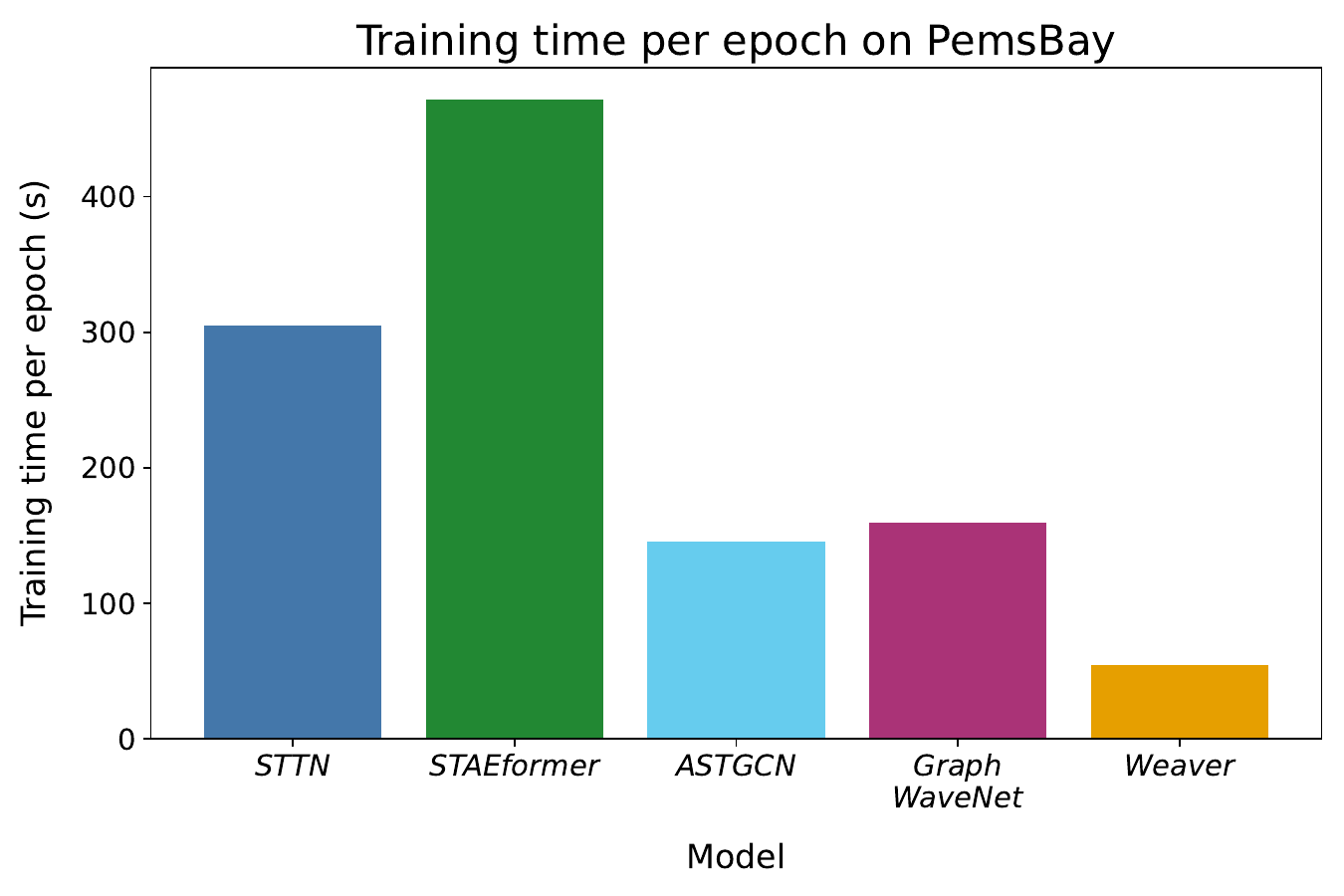}
    \caption{Trial-averaged training times on PemsBay for \textit{Weaver} and baseline models using the A100 GPU (Google Colab), see Figure~\ref{fig:model_efficiency_tradeoff} for training cost conversion for A100 on Google Colab.}
    \label{fig:model-training-times-raw}
\end{figure}

Weaver attains the highest performance score (1.000) while requiring the lowest monetary cost (\$0.82 for 100 epochs). These costs correspond directly to the raw training durations in Figure~\ref{fig:model-training-times-raw}: for 100 training epochs, Weaver requires 1.52 hours, compared to 4.04 hours for ASTGCN, 4.42 hours for Graph WaveNet, 8.48 hours for STTN, and 13.10 hours for STAEformer. Thus, Weaver's lower cost is a direct consequence of shorter per-epoch training duration and reflects reduced computational demand per optimization step.

This efficiency is consistent with the computational complexity reported in Table~\ref{tab:complexity_comps}. Weaver scales as $\mathcal{O}(PN(P+N))$ in time and $\mathcal{O}(P^2 + N^2)$ in space, without a multiplicative layer factor $L$ because spatiotemporal interactions are computed in a single joint operation rather than accumulated across layers. In contrast, STAEformer (\$7.03, 1.40M parameters) and STTN (\$2.37, 0.30M parameters) incur layer-dependent costs—$\mathcal{O}(LPN(P+N))$ and $\mathcal{O}(L(P^2 + N^2))$, respectively—while standard Transformers require $\mathcal{O}(LP^2N^2)$ due to simultaneous spatial–temporal attention. ASTGCN shows lower accuracy yet still incurs non-negligible cost (\$2.17, 0.50M parameters). The alignment between empirical compute cost and theoretical complexity indicates that coupling spatial and temporal dimensions during attention or convolution directly influences training efficiency.

\paragraph{Practical implications and use cases.}
Weaver's competitive prediction performance (Table~\ref{tab:result}), efficient training cost (Figure~\ref{fig:model_efficiency_tradeoff}), and low training times (Figure~\ref{fig:model-training-times-raw}) make it a strong candidate for practical deployment.

Weaver’s stable long–horizon forecasts enable downstream decision-making that depends on anticipating future traffic states. In operational settings, this supports proactive congestion management by allowing transportation agencies (e.g., DOTs, traffic management centers) to adjust signal timing, modify ramp-metering rates, and issue route-choice or departure-time recommendations. The same forecasts can be used for \textit{predictive travel-time analysis} in Advanced Traveler Information Systems (ATIS), providing expected delays and suggested alternatives before congestion forms. 

Weaver also trains significantly faster than baseline models, reducing the cost of retraining and enabling frequent model finetuning as new information becomes available. This is valuable when traffic conditions evolve due to seasonal variation, population shifts, infrastructure changes, or sensor updates. Because the model can be retrained quickly and cheaply, maintaining relevant predictions over time becomes operationally feasible, making Weaver well suited for continuous learning and deployment in evolving traffic systems.

\section{ABLATIONS} \label{section:ablations}
In this section, we perform ablation studies on \textit{Default}(T) settings for the three major \textit{Weaver} components: (i) Kronecker Attention and $\kten{}$ (Section~\ref{section:problem1-kten}), (ii) Tanimoto Valence Attention (Section~\ref{section:problem2-signed-graphs}), and (iii) the Traffic Phase Dictionary (Section~\ref{section:problem3-self-conditioning}).

\subsection{Model Capacity and KPS: Embed Size and Multihead Count}
% Table~\ref{tab:embedsize} -> proposal 1? K-TEN?

\textit{Weaver} models spatial and temporal relationships through $H$ latent graphs generated via dot-product kernels during local spatial- and temporal-attention, then weaves them into a unified spatiotemporal representation using the P$^2$-KMV for global message passing (Section~\ref{section:problem1-kten}). Finally, a weighted implicit Kronecker product summation (W-iKPS, Appendix~\ref{appx:weighted-KPS}) is performed during head mixing (Equation~\eqref{eqn:kmv-head-mixing}). To study this mechanism, we vary the following parameters:
\begin{itemize}
    \item \textbf{Embedding dimension} ($E$), to study its effect on overall representational capacity;
    \item \textbf{Number of heads} ($H$), to study the effects of subgraph partitioning in P$^2$-KMV, which determines the number of summands in the W-iKPS; and
    \item \textbf{Head dimension} ($d_{\text{head}}=E/H$), computed for each $(E,H)$ combination as a more direct measure of per-head representational capacity.
\end{itemize}
Two test sets are designed: (a) Fix $H$ and vary $E$ to study the effects of overall representational capacity, and (b) fix $d_{\text{head}}$ and vary $H$ to study the effect of varying W-iKPS summands while maintaining constant per-head representational capacity. The test metrics are averaged over several trials and summarized in Table~\ref{tab:embedsize}, which also shows that the parameter count is highly dependent on $E$.

\textit{Weaver} demonstrates general robustness and stability across all combinations of $E$ and $H$. Test metrics improve consistently as $E$ and $H$ increase while $95\%$ error margins remain negligible across the board. However, in set (b), both $E=96,H=6$ and $E=64,H=4$ result in elevated error margins (RMSE: $\sim\pm0.01$, MAPE: $\sim\pm0.04$-$0.06$, MAE: $\sim\pm0.01$) compared to $E=32,H=8$ from set (a) despite having significantly higher parameter counts (0.21M and 0.12M versus 0.06M). This suggests that the number of W-iKPS summands (and by extension, latent subgraphs) plays a significant role in determining training stability and convergence consistency in Kronecker Attention, consistent with known properties of Kronecker product summations.

\begin{table}[width=.99\linewidth,cols=4,pos=b]
    \caption{Ablation test results on PemsBay data when \textbf{varying model embed dimension} $E$ and \textbf{number of heads} $H$; Head dim. $d_{\text{head}}=E/H$. `All' indicates metrics across all 15, 30, 45, 60 min horizons. Other hyperparameters follow the Default(T) hyperparameter set. The subscript "D" indicates the defaults. Difference in trials are due to training runs completed when compiling this work.}
    \centering
    \resizebox{\textwidth}{!}{%
        \begin{tabular}{c|c|c|c|c|cccc|cccc|cccc|c}
            \toprule\toprule
            {\makecell{\textbf{Test}\\\textbf{set}}} &
            {\makecell{\textbf{Embed}\\\textbf{dim.}}} & 
            {\makecell{\textbf{Num.}\\\textbf{heads}}} & 
            {\makecell{\textbf{Head}\\\textbf{dim.}}} & 
            {\makecell{\textbf{Num.}\\\textbf{trials}}} &
            \multicolumn{4}{c|}{\textbf{Avg. RMSE}} &
            \multicolumn{4}{c|}{\textbf{Avg. MAPE}} &
            \multicolumn{4}{c|}{\textbf{Avg. MAE}} &
            {\makecell{\textbf{Num.}\\\textbf{Params.}}} \\
            \cmidrule(lr){6-9} \cmidrule(lr){10-13} \cmidrule(lr){14-17}
            & ($E$) & ($H$) & ($d_{\text{head}}$) & & 
            \textbf{15 min} & \textbf{30 min} & \textbf{60 min} & \textbf{All} & 
            \textbf{15 min} & \textbf{30 min} & \textbf{60 min} & \textbf{All} & 
            \textbf{15 min} & \textbf{30 min} & \textbf{60 min} & \textbf{All} & 
             \\
            \midrule\midrule
            - & 128$_D$ & 8$_D$ & 16 & {10} 
                & $\unperror{2.79}{0.01}$ & $\unperror{3.67}{0.01}$ & $\unperror{4.27}{0.01}$ & $\unperror{3.55}{0.01}$ 
                & $\unperror{2.77}{0.00}$ & $\unperror{3.63}{0.01}$ & $\unperror{4.36}{0.02}$ & $\unperror{3.46}{0.01}$ 
                & $\unperror{1.32}{0.00}$ & $\unperror{1.61}{0.00}$ & $\unperror{1.85}{0.00}$ & $\unperror{1.54}{0.00}$ 
                & {0.34M} \\
            \midrule
            \multirow{2}{*}{(a)} & 80 & 8 & 10 & 11 
                & $\unperror{2.82}{0.00}$ & $\unperror{3.70}{0.01}$ & $\unperror{4.30}{0.01}$ & $\unperror{3.60}{0.01}$ 
                & $\unperror{2.79}{0.01}$ & $\unperror{3.68}{0.01}$ & $\unperror{4.45}{0.01}$ & $\unperror{3.51}{0.01}$ 
                & $\unperror{1.33}{0.00}$ & $\unperror{1.63}{0.00}$ & $\unperror{1.89}{0.00}$ & $\unperror{1.56}{0.00}$ 
                & {0.16M} \\
            & 32 & 8 & 4 & 12 
                & $\unperror{2.87}{0.00}$ & $\unperror{3.77}{0.01}$ & $\unperror{4.44}{0.01}$ & $\unperror{3.66}{0.01}$ 
                & $\unperror{2.86}{0.01}$ & $\unperror{3.83}{0.01}$ & $\unperror{4.74}{0.02}$ & $\unperror{3.67}{0.01}$ 
                & $\unperror{1.36}{0.02}$ & $\unperror{1.69}{0.00}$ & $\unperror{1.99}{0.00}$ & $\unperror{1.62}{0.00}$ 
                & {0.06M} \\
            \midrule
            \multirow{2}{*}{(b)} & 96 & 6 & 16 & {10} 
                & $\unperror{2.81}{0.01}$ & $\unperror{3.68}{0.01}$ & $\unperror{4.30}{0.01}$ & $\unperror{3.56}{0.01}$ 
                & $\unperror{2.80}{0.03}$ & $\unperror{3.70}{0.04}$ & $\unperror{4.50}{0.04}$ & $\unperror{3.54}{0.04}$ 
                & $\unperror{1.32}{0.00}$ & $\unperror{1.63}{0.01}$ & $\unperror{1.89}{0.00}$ & $\unperror{1.56}{0.00}$ 
                & {0.21M} \\
            & 64 & 4 & 16 & {10} 
                & $\unperror{2.84}{0.01}$ & $\unperror{3.73}{0.01}$ & $\unperror{4.37}{0.02}$ & $\unperror{3.61}{0.01}$ 
                & $\unperror{2.88}{0.06}$ & $\unperror{3.82}{0.06}$ & $\unperror{4.66}{0.05}$ & $\unperror{3.66}{0.06}$ 
                & $\unperror{1.35}{0.01}$ & $\unperror{1.66}{0.01}$ & $\unperror{1.94}{0.01}$ & $\unperror{1.60}{0.01}$ 
                & {0.12M} \\
            \bottomrule\bottomrule
        \end{tabular}
    }
    \label{tab:embedsize}
\end{table}

\subsection{Valence Attention}
Previous research on traffic network behavior has determined that negative correlations between roads are non-trivial in traffic networks. In Section~\ref{section:problem2-signed-graphs}, we propose Valence Attention, an attention variant that also models negative node relationships, and employ the continuous Tanimoto Coefficient (CTC) due to its ability to represent negative relationships while having characteristics that favor stable model learning dynamics (e.g. self-normalization, local convex/concavity, non-trivial minima/maxima). In this section, we test Tanimoto Valence Attention against typical scaled dot-product attention (SDPA, Equation~\eqref{eqn:multihead_attn}) and the Cosine coefficient (tensor form):
\begin{align}
    \Theta_{\text{cos}} &= \frac{\tensor{Q}\tensor{K}^{\top}}{ \Vert \tensor{Q} \Vert_{\text{vec}} \,\, \Vert \tensor{K}^{\top} \Vert_{\text{vec}} + \epsilon_0 } , \,\epsilon_0 = 10^{-6}
\end{align}
Three test sets are designed with varying embedding dimensions: (a) $E=32$, (b) $E=80$, and (c) $E=128$, to evaluate kernel performance across different model capacities. The averaged test metrics over several trials are summarized in Figure~\ref{fig:valattention-ablation} (see Table~\ref{tab:valattention-ablation} in Appendix~\ref{appx:ablations} for full values).

Tanimoto consistently outperforms both SDPA and Cosine across all embedding dimensions and metrics. The performance gap is most pronounced at lower capacities, with Cosine showing severe degradation in set (a) (RMSE: $3.99\pm0.06$, MAPE: $4.64\pm0.12$, MAE: $1.84\pm0.04$) and elevated error margins ($\sim\pm0.12$-$0.15$), while Tanimoto maintains stable test metrics and low error margins even at $E=32$. SDPA shows moderate performance between Tanimoto and Cosine, but exhibits consistently higher test metrics across all horizons. As embedding dimension increases in sets (b) and (c), the gap narrows but Tanimoto retains its advantage with consistently lower test metrics and error margins across all configurations. The stability of Tanimoto is particularly evident in its consistently low error margins ($\sim\pm0.01$-$0.02$) across all embedding dimensions, while Cosine's error margins remain elevated even at higher capacities. 

In conclusion, these results demonstrate that Tanimoto's inherent properties (self-normalization and bounded output range) provide both improved accuracy and training stability, particularly in capacity-constrained settings where modeling negative relationships becomes critical for capturing complex traffic dynamics. Interestingly, we observe that at $E = 128$, Cosine marginally outperforms SDPA, which suggests that capturing negative relationships is inherently valuable for traffic forecasting.

\begin{figure}[width=.99\linewidth,cols=4,pos=t]
    \centering
    \includegraphics[width=0.85\linewidth]{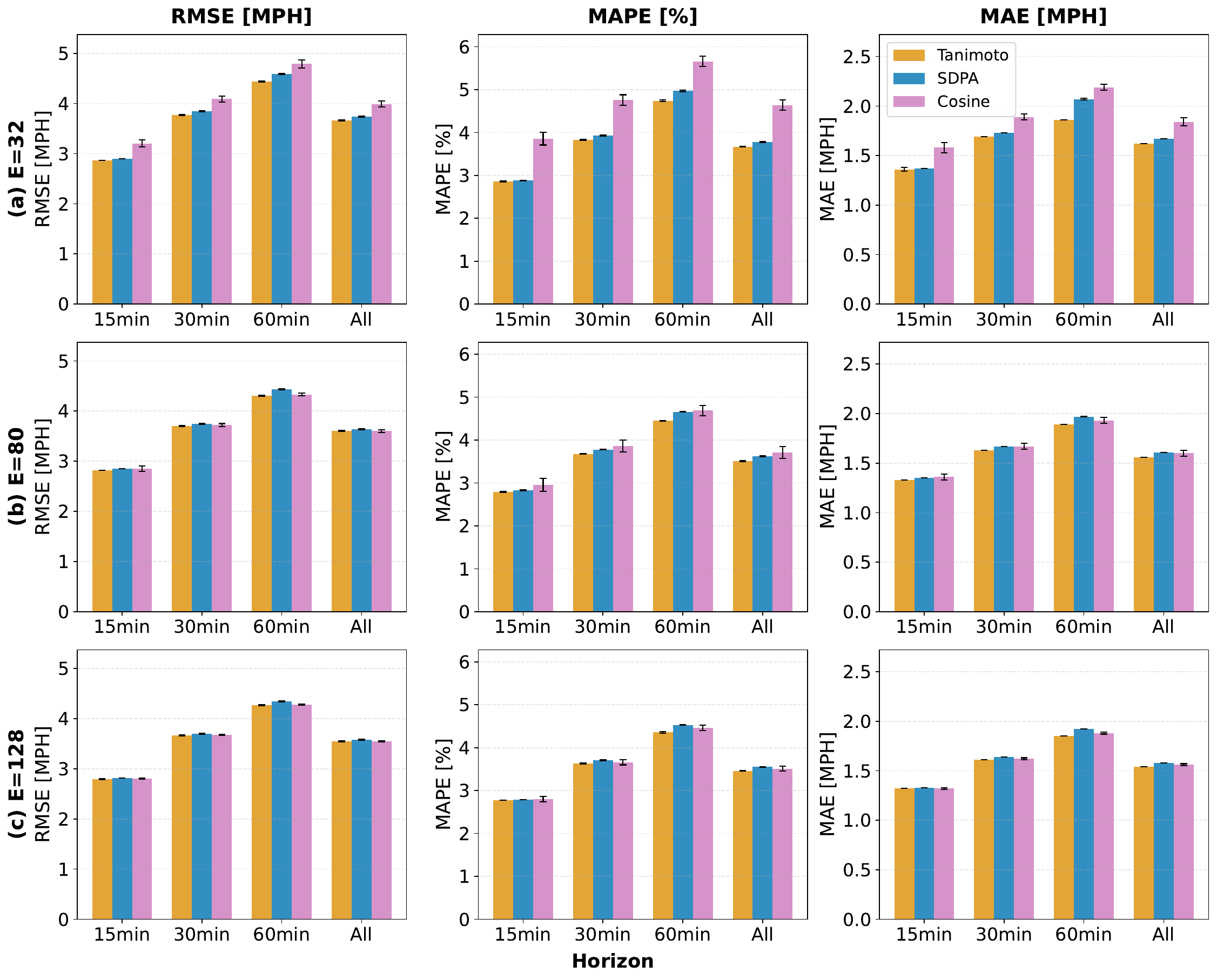}
    \caption{Comparisons of Tanimoto Valence Attention (Tanimoto) versus scaled dot-product attention (SDPA, using softmax) and Cosine coefficient (Cosine) on PemsBay data using the Default(T) hyperparameter set. Detailed results are available in Table~\ref{tab:valattention-ablation} in Appendix~\ref{appx:ablations}. Lower metrics value is better.}
    \label{fig:valattention-ablation}
\end{figure}

\subsection{Traffic Phase Dictionary}
The Traffic Phase Dictionary (Section~\ref{section:problem3-self-conditioning}) enables Weaver to learn latent representations of traffic phases (e.g., free flow, congestion), generating \textit{latent cofactors} $\boldsymbol{\Xi}\in \mathbb{R}^{P\times N\times K_{\Xi}}$ based on the input traffic measures at each node (node-state) $\midtilde{\tensor{X}}\in \mathbb{R}^{N\times PC}$. 
The dictionary size $(M_{\Xi})$ determines the number of landmarks in the dictionary manifold, whereas the cofactor dimension $(K_{\Xi})$ determines the representational capacity of each latent cofactor. 
Three test sets are designed: (a) vary $M_{\Xi}$ and fix $K_{\Xi}$, (b) vary $K_{\Xi}$ and fix $M_{\Xi}$, and (c) vary both $M_{\Xi}$ and $K_{\Xi}$ maintaining a 2:1 ratio. 
The averaged test metrics over several trials are summarized in Table~\ref{tab:ablation_traffic-phase-dictionary}. 
 
Across all test sets, Weaver demonstrates robustness to variations in Traffic Phase Dictionary hyperparameters. In set (a), reducing dictionary size from 64 to 4 results in elevated error margins (MAPE: $\sim\pm0.08$-$0.10$) compared to higher counts, suggesting that insufficient dictionary landmarks degrade training stability. However, test metrics remain comparable across $M_{\Xi}=8$, 16, and 64, indicating diminishing returns beyond a minimum threshold. In set (b), varying cofactor dimension from 4 to 32 shows minimal impact on both test metrics and error margins, suggesting that $K_{\Xi}$ has limited influence on model performance within this range. Set (c) demonstrates that scaling both parameters proportionally yields consistent improvements, with $M_{\Xi}=128, K_{\Xi}=64$ achieving the best performance across all metrics while maintaining low error margins. These results suggest that dictionary size plays a more critical role than cofactor dimension in capturing diverse traffic patterns, though both contribute to overall model capacity.

\begin{table}[width=.99\linewidth,cols=4,pos=b]
    \caption{Ablation test results on PemsBay data when \textbf{varying Traffic Phase Dictionary hyperparameters}. Other hyperparameters follow the Default(T) hyperparameter set. The subscript "D" indicates the defaults. Difference in trials are due to training runs completed when compiling this work.}
    \centering
    \resizebox{\textwidth}{!}{%
        \begin{tabular}{c|c|c|c|cccc|cccc|cccc|c}
            \toprule\toprule
            {\makecell{\textbf{Test}\\\textbf{set}}} &
            {\makecell{\textbf{Landmark}\\\textbf{count}}} & 
            {\makecell{\textbf{Cofactor}\\\textbf{width}}} & 
            {\makecell{\textbf{Num.}\\\textbf{trials}}} &
            \multicolumn{4}{c|}{\textbf{Avg. RMSE}} &
            \multicolumn{4}{c|}{\textbf{Avg. MAPE}} &
            \multicolumn{4}{c|}{\textbf{Avg. MAE}} &
            {\makecell{\textbf{Num.}\\\textbf{Params.}}} \\
            \cmidrule(lr){5-8} \cmidrule(lr){9-12} \cmidrule(lr){13-16}
            & ($M_{\Xi}$) & ($K_{\Xi}$) & & 
            \textbf{15 min} & \textbf{30 min} & \textbf{60 min} & \textbf{All} &
            \textbf{15 min} & \textbf{30 min} & \textbf{60 min} & \textbf{All} &
            \textbf{15 min} & \textbf{30 min} & \textbf{60 min} & \textbf{All} &
            \\
            \midrule\midrule
            - & {$64_D$}  & {$32_D$} & {$10$} 
                & $\unperror{2.79}{0.01}$ & $\unperror{3.67}{0.01}$ & $\unperror{4.27}{0.01}$ & $\unperror{3.55}{0.01}$
                & $\unperror{2.77}{0.00}$ & $\unperror{3.63}{0.01}$ & $\unperror{4.36}{0.02}$ & $\unperror{3.46}{0.01}$
                & $\unperror{1.32}{0.00}$ & $\unperror{1.61}{0.00}$ & $\unperror{1.85}{0.00}$ & $\unperror{1.54}{0.00}$ 
                & {0.34M} \\
            \midrule
            \multirow{3}{*}{(a)} & {$4$}  & {$32$} & {10} 
                & $\unperror{2.86}{0.03}$ & $\unperror{3.72}{0.02}$ & $\unperror{4.32}{0.02}$ & $\unperror{3.60}{0.02}$
                & $\unperror{3.07}{0.10}$ & $\unperror{3.89}{0.09}$ & $\unperror{4.62}{0.08}$ & $\unperror{3.74}{0.09}$
                & $\unperror{1.37}{0.02}$ & $\unperror{1.67}{0.02}$ & $\unperror{1.91}{0.01}$ & $\unperror{1.60}{0.02}$
                & {0.32M} \\
            & {$8$}  & {$32$} & {11} 
                & $\unperror{2.82}{0.01}$ & $\unperror{3.70}{0.01}$ & $\unperror{4.30}{0.01}$ & $\unperror{3.57}{0.01}$
                & $\unperror{2.79}{0.01}$ & $\unperror{3.65}{0.01}$ & $\unperror{4.39}{0.02}$ & $\unperror{3.48}{0.01}$
                & $\unperror{1.33}{0.00}$ & $\unperror{1.62}{0.00}$ & $\unperror{1.87}{0.00}$ & $\unperror{1.56}{0.00}$
                & {0.32M} \\
            & {$16$}  & {$32$} & {10} 
                & $\unperror{2.81}{0.00}$ & $\unperror{3.68}{0.01}$ & $\unperror{4.28}{0.01}$ & $\unperror{3.56}{0.01}$
                & $\unperror{2.77}{0.01}$ & $\unperror{3.63}{0.01}$ & $\unperror{4.37}{0.01}$ & $\unperror{3.47}{0.01}$
                & $\unperror{1.32}{0.00}$ & $\unperror{1.62}{0.00}$ & $\unperror{1.86}{0.00}$ & $\unperror{1.55}{0.00}$
                & {0.32M} \\
            \midrule
            \multirow{3}{*}{(b)} & {$64$}  & {$4$} & {10} 
                & $\unperror{2.82}{0.01}$ & $\unperror{3.70}{0.01}$ & $\unperror{4.30}{0.02}$ & $\unperror{3.57}{0.01}$
                & $\unperror{2.83}{0.03}$ & $\unperror{3.70}{0.04}$ & $\unperror{4.46}{0.05}$ & $\unperror{3.53}{0.04}$
                & $\unperror{1.33}{0.01}$ & $\unperror{1.63}{0.01}$ & $\unperror{1.88}{0.01}$ & $\unperror{1.56}{0.01}$
                & {0.32M} \\
            & {$64$}  & {$8$} & {$10$} 
                & $\unperror{2.81}{0.00}$ & $\unperror{3.68}{0.00}$ & $\unperror{4.28}{0.01}$ & $\unperror{3.56}{0.00}$
                & $\unperror{2.78}{0.01}$ & $\unperror{3.65}{0.01}$ & $\unperror{4.39}{0.01}$ & $\unperror{3.48}{0.01}$
                & $\unperror{1.32}{0.00}$ & $\unperror{1.62}{0.00}$ & $\unperror{1.86}{0.00}$ & $\unperror{1.55}{0.00}$ 
                & {0.32M} \\
            & {$64$}  & {$16$} & {$10$} 
                & $\unperror{2.80}{0.00}$ & $\unperror{3.68}{0.01}$ & $\unperror{4.28}{0.01}$ & $\unperror{3.55}{0.01}$
                & $\unperror{2.77}{0.01}$ & $\unperror{3.64}{0.02}$ & $\unperror{4.39}{0.02}$ & $\unperror{3.47}{0.02}$
                & $\unperror{1.32}{0.00}$ & $\unperror{1.62}{0.00}$ & $\unperror{1.86}{0.00}$ & $\unperror{1.55}{0.00}$ 
                & {0.33M} \\
            \midrule
            \multirow{2}{*}{(c)} & {$8$}  & {$4$} & {$10$} 
                & $\unperror{2.84}{0.03}$ & $\unperror{3.72}{0.02}$ & $\unperror{4.32}{0.02}$ & $\unperror{3.59}{0.02}$
                & $\unperror{2.87}{0.12}$ & $\unperror{3.71}{0.10}$ & $\unperror{4.46}{0.07}$ & $\unperror{3.55}{0.10}$
                & $\unperror{1.35}{0.02}$ & $\unperror{1.64}{0.01}$ & $\unperror{1.89}{0.01}$ & $\unperror{1.58}{0.02}$ 
                & {0.31M} \\
            & {$128$}  & {$64$} & {$10$} 
                & $\unperror{2.79}{0.00}$ & $\unperror{3.66}{0.01}$ & $\unperror{4.26}{0.01}$ & $\unperror{3.54}{0.01}$
                & $\unperror{2.75}{0.01}$ & $\unperror{3.61}{0.01}$ & $\unperror{4.35}{0.02}$ & $\unperror{3.44}{0.01}$
                & $\unperror{1.31}{0.00}$ & $\unperror{1.61}{0.00}$ & $\unperror{1.85}{0.00}$ & $\unperror{1.54}{0.00}$ 
                & {0.42M} \\
            \bottomrule\bottomrule
        \end{tabular}
    }%
    \label{tab:ablation_traffic-phase-dictionary}
\end{table}

\clearpage
\section{CONCLUSION}
We introduced Weaver, a novel spatiotemporal forecasting architecture that addresses fundamental challenges in modeling complex traffic networks through three key contributions: (i) Kronecker Attention for efficient global message passing via P$^2$-KMV and weighted implicit Kronecker product summation (W-iKPS), (ii) Valence Attention using the continuous Tanimoto coefficient (CTC) to model signed relationships in traffic networks, and (iii) the Traffic Phase Dictionary for learning latent traffic phase representations.

Our experimental results demonstrate that Weaver achieves competitive performance against contemporary baselines while exhibiting superior training stability across datasets with varying data quality. On PemsBay and MetrLA benchmarks, Weaver consistently outperforms existing methods at longer prediction horizons (30 and 60 minutes) and maintains low error margins even under challenging conditions with substantial missing data. Notably, Weaver achieves competitive performance even without temporal metadata (NT variants), demonstrating that it learns underlying physical principles rather than relying on cues from temporal metadata alone. Ablation studies confirm the robustness of Weaver's components across varying hyperparameters, with the number of attention heads playing critical roles in training stability through adjusting Kronecker product summands. Furthermore, Weaver demonstrates robustness to parameter count variations, making it adaptable to different computational resource constraints and deployment scenarios.

The success of Weaver's Kronecker-based formulation suggests promising directions for future spatiotemporal modeling in transportation systems. The Kronecker tensor product structure naturally captures system-wide bulk behavior, which may explain Weaver's strength in long-term forecasting. Weaver's architectural organization, inspired by Interaction Networks (IN), is designed to reflect traditional physical modeling principles, making it naturally accommodating for physics-informed neural network approaches in future work. Additionally, Weaver is designed to flexibly accommodate various spatial and temporal metadata depending on availability, enhancing its applicability across diverse data scenarios. Furthermore, its adherence to Geometric Deep Learning (GDL) principles provides a natural foundation for theoretical analysis and extension, offering a grounded mathematical basis that makes the model's behavior more predictable and interpretable while complementing the empirical findings from our ablation studies.

\subsection{Future Work} 

\paragraph{Architectural extensions.} Future work could explore extending Kronecker Attention to higher-order tensor decompositions and investigating adaptive mechanisms for the Traffic Phase Dictionary's manifold structure, including fine-tuning strategies for transfer learning, domain adaptation across different traffic networks, and continual learning for online traffic pattern evolution. Another direction is to incorporate physical road geometry (e.g., lane topology, curvature, and ramp proximity) into the retrieval mechanism as (a fine-tuning step) of the Traffic Phase Dictionary to encode physical correspondence between nodes, keeping the trained dictionary frozen. This would enable geometry-aware retrieval without altering the underlying learned dictionary structure, allowing the model to disentangle behavioral dynamics from static infrastructure.

\paragraph{Baseline and transferability.} The present design deliberately omits physical node attributes to establish baseline modeling characteristics and to demonstrate that the model can learn fundamental traffic dynamics without preemptive structural bias. Future work may therefore investigate fine-tuning only the retrieval layer—while keeping the learned dictionary fixed—to incorporate physical geometry information. This would test whether the a priori trained dictionary serves as a transferable basis of systemic traffic behavior across networks.

\paragraph{Domain-specific applications.} Investigating how spatial and temporal metadata can be leveraged to inductively bias the formation of latent graphs for specific use cases could further enhance Weaver's performance in domain-specific applications. Weaver could also be extended to related transportation forecasting domains including public transit operations, autonomous vehicle coordination, and air traffic management.

\paragraph{Theoretical analysis.} Additionally, theoretical analysis could focus on the convergence properties of W-iKPS, the convergent behavior and stability of CTC, and the topological properties of the induced latent graphs to provide deeper insights into the observed training stability and guide further architectural improvements.

\clearpage

\bibliographystyle{apalike} %dcu
%\nocite{*}
\bibliography{reference}

\clearpage

\appendix

\section*{Appendix}
\section{Notation and terminology}\label{appx:notations}
In this section, we summarize the notation and terminology for tensors and neural networks used in this work. For general tensor notation, see~\citet{TensorOps:Kolda2009}. 

\subsection{Common Notations}

\textbf{Vectors} are denoted by bold lowercase letters, e.g.\ $\asvector{v}$, and are expressed in \textit{column-major} form:
\begin{align}
\asvector{v} \in \mathbb{R}^{D_v \times 1}.
\end{align}
\textbf{Sets} are denoted by calligraphic uppercase letters, e.g.\ $\mathcal{S}$, unless stated otherwise.  

% The \textbf{compound subscript} notation $\spaceset/\timesset$ is used when spatial ($\spaceset$) and temporal ($\timesset$) aspects share algebraically equivalent computations, for example:
% \begin{align}
% f_{\spaceset/\timesset}(\,\cdot\,).
% \end{align}

For brevity, we omit the leading \textbf{sample-batch mode}\footnote{
A tensor \textit{mode} is colloquially referred to as a “tensor dimension.” 
Here we adopt the term \textit{mode} to emphasize operations along specific axes (cf.~\citet{TensorOps:Kolda2009}), 
and to avoid confusion with the \textit{feature/channel dimension} often used in machine learning to denote embedding width or feature count.
}—denoted $B$—that is common to neural networks trained with Stochastic Gradient Descent (SGD) methods~\citep{SGD:RazzoukiEtAl2024}.  
Thus, a tensor with shape $\mathbb{R}^{B \times P \times N \times C}$ is written more compactly as
\begin{align}
\mathbb{R}^{P \times N \times C}.
\end{align}

\subsection{Matrix operations} \label{appx:matrix-operations}
\textbf{column-wise picture of matrix multiplication (CMM)}:
\begin{align}
\asmatrix{A}\,\asmatrix{B} 
= \begin{bmatrix}
\asmatrix{A}\,\mathbf{b}_1 & \asmatrix{A}\,\mathbf{b}_2 & \cdots & \asmatrix{A}\,\mathbf{b}_C
\end{bmatrix}\, , \label{eqn:column-picture-matmul}
\end{align}
Where $\asmatrix{A}\in \mathbb{R}^{A \times B}$ and $\asmatrix{B}\in \mathbb{R}^{B \times C}$, and $\asvector{b}_i \in \mathbb{R}^{B \times 1},\,i\in [C]$ are column vectors of $\asmatrix{B}$.

\subsection{Tensor operations} \label{appx:tensor-operations}

For the tensor $\tensor{A}$:
\begin{align}
\tensor{A} \in \mathbb{R}^{I_1 \times I_2 \times \cdots \times I_m \times \cdots \times I_M}\,, \label{eqn:definition-tensor-mode}
\end{align}
the \textbf{mode $m$} indexes specific axes of the tensor, with $m \in \{1,2,\ldots,M\}$, and $I_m$ is the size of the mode (i.e., the tensor dimension).

The \textbf{concatenation} operation is denoted simply by single "$\Vert$", and serial concatenations with "$\Vert_{i=1}^n$", where all concatenations are performed along the last dimension unless stated. Otherwise, we generally distinguish \textbf{tensor-level operations} by $\Langle\,\cdot\,\Rangle$. For instance, \textbf{concatenation along mode-$m$} is stated as: $\Langle \tensor{A} \Vert \tensor{B} \Rangle_{\text{mode-}m}$ for $\tensor{A} \in \mathbb{R}^{I_{A,1} \times \ldots \times I_{A,m} \times \ldots I_{A,D}}, \tensor{B} \in \mathbb{R}^{I_{B,1} \times \ldots \times I_{B,m} \times \ldots I_{B,D}}$ where necessarily $I_{A,i}=I_{B,i}$ for $i\neq m$. Hence:
\begin{align}
    \Langle \tensor{A} \Vert \tensor{B} \Rangle_{\text{mode-}m} &= \tensor{C}\,\in\,\mathbb{R}^{I_{A,1} \times \ldots \times ( I_{A,m} + I_{B,m} ) \times \ldots \times I_{A,D} }\,. \label{eqn:definition-tensor-concatenation}
\end{align}

\textbf{Tensor mode rearrangement} (tensor rearrangement) covers permutation and flattening operations. They are denoted using:
\begin{align}
    \tensor{B} &= \LLangle \tensor{A}_{[ \mathcal{M}_{\text{original}}   \rightarrow   \mathcal{M}_{\text{target}} ]}\, \RRangle, \label{eqn:definition-tensor-rearrangement}
\end{align}
where $\mathcal{M}_{\text{original}}$ denotes the ordered list of tensor modes in the input, and $\mathcal{M}_{\text{target}}$ denotes the target mode configuration in the output.

\textbf{Tensor mode summation} along a specific tensor mode $m$ is denoted by $\sum_{i=1}^{I_m}\Langle\,\cdot\,\Rangle_m: \mathbb{R}^{I_1 \times I_2 \times \ldots \times I_m \times \ldots \times I_{D-1} \times I_{D}}$, performing element-wise matrix summation along mode $m$:
\begin{align}
\sum_{i=1}^{I_m} \LLangle \tensor{X} \RRangle_{\text{mode-m}}
    &= \sum_{i_m=1}^{I_m} \tensor{X}_{[:\,,\,\ldots\,,\,i_m\,,\,:,\,\ldots\,,\,:]} \,, \label{eqn:definition-tensor-mode-summation}
\end{align}
where the subscript $[:,\ldots,i_m,\,:,\ldots\,,:\,]$ denotes \textbf{slicing}\footnote{Refer to \citet{TensorOps:Kolda2009} on slicing.} along mode $m$ at index $i_m \in [I_m]$.

\textbf{Tensor mode variance} computes the biased variance of a tensor along mode~$m$ using tensor mode summation:
\begin{equation}
    \mathrm{Var}_{\text{mode-}m}(\tensor{X})
    = \frac{1}{I_m} \sum_{i=1}^{I_m}
      \LLangle (\tensor{X} - \bar{\tensor{X}})^2 \RRangle_{\text{mode-}m}
    \quad \in \mathbb{R}^{I_1 \times \cdots \times I_{m-1} \times I_{m+1} \times \cdots \times I_M},
    \label{eqn:tensor-mode-variance}
\end{equation}
where $\tensor{X} \in \mathbb{R}^{I_1 \times \cdots \times I_M}$, 
$\bar{\tensor{X}} \in \mathbb{R}^{I_1 \times \cdots \times 1 \times \cdots \times I_M}$ 
is the mean of $\tensor{X}$ along mode~$m$ (with its $m$-th dimension collapsed to~1), and 
the squared difference $(\tensor{X} - \bar{\tensor{X}})^2$ is applied element-wise.

The \textbf{top-$k$ selection} operation extracts the $k$ largest elements along a specified mode based on a scoring tensor. 
Let $\tensor{X} \in \mathbb{R}^{I_1 \times I_2 \times \cdots \times I_D}$ be the \textbf{source tensor} and 
$\boldsymbol{\Gamma} \in \mathbb{R}^{I_1 \times I_2 \times \cdots \times I_D}$ its corresponding \textbf{score tensor}. 
For mode $m$, the top-$k$ selection induced by $\boldsymbol{\Gamma}$ is defined as:
\begin{equation}
    \tensor{X}^{\text{Top-}k(I_m, \boldsymbol{\Gamma})} 
    = \tensor{X}_{[:,\,\ldots,\,i_{m,1},\,\ldots,\,i_{m,k},\,\ldots,\,:]} 
    \quad \in \mathbb{R}^{I_1 \times \cdots \times k \times \cdots \times I_D}, 
    \label{eqn:top-k-selection}
\end{equation}
where $\{i_{m,1}, \ldots, i_{m,k}\}$ are the indices of the $k$ largest entries of $\boldsymbol{\Gamma}$ along mode $m$, 
sorted in descending order by their score values. 
The same index set can be used to select $\boldsymbol{\Gamma}$ itself, i.e., 
$\boldsymbol{\Gamma}^{\text{Top-}k(I_m)} = \boldsymbol{\Gamma}_{[:,\,\ldots,\,i_{m,1},\,\ldots,\,i_{m,k},\,\ldots,\,:]}$. 
This operation preserves the tensor structure while reducing the dimension $I_m$ to $k$.

\textbf{Tensor broadcasting} is a mechanism in deep learning frameworks where dimensions of size $1$ are implicitly expanded (replicated) to match the size of corresponding dimensions in other tensors during operations. We denote \textbf{broadcasted modes} with the ``!'' superscript, with the target dimension size shown in the subscript. For example, $\tensor{A}^! \in \mathbb{R}^{I_1 \times \cdots \times 1_{I_i}^! \times \cdots \times I_D \times M_1 \times C}$ indicates that the broadcasted mode $1_{I_i}^!$ will be expanded to size $I_i$ during computation. Broadcasting alters both the interpretation of batch structure and the effective memory/computation cost~\citep{NVIDIA2023CUDA}.

\textbf{Batch matrix multiplication} (BMM) is a common tensor operation in contemporary deep learning. For tensors $\tensor{A} \in \mathbb{R}^{I_1 \times \cdots \times I_D \times M_1 \times C}$ and $\tensor{B} \in \mathbb{R}^{I_1 \times \cdots \times I_D \times M_2 \times C}$, the BMM operation yields $\tensor{Y} \in \mathbb{R}^{I_1 \times \cdots \times I_D \times M_1 \times M_2}$:
\begin{align}
Y_{i_1,\ldots,\,i_D,\ell,\,j}
= \sum_{k=1}^{C} A_{i_1,\ldots,\,i_D,\ell,k}\, B_{i_1,\ldots,\,i_D,\,j,k}\, . \label{eqn:definition-BMM-basic}
\end{align}
We denote this compactly as $\tensor{Y}=\tensor{A}\,\tensor{B}^\top$, where multiplications occur over the last index $k$, and the batch modes $I_1,\ldots,I_D$ \textbf{stratify} sets of independent operations. This follows the convention of \texttt{torch.matmul} in PyTorch, with the special case $D=1$ corresponding to \texttt{torch.bmm}.
Importantly, tensor broadcasting affects BMM behavior, which requires additional consideration for the intended computational outcomes:
\begin{itemize}
    \item \textbf{BMM with broadcasting:} Flattening the batch dimensions would collapse replicated operations into a single index, obscuring the stratification structure,
    \item \textbf{BMM without broadcasting:} Batch dimensions can be flattened without loss of generality:
    \begin{align}
        \tensor{A}\,\tensor{B}^\top \equiv 
        \midtilde{\tensor{A}}\, \midtilde{\tensor{B}}^\top,\quad \text{where }\,
        \midtilde{\tensor{A}} \in \mathbb{R}^{(I_1\cdots I_D)\times M_1 \times C},\,\, \midtilde{\tensor{B}} \in \mathbb{R}^{(I_1\cdots I_D)\times M_2 \times C} \,.
    \end{align}
\end{itemize}

\subsection{Neural network components.}  \label{appx:neural-network-components}

Generally, \textbf{learnable weights and biases} are respectively denoted by $\asmatrix{W}$ and $\asvector{b}$, with \textbf{multiplication of tensors} $\tensor{X}$ \textbf{and matrices} $\asmatrix{W}$ occur on the last two dimensions, i.e.:
\begin{align}
    \tensor{X}\in\mathbb{R}^{\ldots\times I_{D-1}\times I_{D}}, \quad 
    \asmatrix{W}\in\mathbb{R}^{I_{D}\times E}
    \Rightarrow
    \tensor{X}\asmatrix{W}\in\mathbb{R}^{\ldots\times I_{D-1}\times E},
\end{align}
where "$\ldots$" are placeholders for prior tensor dimensions. The \textbf{addition of} $\tensor{X}$ \textbf{and vectors} $\asvector{b}$ are applied to the last mode:
\begin{align}
    \asvector{b}\in\mathbb{R}^{E}
    \Rightarrow
    \tensor{X}\asmatrix{W}+\asvector{b}\in\mathbb{R}^{\ldots\times I_{D-1 }\times E}.
\end{align}
Additionally, \textbf{\textit{ResNet}-style residual connections} \citep{ResNet:he2016residual} are applied to improve model training dynamics by reformulating learned functions in residual form:
\begin{align}
    h(\tensor{X}) &= (\tensor{X})_{\text{res}} + \Delta h(\tensor{X}), \\
    \Delta h(\tensor{X}) &= F(\tensor{X};\,\mathcal{W},\mathcal{B}),
\end{align}
where the subscript “res” denotes the residual (identity) connection, i.e., $(\tensor{X})_{\text{res}}=\tensor{X}$, and $F(\,\cdot\,;\mathcal{W},\mathcal{B})$ denotes an arbitrary MLP-type function parameterized by weights $\asmatrix{W}_{\ell}\in\mathcal{W}$ and biases $\asvector{b}_{\ell}\in\mathcal{B}$, for layers $\ell\in\{1,2,\ldots, L\}$.

\textbf{Gated Linear Units} (GLU) \citep{GLU:pmlr-v70-dauphin17a} are utilized for its simplicity, compactness, and expressiveness, which allows us to focus on core architectural innovations by standardizing hidden dimensions as $2\times$ the target output dimension. However, standard MLP implementations (e.g. using ReLU, GELU, etc.) remain viable alternatives, and it is still used in output processing. GLUs are self-gating mechanisms where for $\tensor{U} \in \mathbb{R}^{\,\ldots\times \,2 D_{\text{out}}}$:
\begin{alignat}{2}
    \tensor{U}_{\text{signal}} &= \tensor{U}_{[\,\ldots\,,\,0:D_{\text{out}}]} \, &&\in\,\mathbb{R}^{\ldots \times \, D_{\text{out}} }\,, \\
    \tensor{U}_{\text{gate}}   &= \tensor{U}_{[\,\ldots\,,\,E:2D_{\text{out}}]} \,&&\in\,\mathbb{R}^{\ldots \times \, D_{\text{out}} }\,, \\
    \glu(\tensor{U}) &= \tensor{U}_{\text{signal}} \odot \sigma(\tensor{U}_{\text{gate}}).
\end{alignat}
Here, the subscript $[\ldots,\, I_a : I_b]$, for $I_a < I_b$ and $I_a, I_b \in [D_{\text{out}}]$, indicates slicing along the corresponding dimensions, $\odot$ denotes element-wise multiplication and $\sigma(\cdot)$ is the logistic sigmoid function. Overall, this yields the mapping:
\begin{align}
    \glu : \mathbb{R}^{\,\ldots\times 2E} \to \mathbb{R}^{\,\ldots\times E}.
\end{align}

\section{Benchmark Models Setup} \label{appx:benchmarks-setup}
In this section, we list the model hyperparameters and PyTorch setup used to train each benchmark model in Section~\ref{sec:experimental-setup}. For compatibility with the Torch Spatiotemporal (TSL)~\citep{TSL:Cini_Torch_Spatiotemporal_2022} training interface, we use code from existing repositories and apply a minimally invasive wrapper class (see Listing~\ref{lst:model_wrapper}) for input/output and hyperparameter handling. The summarized benchmark configurations are provided in Table~\ref{tab:appendix_model_manifest}.

\begin{lstlisting}[language=Python,
    caption={Example of generic model wrapper for TSL training interface.},
    label={lst:model_wrapper}]
class ModelWrapper(nn.Module):
    def __init__(self, model_class, model_params, static_exogenous=None):
        ...
        self.model = model_class(**model_params, **(static_exogenous or {}))

    def forward(self, x, metadata=None):
        if metadata is not None:
            x = torch.cat([x, metadata], dim=-1)
        return self.model(x)
\end{lstlisting}

\begin{table}[width=.99\linewidth,cols=4,pos=b]
\centering
\caption{Model-specific structural hyperparameters, associated papers, and source code references. Implementations were used for compatibility with the TSL framework.}
\label{tab:appendix_model_manifest}
\resizebox{\linewidth}{!}{
\begin{tabular}{l p{0.44\linewidth} p{0.20\linewidth} p{0.25\linewidth}}
\toprule
\textbf{Model} & \textbf{Hyperparameters / Metadata} & \textbf{Paper} & \textbf{Source Code} \\
\midrule

STTN &
input channels, $C_{\text{in}}=1$; embedding dimension, $d_{\text{embed}}=64$; layers, $L=3$; Chebyshev order, $K=3$; heads, $H=1$; forward expansion, $r=4$; dropout, 0.001. &
\citet{STTN:xu2021spatial}. &
\href{https://github.com/wubin5/STTN/tree/main/Batch_Training_Version}{STTN (PyTorch)\textsuperscript{$\dagger$}}. \\

\addlinespace[2pt]

STAEformer &
feature embedding dimension, $d_f=24$; adaptive embedding dimension, $d_a=80$; layers, $L=3$; heads, $H=4$; dropout, 0.1; temporal metadata: TOD (indexed, 0–287), DOW (one-hot). &
\citet{STAE:liu2023staeformer}. &
\citet{STAE:liu2023staeformer}. \\

\addlinespace[2pt]

ASTGCN &
spatial-temporal blocks, $B=2$; input channels, $C_{\text{in}}=3$; Chebyshev order, $K=3$; graph convolution kernels, 64; temporal convolution kernels, 64; temporal stride, 1. &
\citet{ASTGCN:guo2019astgcn}. &
\citet{geometric_temporal:rozem2021}. \\

\addlinespace[2pt]

Graph WaveNet &
input channels, $C_{\text{in}}=2$; hidden dimension, $d_{\text{hid}}=32$; layers, $L=8$; diffusion step, $K=2$; node-embedding dimension, $d_{\text{emb}}=10$; static graph adjacency; dropout, 0.3. &
\citet{GraphWavenet:wu2019}. &
\citet{GraphWavenet:wu2019}. \\

\bottomrule
\end{tabular}
}
\vspace{2pt}
\raggedright
\footnotesize{
\textsuperscript{$\dagger$}\,Official repository uses TensorFlow. For compatibility with TSL, a PyTorch reimplementation of STTN was used: \url{https://github.com/wubin5/STTN/tree/main/Batch_Training_Version}.
}

\end{table}

\section{Weaver Ablation Results} \label{appx:ablations}

The ablation results for Valence attention is shown in Table~\ref{tab:valattention-ablation}.

\begin{table}[width=.99\linewidth,cols=4,pos=h]
    \centering
    \caption{Comparisons of Tanimoto Valence Attention (Tanimoto) versus scaled dot-product attention (SDPA, using softmax) and Cosine coefficient (Cosine) on PemsBay data. Other hyperparameters use the Default(T) hyperparameter set. Differences in trials are due to training runs completed when compiling this work. See Figure~\ref{fig:valattention-ablation} for visualizations.}
    \resizebox{\textwidth}{!}{%
        \begin{tabular}{ccl|cccc|cccc|cccc}
            \toprule\toprule
            \multicolumn{3}{c|}{\textbf{Configuration}} & \multicolumn{4}{c|}{\textbf{RMSE}} & \multicolumn{4}{c|}{\textbf{MAPE}} & \multicolumn{4}{c}{\textbf{MAE}} \\
            \midrule
            \textbf{Test set} & \textbf{Embed size (E)} & \textbf{Kernel} & \textbf{15 min} & \textbf{30 min} & \textbf{60 min} & \textbf{All} & \textbf{15 min} & \textbf{30 min} & \textbf{60 min} & \textbf{All} & \textbf{15 min} & \textbf{30 min} & \textbf{60 min} & \textbf{All} \\
            \midrule\midrule
            \multirow{3}{*}{(a)} & \multirow{3}{*}{\textbf{32}} 
                & Tanimoto (12) 
                & $\unperror{2.87}{0.00}$ & $\unperror{3.77}{0.01}$ & $\unperror{4.44}{0.01}$ & $\unperror{3.66}{0.01}$ 
                & $\unperror{2.86}{0.01}$ & $\unperror{3.83}{0.01}$ & $\unperror{4.74}{0.02}$ & $\unperror{3.67}{0.01}$ 
                & $\unperror{1.36}{0.02}$ & $\unperror{1.69}{0.00}$ & $\unperror{1.86}{0.00}$ & $\unperror{1.62}{0.00}$ \\
                & & SDPA (10) 
                & $\unperror{2.90}{0.00}$ & $\unperror{3.84}{0.01}$ & $\unperror{4.59}{0.01}$ & $\unperror{3.74}{0.01}$ 
                & $\unperror{2.88}{0.01}$ & $\unperror{3.93}{0.01}$ & $\unperror{4.97}{0.02}$ & $\unperror{3.78}{0.01}$ 
                & $\unperror{1.37}{0.00}$ & $\unperror{1.73}{0.00}$ & $\unperror{2.07}{0.01}$ & $\unperror{1.67}{0.00}$ \\
                & & Cosine (10) 
                & $\unperror{3.20}{0.07}$ & $\unperror{4.09}{0.06}$ & $\unperror{4.79}{0.08}$ & $\unperror{3.99}{0.06}$ 
                & $\unperror{3.86}{0.15}$ & $\unperror{4.76}{0.12}$ & $\unperror{5.66}{0.12}$ & $\unperror{4.64}{0.12}$ 
                & $\unperror{1.58}{0.05}$ & $\unperror{1.89}{0.03}$ & $\unperror{2.19}{0.03}$ & $\unperror{1.84}{0.04}$ \\
            \midrule
            \multirow{3}{*}{(b)} & \multirow{3}{*}{\textbf{80}} 
                & Tanimoto (11) 
                & $\unperror{2.82}{0.00}$ & $\unperror{3.70}{0.01}$ & $\unperror{4.30}{0.01}$ & $\unperror{3.60}{0.01}$ 
                & $\unperror{2.79}{0.01}$ & $\unperror{3.68}{0.01}$ & $\unperror{4.45}{0.01}$ & $\unperror{3.51}{0.01}$ 
                & $\unperror{1.33}{0.00}$ & $\unperror{1.63}{0.00}$ & $\unperror{1.89}{0.00}$ & $\unperror{1.56}{0.00}$ \\
                & & SDPA (10) 
                & $\unperror{2.85}{0.00}$ & $\unperror{3.74}{0.01}$ & $\unperror{4.43}{0.01}$ & $\unperror{3.64}{0.01}$ 
                & $\unperror{2.83}{0.01}$ & $\unperror{3.78}{0.01}$ & $\unperror{4.66}{0.01}$ & $\unperror{3.62}{0.01}$ 
                & $\unperror{1.35}{0.00}$ & $\unperror{1.67}{0.00}$ & $\unperror{1.97}{0.00}$ & $\unperror{1.61}{0.00}$ \\
                & & Cosine (10) 
                & $\unperror{2.85}{0.05}$ & $\unperror{3.72}{0.03}$ & $\unperror{4.33}{0.03}$ & $\unperror{3.60}{0.03}$ 
                & $\unperror{2.96}{0.15}$ & $\unperror{3.86}{0.14}$ & $\unperror{4.69}{0.12}$ & $\unperror{3.71}{0.14}$ 
                & $\unperror{1.36}{0.03}$ & $\unperror{1.67}{0.03}$ & $\unperror{1.93}{0.03}$ & $\unperror{1.60}{0.03}$ \\
            \midrule
            \multirow{3}{*}{(c)} & \multirow{3}{*}{\textbf{128}} 
                & Tanimoto (10) 
                & $\unperror{2.79}{0.01}$ & $\unperror{3.67}{0.01}$ & $\unperror{4.27}{0.01}$ & $\unperror{3.55}{0.01}$ 
                & $\unperror{2.77}{0.00}$ & $\unperror{3.63}{0.01}$ & $\unperror{4.36}{0.02}$ & $\unperror{3.46}{0.01}$ 
                & $\unperror{1.32}{0.00}$ & $\unperror{1.61}{0.00}$ & $\unperror{1.85}{0.00}$ & $\unperror{1.54}{0.00}$ \\
                & & SDPA (10) 
                & $\unperror{2.81}{0.00}$ & $\unperror{3.70}{0.01}$ & $\unperror{4.34}{0.01}$ & $\unperror{3.58}{0.01}$ 
                & $\unperror{2.79}{0.00}$ & $\unperror{3.71}{0.01}$ & $\unperror{4.53}{0.01}$ & $\unperror{3.55}{0.01}$ 
                & $\unperror{1.33}{0.00}$ & $\unperror{1.64}{0.00}$ & $\unperror{1.92}{0.00}$ & $\unperror{1.58}{0.00}$ \\
                & & Cosine (10) 
                & $\unperror{2.80}{0.01}$ & $\unperror{3.68}{0.01}$ & $\unperror{4.28}{0.01}$ & $\unperror{3.55}{0.01}$ 
                & $\unperror{2.80}{0.06}$ & $\unperror{3.66}{0.06}$ & $\unperror{4.46}{0.06}$ & $\unperror{3.51}{0.06}$ 
                & $\unperror{1.32}{0.01}$ & $\unperror{1.62}{0.01}$ & $\unperror{1.88}{0.01}$ & $\unperror{1.56}{0.01}$ \\
            \bottomrule\bottomrule
    \end{tabular}
    }%
    \label{tab:valattention-ablation}
\end{table}

\clearpage

\section{KRONECKER PRODUCT and OPERATIONS} \label{appx:Kronecker-product-and-operations}

\paragraph{Kronecker product} The \textbf{Kronecker Product} ($\otimes$, \textbf{KP}) of two arbitrary matrices $\asmatrix{A}\in \mathbb{R}^{I_{A_1} \times I_{A_2}},\, \asmatrix{B}\in\mathbb{R}^{I_{B_1} \times I_{B_2}}$ is defined as:
\begin{align}
    \asmatrix{A} \otimes \asmatrix{B} = 
        \begin{bmatrix}
        a_{11}\asmatrix{B} & a_{12}\asmatrix{B} & \cdots & a_{1I_{A_2}}\asmatrix{B} \\
        a_{21}\asmatrix{B} & a_{22}\asmatrix{B} & \cdots & a_{2I_{A_2}}\asmatrix{B} \\
        \vdots & \vdots & \ddots & \vdots \\
        a_{I_{A_1}1}\asmatrix{B} & a_{I_{A_1}2}\asmatrix{B} & \cdots & a_{I_{A_1}I_{A_2}}\asmatrix{B}
        \end{bmatrix}\,.
\end{align}

\subsection{Kronecker Matrix-Vector product (KMV)}  \label{appx:Kronecker-matrix-vector-definition-KMV}

The \textbf{Kronecker Matrix-Vector product (KMV)} is a useful Kronecker product identity that converts $(\asmatrix{A} \otimes \asmatrix{B}) \vecop(\asmatrix{C})$ to the computationally efficient form $\vecop(\asmatrix{B}\,\asmatrix{C}\,\asmatrix{A}^\top)$, where for some $\asmatrix{C} \in \mathbb{R}^{I_{A_2} \times I_{B_2}}$:
\begin{align}
    \overbrace{ (\asmatrix{A} \otimes \asmatrix{B}) \vecop(\asmatrix{C}) = 
        \underbrace{ \vecop(\asmatrix{B} \asmatrix{C} \asmatrix{A}^\top)}_{\text{KMV vectorization (KMVV)}} }^{\text{Kronecker Matrix-Vector product (KMV)}} \quad \in \mathbb{R}^{ I_{A_1} I_{B_1} \times 1 } \,, 
        \label{eqn:KMV-basic-definition}
\end{align}
and for arbitrary $\asmatrix{D}\in \mathbb{R}^{I_{1} \times I_2}$, the vectorization operation is $\vecop(\asmatrix{D}) := \LLangle\, \asmatrix{D}_{[I_{1} \times I_{2} \,\rightarrow\, I_{2} I_{1} \times 1 ]}  \,\RRangle\,$ and let \textbf{vector folding (VF)} for $\asvector{d}\in \mathbb{R}^{I_2 I_1 \times 1 }$ be
$\vecop^{-1}( \asvector{d} ) := \LLangle\, \asvector{d}_{ [I_2 I_1 \times 1 \,\rightarrow\, I_1 \times I_2] } \,\RRangle\,$.

\subsection{Kronecker graph product in spatiotemporal graph composition}  \label{appx:Kronecker-graph-spatiotemporal}

The KP has long been recognized as a graph composition operator that combines graph structures systematically~\citep{KroneckerGraph:weichsel1962kronecker, KroneckerGraph:leskovec2005realistic}, we demonstrate here that the Kronecker kernel yields a valid spatiotemporal graph admissible to Spatiotemporal Message Passing (STMP) operations.

Let $\graph_S = (\spacenodeset, \set{E}_S),\,\graph_T = (\timenodeset, \set{E}_T)$ define the spatial and temporal graphs, respectively, with corresponding nodes $\omega_{S[i]} \in \Omega_S, \omega_{T[i]} \in \Omega_T$ and edges $\varepsilon_{S[i,j]} \in \set{E}_{S}, \, \varepsilon_{T[i,j]} \in \set{E}_T$, where $\varepsilon_{[i,j]}$ is a pairwise connection from node $\omega_i$ to $\omega_j$. 
By definition~\citep{KroneckerGraph:weichsel1962kronecker,KroneckerGraph:leskovec2010kronecker}, the KP of two graphs, $\mathcal{G}_{T} \otimes \mathcal{G}_{S}$, is the KP of their respective adjacency matrices, $\adjmat_S = A(\graph_{S})$ and $\adjmat_T = A(\graph_{T})$:
\begin{align}
    \asmatrix{A}(\graph_{ST}) = \asmatrix{A}(\graph_T) \otimes \asmatrix{A}(\graph_S) := \adjmat_{T} \otimes \adjmat_{S}, \label{eqn:kronecker-graph-definition}
\end{align}
which directly determines the edge structure in the spatiotemporal graph $\graph_{ST}$.

Having established that the Kronecker product generates valid spatiotemporal graphs, we leverage this result by applying graph-generating dot-product kernels (G-DPKs) (Equation~\eqref{eqn:graph-generating-dot-G-DPK}) to extend the definition in Equation~\eqref{eqn:kronecker-graph-definition} to Spatiotemporal attention maps (STAMs):
\begin{align}
    \Theta( \graph_{ST} ) &\approx \Theta( \kten{} ) = \Theta(\graph_T) \otimes \Theta(\graph_S) := \attn_T \otimes \attn_S\,, \label{eqn:spatiotemporal-graph-attn-form}
\end{align}
to establish that the Kronecker attention kernel (Equation~\eqref{eqn:kronecker-attention-kernel}), proposed earlier via KPA, generates valid spatiotemporal graphs through the Kronecker product of the spatial ($\attn_S$) and temporal ($\attn_T$) attention maps.

Following~\citet{KroneckerGraph:leskovec2010kronecker}, the structure of edges in $\graph_{ST}$ is determined by the edges in the constituent spatial and temporal graphs. For spatial nodes $\omega_{S[\varsigma_1]}, \omega_{S[\varsigma_2]}$ and temporal nodes $\omega_{T[\tau_1]}, \omega_{T[\tau_2]}$:
\begin{align}
    ( \varepsilon_{ST\,[\varsigma_1\tau_1,\varsigma_2\tau_2]} \in \,\graph_{ST}) \,\iff\, 
    ( \varepsilon_{S[\varsigma_1,\varsigma_2]} \in \graph_{S}) \,\cap \, 
    ( \varepsilon_{T[\tau_1,\tau_2]} \in \graph_{T})\,.
\end{align}
In other words, spatiotemporal edges exist if and only if both constituent spatial and temporal edges exist. Here, $\omega_{ST[\varsigma_1\tau_1]},\,\omega_{ST [\varsigma_2\tau_2]} \in \Omega_{ST}$ are spatiotemporal nodes indexed by ordered pairs $[\varsigma_1\tau_1]$, $[\varsigma_2\tau_2]$ and connected by edge $\varepsilon_{ST\,[\varsigma_1\tau_1,\varsigma_2\tau_2]}$. This establishes that the Kronecker product of $\graph_S$ and $\graph_T$ yields a valid spatiotemporal graph $\graph_{ST}=(\Omega_{ST},\,\set{E}_{ST})$ with structural correspondence between constituent and spatiotemporal graph elements.

\section{PROPERTIES of WEAVER COMPONENTS} \label{appx:weaver-component-details}

\subsection{W-iKPS: Relaxation of Kronecker Separability via Head-mixing Weights} \label{appx:weighted-KPS}
We demonstrate with a short proof that a \textbf{weighted implicit Kronecker product summation (W-iKPS)} is performed when linear projection is applied via the head-mixing weight $\asmatrix{W}_O$ (Equation~\eqref{eqn:multihead_mixhead}). Consequently, without additional mechanisms, this implies that the spatiotemporal attention map $\attnst$ does not necessarily have to conform to a separable Kronecker structure for P-KMV to be applicable.
To establish a link between multihead and KPS, we start with a canonical form of KPS on $\attnspace$ and $\attntimes$ using multihead indexing:
\begin{align}
    \attnst &\approx \sum_{h=1}^{H}{(\attntimes^{(h)} \otimes \attnspace^{(h)})}\,, \\
    \attnst\,\asmatrix{V} &\approx \sum_{h=1}^{H}{(\attntimes^{(h)} \otimes \attnspace^{(h)})}\, \asmatrix{V}^{(h)} \,.
\end{align}
To match the P-KMV form, this can be re-written in as tensors batched on the head dimension, first in the unconsolidated form where $\gtensor{\Theta}_{\spaceset\timesset} \in \mathbb{R}^{H\times PN \times PN}$, $\gtensor{\Theta}_{\timesset} \in \mathbb{R}^{H\times N \times N}$, $\gtensor{\Theta}_{\spaceset} \in \mathbb{R}^{H\times P \times P}$, and  $\tensor{V}\in\mathbb{R}^{H \times PN \times E}$:
\begin{align}
    \gtensor{\Theta}_{\spaceset \timesset} \tensor{V} &\approx ( \gtensor{\Theta}_{\timesset} \otimes \gtensor{\Theta}_{\spaceset} )\,\tensor{V}\, \quad \in \mathbb{R}^{H \times PN \times E}.
\end{align}
Then KPS is applied by summation on $H$ in tensor notation:
\begin{align}
    \attnst\,\asmatrix{V} &\approx ( \gtensor{\Theta}_{\timesset} \otimes \gtensor{\Theta}_{\spaceset} )\,\tensor{V}\, \times_{1} \asvector{1}\, \quad \in \mathbb{R}^{PN \times E},  \label{eqn:multihead-KPS-tensor-appx} 
\end{align}
with $\asvector{1} = [1,...,1] \in \mathbb{R}^{H}$ a vector of ones, while $\times_1$ denotes the \textit{mode-1 tensor product} (i.e. on $H$, see \citet{TensorOps:Kolda2009}). 

Applying a standard graph propagation framework (refer to Equation~\eqref{eqn:graph-convolution-basic} for example) to Equation~\eqref{eqn:multihead-KPS-tensor-appx}, we may treat each head as independent message passing (MP) on graph variants. Hence, one possible way of applying the feature-mixing step uses \textbf{head-specific weight matrices} $\asmatrix{W}^{(h)} \in \mathbb{R}^{E \times E'}$:
\begin{align}
    \sum_{h=1}^{H}{(\attntimes^{(h)} \otimes \attnspace^{(h)})}\, \asmatrix{V}^{(h)} \asmatrix{W}^{(h)} &= \Big( ( \gtensor{\Theta}_{\timesset} \otimes \gtensor{\Theta}_{\spaceset} )\,\tensor{V} \,\tensor{W}_{H} \Big)\, \times_{1} \asvector{1}\, \quad \in \mathbb{R}^{PN \times E} , \label{eqn:WKPS-PerHeadWeighting}
\end{align}
where $\tensor{W}_H \in \mathbb{R}^{H \times E \times E'}$ stacks $\asmatrix{W}^{(h)}$ on $H$. Setting $E' = HE$, we obtain $\tensor{W}_H \in \mathbb{R}^{H \times E \times HE}$, which corresponds to $H \cdot E \cdot HE = H^2E^2$ parameters—matching the capacity of the head-mixing weight $\asmatrix{W}_O \in \mathbb{R}^{HE \times HE}$. 

The RHS of Equation~\eqref{eqn:WKPS-PerHeadWeighting} makes it clearer when applying head-mixing weights $\asmatrix{W}_O$ using tensor operations:
\begin{alignat}{2}
    \midtilde{\tensor{Z}} &= (\gtensor{\Theta}_{\timesset} \otimes \gtensor{\Theta}_{\spaceset} )\,\tensor{V} \quad \in \mathbb{R}^{H \times PN \times E}\,, \\
    \tensor{Z} &= \Langle \midtilde{\tensor{Z}}_{[H \times PN \times E \rightarrow PN \times HE]} \Rangle\, \quad \in \mathbb{R}^{PN \times HE} &\quad&\text{(Tensor rearrangement)}, \\
    \Big( ( \gtensor{\Theta}_{\timesset} \otimes \gtensor{\Theta}_{\spaceset} )\,\tensor{V} \,\tensor{W}_{H} \Big)\, \times_{1} \asvector{1}
    &\cong \tensor{Z}\,\asmatrix{W}_{O} \,, \label{eqn:WiKPS_head-mixing-appx}
\end{alignat}
where the congruency in Equation~\eqref{eqn:WiKPS_head-mixing-appx} comes from the fact that linear combinations can be written as other linear combinations. Thus, by transitive property of equality:
\begin{align}
    \sum_{h=1}^{H}{(\attntimes^{(h)} \otimes \attnspace^{(h)})}\, \asmatrix{V}^{(h)} \asmatrix{W}^{(h)} &\cong \tensor{Z}\,\asmatrix{W}_{O}\,,
\end{align}
which demonstrates that application of head-mixing weights $\asmatrix{W}_O$ in $\tensor{Z}\asmatrix{W}_O$ is an \textit{\textbf{implicit}} version of the head-specific \textit{weighted Kronecker product summation} (W-KPS) in Equation~\eqref{eqn:WKPS-PerHeadWeighting}. Therefore, $\tensor{Z}\asmatrix{W}_O$ is a \textbf{Weighted Implicit KPS (W-iKPS)}.

\subsection{Characteristics of the Continuous Tanimoto Coefficient (CTC)} \label{appx:tanimoto-analysis-main}

\subsubsection{Local Stability Analysis of CTC Critical Points} \label{appx:tanimoto-critical-points}

The Jacobian of Equation~\eqref{eqn:definition-tanimoto-vector-qk} with respect to (\textit{wrt.}) $\asvector{q}$ is derived via quotient rule and set to $\boldsymbol{0}$ to determine the critical points (roots):
\begin{align}
    \nabla_{\asvector{q}} \frac{\asvector{q}^\top \asvector{k}}{ \Vert \asvector{q} \Vert^2 + \Vert \asvector{k} \Vert^2 - \asvector{q}^\top \asvector{k} } 
    &= \frac{ ( \asvector{q}^\top \asvector{q} + \asvector{k}^\top \asvector{k} )\,\asvector{k} - 2(\asvector{q}^\top \asvector{k})\,\asvector{q} }{ (  \asvector{q}^\top \asvector{q} + \asvector{k}^\top \asvector{k} - \asvector{q}^\top \asvector{k} )^2 } = \boldsymbol{0} \,,  \label{eqn:ctc-analysis1} \\
    &\Rightarrow 2(\asvector{q}^\top \asvector{k})\,\asvector{q}  = ( \asvector{q}^\top \asvector{q} + \asvector{k}^\top \asvector{k} )\,\asvector{k}\,, \label{eqn:ctc-analysis2}
\end{align}
where $\boldsymbol{0} \,\in\, 0^{E \times 1}$ i.e., a vector of zeroes, and using the identity $\Vert\asvector{a}\Vert^2 = \asvector{a}^\top \asvector{a}$. Here, Equation~\eqref{eqn:ctc-analysis2} is feasible \textit{iff} $\asvector{q}$ and $\asvector{k}$ are \textbf{\textit{collinear vectors}}, i.e.: $\asvector{k} = c\,\asvector{q}$ for any $c \in \mathbb{R}$. Hence, substituting for $\asvector{k}= c\,\asvector{q}$ to Equation~\eqref{eqn:ctc-analysis2}:
\begin{alignat}{4}
    (c^2 - 1)\,\asvector{q} &= \boldsymbol{0} 
    &&\Rightarrow \quad &&c = \pm 1, \label{eqn:ctc-analysis3} \\
    c &= \pm 1 
    &&\Rightarrow \quad &&\asvector{q} = \pm\,\asvector{k}, 
    &&\text{(Collinearity condition; roots 1 and 2).} \label{eqn:ctc-analysis4}
\end{alignat}

To characterize the local behavior at the critical points, we perform a second-order analysis. Let $\asvector{q} = \asvector{q}_* + \delta\asvector{q}$ where $\asvector{q}_*$ denotes a critical point and $\Vert\delta\asvector{q}\Vert$ is small. Equation~\eqref{eqn:ctc-analysis2} is restated with $\asvector{q}, \asvector{k}\in \mathbb{R}^{E \times 1}$ and assuming $\asvector{k}$ is constant:
\begin{align*}
    2(\asvector{q}^\top \asvector{k})\,\asvector{q}  = ( \asvector{q}^\top \asvector{q} + \asvector{k}^\top \asvector{k} )\,\asvector{k}\,,
\end{align*}
To derive the Hessian, let $N = \asvector{q}^\top \asvector{k}$ and $D = \Vert\asvector{q}\Vert^2 + \Vert\asvector{k}\Vert^2 - \asvector{q}^\top \asvector{k}$. We can rewrite the gradient as:
\begin{equation}
    \nabla T = \frac{\asvector{g}}{D^2} \quad \text{where} \quad \asvector{g} = (D + N)\asvector{k} - 2N\asvector{q}\,,
\end{equation}
The Hessian is obtained by applying the quotient rule:
\begin{equation}
    \asmatrix{H} = \frac{\partial}{\partial \asvector{q}}\left(\frac{\asvector{g}}{D^2}\right) = \frac{1}{D^2}\frac{\partial \asvector{g}}{\partial \asvector{q}} - \frac{2\asvector{g}}{D^3}\left(\frac{\partial D}{\partial \asvector{q}}\right)^\top\,,
\end{equation}
where:
\begin{align}
    \frac{\partial D}{\partial \asvector{q}} &= 2\asvector{q} - \asvector{k}\,\,, \quad 
    \frac{\partial \asvector{g}}{\partial \asvector{q}} = \asvector{k}(2\asvector{q} - \asvector{k})^\top + \asvector{k}\asvector{k}^\top - 2\asvector{k}\asvector{q}^\top - 2(\asvector{q}^\top \asvector{k})\asmatrix{I}\,,
\end{align}
and $\asmatrix{I}$ is the $E \times E$ identity matrix.

\paragraph{Root 1: $\asvector{q}_* = \asvector{k}$.}
At this critical point, $N = \Vert\asvector{k}\Vert^2$, $D = \Vert\asvector{k}\Vert^2$, and $\asvector{g} = \asvector{0}$. Since $\asvector{g}$ vanishes, the Hessian simplifies to:
\begin{equation}
    \asmatrix{H}\bigg|_{\asvector{q}=\asvector{k}} = \frac{1}{D^2}\frac{\partial \asvector{g}}{\partial \asvector{q}}\bigg|_{\asvector{q}=\asvector{k}} = \frac{-2\Vert\asvector{k}\Vert^2 \asmatrix{I}}{\Vert\asvector{k}\Vert^4} = -\frac{2\asmatrix{I}}{\Vert\asvector{k}\Vert^2} \,,
\end{equation}
The Hessian is negative definite (all eigenvalues equal $-2/\Vert\asvector{k}\Vert^2 < 0$), confirming that $\asvector{q}_* = \asvector{k}$ is a \textbf{local maximum} with Tanimoto coefficient value $T = 1$.

\paragraph{Root 2: $\asvector{q}_* = -\asvector{k}$.}
At this critical point, $N = -\Vert\asvector{k}\Vert^2$, $D = 3\Vert\asvector{k}\Vert^2$, and $\asvector{g} = \asvector{0}$. Evaluating the Hessian:
\begin{equation}
    \frac{\partial \asvector{g}}{\partial \asvector{q}}\bigg|_{\asvector{q}=-\asvector{k}} = \asvector{k}(-3\asvector{k})^\top + \asvector{k}\asvector{k}^\top + 2\asvector{k}\asvector{k}^\top + 2\Vert\asvector{k}\Vert^2 \asmatrix{I} = 2\Vert\asvector{k}\Vert^2 \asmatrix{I} \,.
\end{equation}
Therefore:
\begin{equation}
    \asmatrix{H}\bigg|_{\asvector{q}=-\asvector{k}} = \frac{2\Vert\asvector{k}\Vert^2 \asmatrix{I}}{9\Vert\asvector{k}\Vert^4} = \frac{2\asmatrix{I}}{9\Vert\asvector{k}\Vert^2} \,.
\end{equation}
The Hessian is positive definite (all eigenvalues equal $2/(9\Vert\asvector{k}\Vert^2) > 0$), confirming that $\asvector{q}_* = -\asvector{k}$ is a \textbf{local minimum} with Tanimoto coefficient value $T = -1/3$. The stability of these critical points depends on the optimization objective.

\textbf{Under maximization of $\Theta_{\text{CTC}}$ (gradient ascent: $\dot{\asvector{q}} = \nabla T$):}
\begin{itemize}
    \item $\asvector{q}_* = \asvector{k}$: Negative definite Hessian $\Rightarrow$ \textbf{stable equilibrium} (attractor).
    \item $\asvector{q}_* = -\asvector{k}$: Positive definite Hessian $\Rightarrow$ \textbf{unstable equilibrium} (repeller).
\end{itemize}

\textbf{Under minimization of $\Theta_{\text{CTC}}$ (gradient descent: $\dot{\asvector{q}} = -\nabla T$):}
\begin{itemize}
    \item $\asvector{q}_* = \asvector{k}$: Negative definite Hessian $\Rightarrow$ \textbf{unstable equilibrium} (repeller).
    \item $\asvector{q}_* = -\asvector{k}$: Positive definite Hessian $\Rightarrow$ \textbf{stable equilibrium} (attractor).
\end{itemize}
 
\subsubsection{CTC and Positive Edge Dominance} \label{appx:CTC-positive-dominance}
We provide a sketch of proof illustrating how the asymmetric CTC range promotes stable signed network dynamics through positive edge dominance. 
Consider a simplified message-passing process with score vector $\boldsymbol{\theta}$ and feature vector $\mathbf{z}$, where we assume $\theta_i \sim \text{Bernoulli}(p)$ with non-stochastic $z_i = z_j$ for all $i \neq j$, representing a worst-case boundary scenario. 
The transformation $\left(\tfrac{4}{3}\theta_i - \tfrac{1}{3}\right)$ remaps Bernoulli samples from $\{0,1\}$ to the CTC range $\{-\tfrac{1}{3}, 1\}$. 
Hence, the message $M$ is computed as:
\begin{align} 
M &= \boldsymbol{\theta}^\top \mathbf{z} = \sum_{i=1}^{N} \left( \tfrac{4}{3}\theta_i - \tfrac{1}{3} \right) z_i.
\end{align}
The expected message magnitude is
\begin{equation}
\mathbb{E}[M] = \frac{4p - 1}{3}\sum_{i=1}^{N} z_i,
\end{equation}
which remains positive for $p > 0.25$ when $\sum_i z_i > 0$. 
This reflects an intrinsic bias toward positive edges, aligning with structural balance theory in signed networks~\citep{DynSignedNetworks:Shi2018DynamicsOS}, which posits that stable signed networks exhibit positive edge dominance. 

In contrast, a symmetric $[-1,1]$ range (e.g., from $\tanh$ or Cosine) obtained via $(2\theta_i - 1)$ for $\{0,1\} \rightarrow \{-1,1\}$ yields $\mathbb{E}[M] = (2p - 1)\sum_i z_i$, which crosses zero at $p = 0.5$. 
This symmetry introduces balanced positive–negative competition that can destabilize early signal propagation. 
By biasing expectations toward cooperation ($p>0.25$), the CTC formulation naturally favors stable, predominantly positive connectivity patterns—consistent with empirical findings in traffic networks where correlations between adjacent road segments are dominantly positive~\citep{trafficstudy:ermagun2017using}. 

% Our derivation assumes deterministic $\mathbf{z}$. 
% Under fully stochastic conditions where $\mathbf{z}$ is random (e.g., $z_i \sim \mathcal{N}(\mu_i,\sigma_i^2)$ in deep networks~\citep{DNNGaussian:schoenholz2017deep}), this positive bias remains but its variance increases, making stability considerations even more critical. 
% A comprehensive stochastic analysis is left for future work.
Since this derivation assumes deterministic $\mathbf{z}$, a comprehensive analysis under fully stochastic conditions where $\mathbf{z}$ is random (e.g., $z_i \sim \mathcal{N}(\mu_i,\sigma_i^2)$ in deep networks~\citep{DNNGaussian:schoenholz2017deep}) is left for future work.

\section{Generalization of P$^\Delta$-KMV} \label{appx:generalized-pkmv-main}
This section derives the higher-order generalization of P$^2$-KMV: P$^\Delta$-KMV, by extending a multi-modal ($\Delta>2$) KMV-vectorization~(Equation~\eqref{eqn:3Dkmvv1}--\eqref{eqn:3Dkmvv4}) and resolving it by inferring the necessary tensor rearrangements via induction (Equations~\eqref{eqn:kmvv-arrange1}--\eqref{eqn:kmvv-arrange4}). We follow directly from the basic P$^2$-KMV in Equations~\eqref{eqn:P2KMV_basic1}--\eqref{eqn:P2KMV_basic2} from Section~\ref{section:problem1-kten}, replacing $\asvector{v}_e, \asmatrix{V}_e$ with $\asvector{z}_e, \asmatrix{Z}_e$ for generalizable notation. This yields the \textbf{right-handed P-KMV ((R)-P-KMV) forms} computable via recursive tensor rearrangement, enabling GPU optimizations and other computational advantages. 

\subsection{Nested multi-modal hierarchical Kronecker-TEN ($\Delta>2$)} 

We begin by describing extensions of the Kronecker-TEN framework from Section~\ref{section:problem1-kten} to nested Kronecker hierarchical graphs, which provides consistent notation for subsequent derivations. Let $\nodeset = \{\node_\ell\}_{\ell=1}^L$ be a \textbf{node-set} with cardinality $R=|\nodeset|$, and \textbf{node index} $\ell\in [L]$. We define the \textbf{hierarchical order} of node-sets as $\nodeset^\hi \succ \nodeset \succ \nodeset^\lo$, where $\nodeset^\hi$ and $\nodeset^\lo$ are the \textit{immediate superordinate} and \textit{immediate subordinate} to $\nodeset$, respectively. Overloading the definition of a \textbf{node} $m_\ell$ as a hyperedge~\citep{Hypergraph:antelmi2023representation}, each node contains itself and its subordinate node-set:
\begin{align}
    m_\ell = \{ \ell, \asvector{u}_{\ell}, \nodeset^\lo | \nodeset \}\,,
\end{align}
where $\asvector{u}_{\ell} \in \mathbb{R}^E$ are the \textbf{routing vectors}, and $\{\cdot\,| \nodeset\}$ denotes \textbf{node-set membership}. This allows compact representations of fully connected, recursively nested Kronecker hierarchical graphs.

To define $\kten{\Delta}$ of depth-$\Delta$, we now further refine $\nodeset$ with \textbf{hierarchical-depth indexing} $\delta \in \{0,1,\ldots, \Delta\}$ and introduce the \textbf{hierarchical rule} $\mathcal{D}_H$ defining the hierarchy of \textit{node-sets ordered by decreasing abstraction}: 
\begin{align}
    % \mathcal{D}_H = ( \, \Omega^{[0]} \, \succ \,\cdots\, \succ \nodeset^{[\delta-1]} \succ \nodeset^{[\delta]} \succ \nodeset^{[\delta+1]} \succ \cdots\, \succ \{1\}^{[\Delta+1]} \,)\,\, ,
    \mathcal{D}_H = ( \, \nodeset^{[1]} \, \succ \,\cdots\, \succ \nodeset^{[\delta-1]} \succ \nodeset^{[\delta]} \succ \nodeset^{[\delta+1]} \succ \cdots\, \succ \{1\}^{[\Delta+1]} \,)\,\, ,
\end{align}
% where $\latentset^{[0]}=\latentset$ is the "latent node-set" representing the highest abstraction (the "information layer"), and $1\in\{1\}$ is the identity set for mathematical completeness. 
where $1\in\{1\}$ is the identity set indicating the presence of a node, for completeness. 

% Our \textbf{propagation rule} on $\kten{\Delta}$ for latents $\asvector{z}$ follows from 

Thus, the corresponding $\Delta$\textbf{-hierarchical KMV} (\hkmv{\Delta}) representation of $\mathcal{D}_H$ is:
% Thus, recursive application of MP1 and MP2 from Section~\ref{section:problem1-kten} to $\mathcal{D}_H$ yields the \textit{$\Delta$-hierarchical KMV} (\hkmv{\Delta}) form, for which P$^2$-KMV ($\Delta=2$) is a special case:
\begin{align}
    \asvector{z}'_e &= \left(\, \Theta_{\nodeset^{[1]}} \otimes \Theta_{\nodeset^{[2]}} \cdots \, \otimes\, \Theta_{\nodeset^{[\Delta]}} \otimes\, 1 \, \right) \asvector{z}_e\,  \, \, \, =\left(\bigotimes_{\delta=1}^{\Delta+1}\, \Theta_{\,\delta}\,\right)\,\asvector{z}_e\, , \label{eqn:mono_prop_rule}
\end{align}
where $\Theta_{\nodeset^{[\delta]}}$ are attention maps at depth-$\delta$, with $\Theta_{\nodeset^{[\Delta+1]}}:= 1$ for completeness. 
% The hierarchical nature of \hdelta-KMV becomes apparent after depth-1 ($d=1$) KMV-vectorization (Section~\ref{section:problem1-kten}), with nested structure emerging $d=2,3,\ldots,\Delta$.

\subsubsection{Hierarchical KMV-vectorization (KMVV).} Using superscript $\depth{\delta}$ to indicate KMV-vectorization depth on $\asvector{z}$, we can thus define two equivalent \textbf{hierarchical KMVV} patterns of Equation~\eqref{eqn:mono_prop_rule} for $\Delta>2$:
\begin{alignat}{2}
    \asvector{z}'_e &= \left( \Theta_{\nodeset^{[1]}}\, \bigotimes_{\delta=2}^{\Delta+1}\,\Theta_{\nodeset^{[\delta]}} \right)\, \asvector{z}_e\, \,
    &&= \vecop \Bigg( \left[ \bigotimes_{\delta=2}^{\Delta+1}\,\Theta_{\nodeset^{[\delta]}} \right]\, \asmatrix{Z}^{\depth{1}}_e\, \Theta_{\nodeset^{[1]}}^\top \Bigg)\, \,\,, \label{eqn:KMV-rightrecur}  \\
    \nonumber\\
    \asvector{z}'_e &= \left( \bigotimes_{\delta=1}^{\Delta} \Theta_{\nodeset^{[1]}}\, \otimes\,\Theta_{\nodeset^{[\Delta+1]}} \right)\, \asvector{z}_e\,\, 
    &&= \vecop\Bigg( \Theta_{\nodeset^{[\Delta+1]}} \, \left[ \, \bigotimes_{\delta=1}^{\Delta}\,\Theta_{\nodeset^{[\delta]} } \cdot (\asmatrix{Z}_e^{\depth{1}} )^\top \right]^\top \Bigg)\,\,\,, \label{eqn:KMV-leftrecur}
\end{alignat}
where $\asvector{z}_e = \vecop(\asmatrix{Z}^{\depth{1}}_e)$ or equivalently, $\vecop^{-1}(\asvector{z}_e) = \asmatrix{Z}^{\depth{1}}_e$. We refer to Equation~\eqref{eqn:KMV-rightrecur} as the "right-handed KMVV" (\hrightrecur) and \eqref{eqn:KMV-leftrecur} as the "left-handed KMVV" (\hleftrecur), each resulting in slightly different tensor operations when resolved via P-KMV. Put simply, using \hrightrecur\ avoids the nested transpose operation in \hleftrecur\ and retains the characteristic KMV form $(\asmatrix{A}\,\otimes\,\asmatrix{B})\,\asvector{z}_e$ in subsequent recursions, making it more intuitive in practice. Hence, we shall apply this in Appendix~\ref{section:PKMV3} when deriving the generalized P-KMV.

\subsection{The Right-Handed 3D P-KMV ((R)-P$^3$-KMV) and generalized (R)-P$^\Delta$-KMV} \label{section:PKMV3}

\subsubsection{The (R)-P$^3$-KMV}
For the 3D problem, we assume the Kronecker adjacency map, $\asmatrix{K}_3$ as follows using the \textit{right-handed KMVV} ((R)-KMVV) form in Equation~\eqref{eqn:KMV-rightrecur}:
\begin{align}
    \asmatrix{K}_3 &= ( \Theta_{\nodeset^{[1]}} \otimes \Theta_{\nodeset^{[2]}} \otimes \Theta_{\nodeset^{[3]}} \otimes \Theta_{\nodeset^{[4]}} )\,
    = \Theta_{\nodeset^{[1]}} \bigotimes_{\delta=2}^{\Delta+1}\,\Theta_{\nodeset^{[\delta]}}\,\,,
\end{align}
We first fully expand the \textbf{3D hierarchical KMVV} ($\kappa_{R}^{3}$) from Equation~\eqref{eqn:KMV-rightrecur}:
\begin{align}
    \asmatrix{K}_3\,\asmatrix{Z}^{\depth{0}} 
    &= \Bigg( \left[ \bigotimes_{\delta=2}^{\Delta+1=4}\,\Theta_{\nodeset^{[\delta]}} \right] \asmatrix{Z}^{\depth{1}}\,\Theta_{\nodeset^{[1]}}^\top \Bigg), \label{eqn:3Dkmvv1}  \\
    % = \bigg( \left[ \Theta_{\nodeset^{[2]}} \otimes \Theta_{\nodeset^{[3]}} \otimes \Theta_{\nodeset^{[4]}} \right] \asmatrix{Z}^{\depth{1}}\,\Theta_{\nodeset^{[1]}}^\top \bigg) ,  \\
    \left[ \bigotimes_{\delta=2}^{4}\,\Theta_{\nodeset^{[\delta]}} \right] \asmatrix{Z}^{\depth{1}} 
    &= \Bigg( \left[ \bigotimes_{\delta=3}^{4}\,\Theta_{\nodeset^{[\delta]}} \right] \asmatrix{Z}^{\depth{2}}\,\Theta_{\nodeset^{[2]}}^\top \Bigg) , \label{eqn:3Dkmvv2} \\
    % = \bigg( \left[ \Theta_{\nodeset^{[3]}} \otimes \Theta_{\nodeset^{[4]}} \right] \asmatrix{Z}^{\depth{2}}\,\Theta_{\nodeset^{[2]}}^\top \bigg) , 
    \left[ \bigotimes_{\delta=3}^{4}\,\Theta_{\nodeset^{[\delta]}} \right] \asmatrix{Z}^{\depth{2}} 
    &= \Bigg(  \Theta_{\nodeset^{[4]}} \, \asmatrix{Z}^{\depth{3}}\,\Theta_{\nodeset^{[3]}}^\top \Bigg)  \label{eqn:3Dkmvv3}  \\
    % = \Bigg(  1 \cdot \, \asmatrix{Z}^{\depth{3}}\,\Theta_{\nodeset^{[3]}}^\top \Bigg)  ,
    \Theta_{\nodeset^{[4]}} \asmatrix{Z}^{\depth{3}} 
    &= \Bigg(  \asmatrix{Z}^{\depth{4}}\,\Theta_{\nodeset^{[4]}}^\top \Bigg) = \asmatrix{Z}^{\depth{4}} \label{eqn:3Dkmvv4}
\end{align}
where $\Theta_{\nodeset^{\depth{4}}} := 1$, $|\Theta_{\nodeset^{\depth{4}}}|=1$ is the identity mapping necessary for deriving the general case. Here, we can infer the dimensions of $\asmatrix{Z}^{\depth{\delta}}$ from their corresponding matrix multiplications, 
i.e.: $\asmatrix{Z}^{\depth{0}}\in\mathbb{R}^{\prod_{\delta=1}^{\Delta+1}I_\delta \times I_0 }$, $\asmatrix{Z}^{\depth{1}}\in\mathbb{R}^{\prod_{\delta=2}^{\Delta+1}I_\delta\times I_{1}}$, $\asmatrix{Z}^{\depth{2}}\in\mathbb{R}^{\prod_{\delta=3}^{\Delta+1}I_\delta\times I_{2}}$, and $\asmatrix{Z}^{\depth{3}}\in\mathbb{R}^{I_4 \times I_{3}}$. 

To resolve the $\kappa_{R}^{3}$ without interrupting the isomorphism of $\tensor{Z}^{\depth{\delta}}$, we can extend Equation~\eqref{eqn:P2KMV_basic1} by induction to infer the necessary tensor rearrangements for Equations~\eqref{eqn:3Dkmvv1}--\eqref{eqn:3Dkmvv4}. Specifically, to get from $d=0$ to $d=1$, we require: 
% \begin{align}
%     \tensor{Z}^{\depth{1}} &= \LLLangle  \tensor{Z}^{\depth{0}}_{ \big[\,\prod_{\delta=1}^{4}I_\delta \times I_0 \,\rightarrow\, I_0 \times (\prod_{\delta=2}^{4} I_\delta \times I_1) \, \big] }  \RRRangle
% \end{align}
\begin{align}
    \tensor{Z}^{\depth{0}}\in\mathbb{R}^{\prod_{\delta=1}^{4}I_\delta \times I_0 } \, \mspace{53mu}
    \rightarrow \, \tensor{Z}^{\depth{1}}\in \mathbb{R}^{I_0 \times (\prod_{\delta=2}^{4} I_\delta \times I_1)}, \label{eqn:kmvv-arrange1}
\end{align}
and extending this to subsequent depths ($\delta>1$) resolves to:
\begin{alignat}{2}
    &\tensor{Z}^{\depth{1}}\in\mathbb{R}^{I_0 \times ( \prod_{\delta=1}^{4}I_\delta \times I_1 ) } \, &&\rightarrow \, \tensor{Z}^{\depth{2}}\in \mathbb{R}^{I_0 \times (I_1 \times \prod_{\delta=3}^{4} I_\delta \times I_2)}, \label{eqn:kmvv-arrange2}  \\
    &\tensor{Z}^{\depth{2}}\in \mathbb{R}^{I_0 \times (I_1 \times \prod_{\delta=3}^{4} I_\delta \times I_2)} \, &&\rightarrow \,\tensor{Z}^{\depth{3}}\in \mathbb{R}^{I_0 \times (I_1 \times (I_2 \times I_4 \times I_3))}, \label{eqn:kmvv-arrange3} \\
    &\tensor{Z}^{\depth{3}}\in \mathbb{R}^{I_0 \times (I_1 \times (I_2 \times I_4 \times I_3))} \, &&\rightarrow \, \tensor{Z}^{\depth{4}}\in \mathbb{R}^{I_0 \times (I_1 \times (I_2 \times (I_3 \times I_4)))}. \label{eqn:kmvv-arrange4}
\end{alignat}

Notably, if we assume homogenous attention maps $\Theta_{\nodeset^{\depth{\delta}}}$ at every $\delta$, certain dimensional stratifications "$\times$" do not specialize information in the tensor (see Section~\ref{appx:tensor-operations} on BMM) and can be removed to avoid redundancy while improving batched tensor operations on the GPU~\citep{NVIDIA2023CUDA}. Put simply, we may write instead:
\begin{align}
    \tensor{Z}^{\depth{1}} &\in \mathbb{R}^{I_0 \times (\prod_{\delta=2}^{4} I_\delta \times I_1)}\, , \label{eqn:p3kmv-rearrange1} \\
    \tensor{Z}^{\depth{2}} &\in \mathbb{R}^{ \prod_{\delta=0}^{1} I_{\delta} \times (\prod_{\delta=3}^{4} I_{\delta} \times I_2) }\, , \label{eqn:p3kmv-rearrange2} \\
    \tensor{Z}^{\depth{3}} &\in \mathbb{R}^{ \prod_{\delta=0}^{2} I_{\delta} \times (I_4 \times I_3) }\,, \label{eqn:p3kmv-rearrange3} \\
    \tensor{Z}^{\depth{4}} &\in \mathbb{R}^{ \prod_{\delta=0}^{3} I_{\delta} \times I_4 }\,. \label{eqn:p3kmv-rearrange4}
\end{align}

Thus, we can obtain the \textbf{P$^3$-KMV} as the solution to $\kappa_{R}^{3}$ by resolving Equations~\eqref{eqn:3Dkmvv1}--\eqref{eqn:3Dkmvv4} in reverse order using tensor rearrangements (Kronecker-Tumble ($\dotumble$)) from Equations~\eqref{eqn:p3kmv-rearrange1}--\eqref{eqn:p3kmv-rearrange4} while applying the corresponding attention maps. In other words, the P$^3$-KMV is fully stated below:
\begin{alignat}{2}
    \tensor{U}^{\depth{3}} &=  (\tensor{Z}^{\depth{4}} \cdot 1 )_{\dotumble}
    &&\in\, \mathbb{R}^{ \prod_{\delta=0}^{2} I_{\delta} \times (I_4 \times I_3) } , \label{eqn:3DKtumbl1} \\
    \tensor{U}^{\depth{2}} &= (\tensor{U}^{\depth{3}}  \Theta^\top_{\nodeset^{\depth{3}}})_{\dotumble}
    &&\in\, \mathbb{R}^{ \prod_{\delta=0}^{1} I_{\delta} \times (\prod_{\delta=3}^{4} I_{\delta} \times I_2) }, \label{eqn:3DKtumbl2} \\
    \tensor{U}^{\depth{1}} &= (\tensor{U}^{\depth{2}}  \Theta^\top_{\nodeset^{\depth{2}}})_{\dotumble}
    &&\in\, \mathbb{R}^{I_0 \times (\prod_{\delta=2}^{4} I_\delta \times I_1)}, \label{eqn:3DKtumbl3}  \\
    \asmatrix{K}_3 \, \tensor{Z}^{\depth{0}} = \tensor{U}^{\depth{0}} &= (\tensor{U}^{\depth{1}}  \Theta^\top_{\nodeset^{\depth{1}}})_{\dotumble}
    &&\in\, \mathbb{R}^{\prod_{\delta=1}^{4}I_\delta \times I_0 } . \label{eqn:3DKtumbl4}
\end{alignat}

\subsubsection{The generalized (R)-P$^{\Delta}$-KMV and Kronecker-Tumble}
Although Equation~\eqref{eqn:3DKtumbl1} is an identity map, it enables the \textbf{general form of (R)-P$^\Delta$-KMV}:
\begin{align}
    \tensor{U}^{\depth{d}} &=  \left( \tensor{U}^{\depth{d+1}}\,\Theta_{\nodeset^{\depth{d+1}}}^\top \right)_{\dotumble}\,.
\end{align}
Induction is applied to infer the \textbf{general form of K-Tumble} ($\dotumble$) from Eqn.~\eqref{eqn:3DKtumbl1}--\eqref{eqn:3DKtumbl4} as the tensor rearrangement:
\begin{align}
    \LLangle \tensor{A}^{\depth{d+1}} \RRangle_{\dotumble} 
    &= 
    \LLLangle
    \tensor{A}^{\depth{d+1}}_{{\left[\prod_{\alpha=0}^{d} I_\alpha\times\, \prod_{\delta=d+2}^{\Delta+1} I_{\delta}\times \, I_{d+1} \,\rightarrow\, \prod_{\alpha=0}^{d-1} I_\alpha\times\, \prod_{\delta=d+1}^{\Delta+1} I_{\delta}\times \, I_d \right]}\,} 
    \RRRangle\, = \tensor{A}^{\depth{d}} \,. \label{eqn:kronecker-tumble-general}
\end{align}

\subsubsection{The Multihead (MH) (R)-P$^{\Delta}$-KMV }
Extending (R)-P$^{\Delta}$-KMV to multihead architectures requires a key modification. Standard (R)-P$^{\Delta}$-KMV assumes a homogeneous attention map at each depth, but multihead mechanisms learn distinct mappings per head $h \in [H]$. This is simply addressed by preserving $H$ as an independent mode in all (R)-P$^{\Delta}$-KMV and K-Tumble operations:
\begin{align}
    \tensor{U}_H^{\depth{d}} &= \left( \tensor{U}^{\depth{d+1}}_H\,\Theta_{H \, \,\nodeset^{\depth{d+1}}}^\top \right)_{\dotumble^H} 
\end{align}
And correspondingly, the multihead K-Tumble is also stratified on $H$:
\begin{align}
    \LLangle \tensor{A}_H^{\depth{d+1}}\RRangle_{\dotumble} &= \LLLangle
    \tensor{A}^{\depth{d+1}}_{H\,{\,{\left[H \times \prod_{\alpha=0}^{d} I_\alpha \times \, \prod_{\delta=d+2}^{\Delta+1} I_{\delta} \times \, I_{d+1} \,\rightarrow\,  H \times \prod_{\alpha=0}^{d-1} I_\alpha \times\, \prod_{\delta=d+1}^{\Delta+1} I_{\delta} \times \, I_d  \right]}}\,} \RRRangle = \tensor{A}_H^{\langle d \rangle} \,.
\end{align}

\subsection{Comparative GPU Performance on P$^2$-KMV (Basic vs Efficient)} \label{appx:P-KMV-extra-comparisons}

\begin{figure}[width=.99\linewidth,cols=4,pos=t]
    \centering

    \begin{subfigure}{0.80\linewidth}
        \centering
        \includegraphics[width=\linewidth]{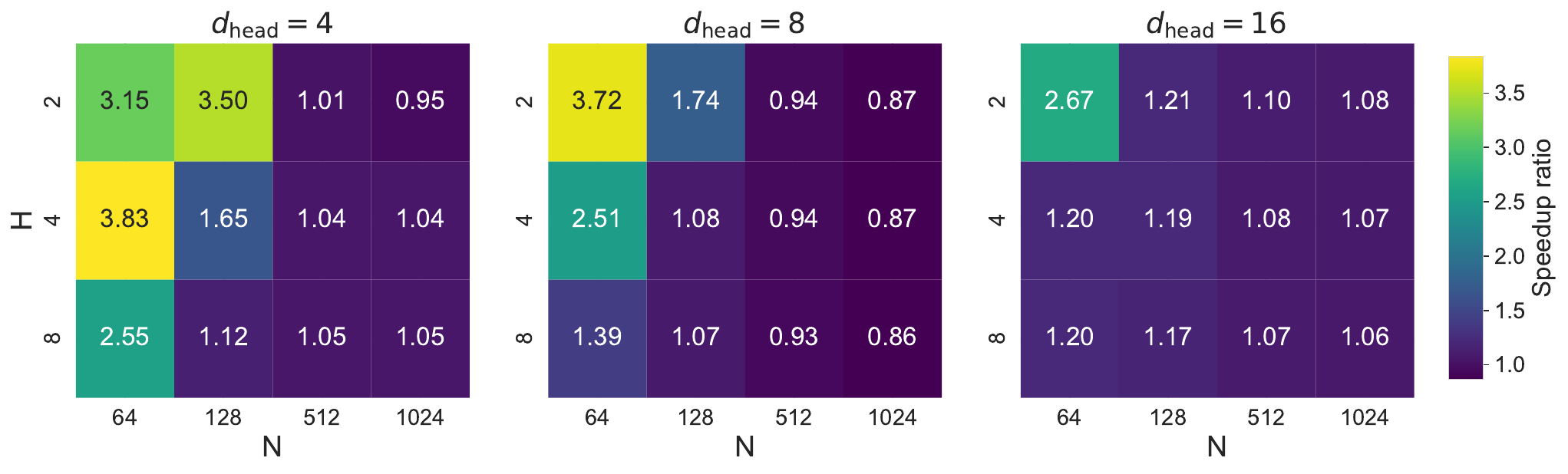}
        \caption{T4 GPU}
        \label{fig:pkmv-speedup-t4}
    \end{subfigure}

    \vspace{1em}

    \begin{subfigure}{0.80\linewidth}
        \centering
        \includegraphics[width=\linewidth]{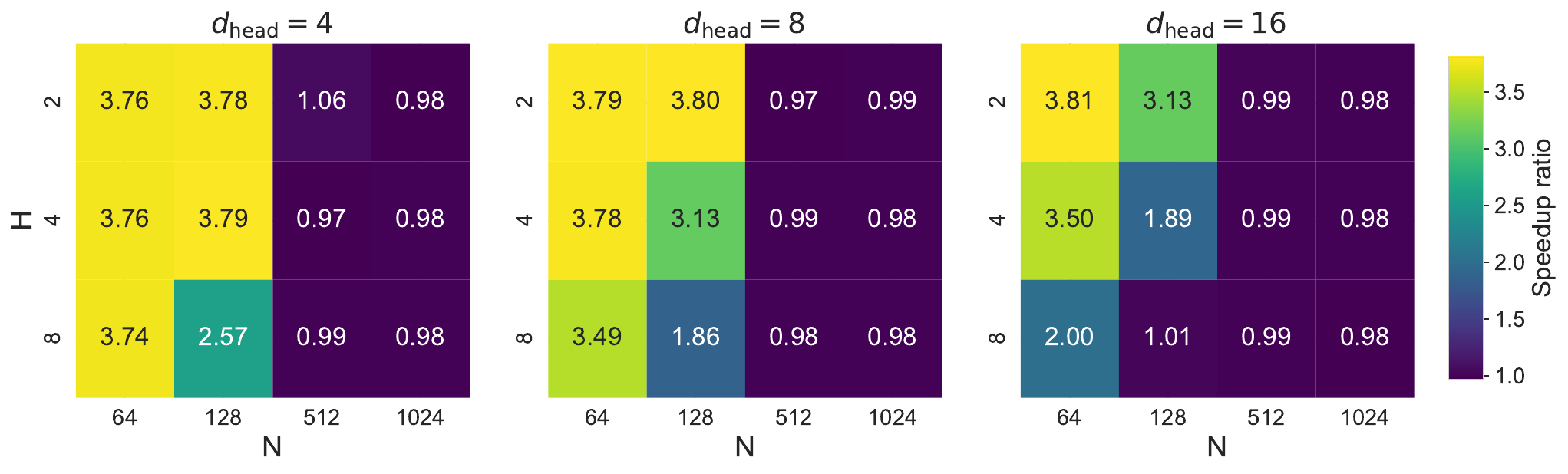}
        \caption{A100 GPU}
        \label{fig:pkmv-speedup-a100}
    \end{subfigure}

    \vspace{1em}

    \begin{subfigure}{0.80\linewidth}
        \centering
        \includegraphics[width=\linewidth]{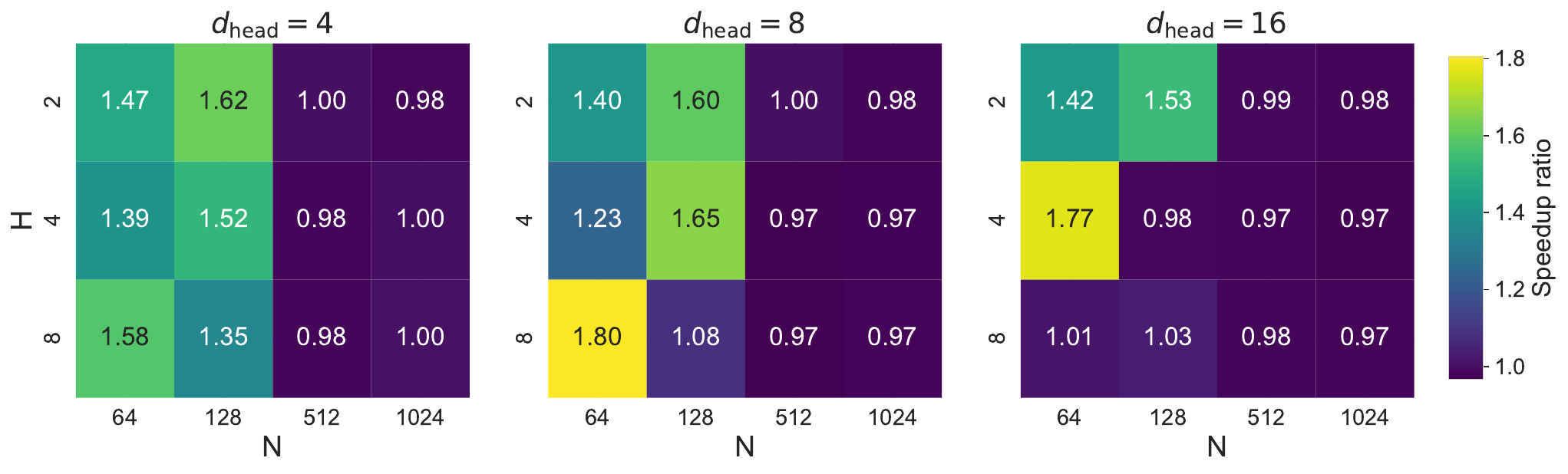}
        \caption{RTX 4090}
        \label{fig:pkmv-speedup-4090}
    \end{subfigure}

    \caption{Speedup heatmaps for the basic vs.\ efficient P$^2$-KMV kernel, computed relative to the same GPU.}
    \label{fig:pkmv-speedups}
\end{figure}

This section compares multihead tensor computation performance between the basic (Equations~\eqref{eqn:P2KMV_basic1}--\eqref{eqn:P2KMV_basic2}) and efficient (Equations~\eqref{eqn:PKMV-efficient1}--\eqref{eqn:PKMV-efficient3}) P$^2$-KMV kernels across GPUs with distinct architectural profiles. The NVIDIA T4 (2018; 16\,GB GDDR6, $\sim$320\,GB/s memory bandwidth) and A100 (2020; 40/80\,GB HBM2e, $\sim$1.5--2.0\,TB/s memory bandwidth) are datacenter--class accelerators optimized for tensor computations, whereas the RTX~4090 (2022; 24\,GB GDDR6X, $\sim$1.0\,TB/s memory bandwidth) is a consumer GPU optimized primarily for graphics and rendering. In terms of raw memory throughput and tensor performance, the devices rank approximately as \textbf{A100} $>$ \textbf{RTX~4090} $>$ \textbf{T4}.

Heatmap values report the ratio of execution times, $\text{Speedup} = t_{\text{Basic}} / t_{\text{Efficient}}$, where values $> 1.0$ indicate that the efficient P$^2$-KMV kernel executes faster. Execution times $t_{\text{Basic}}$ and $t_{\text{Efficient}}$ are averaged over 10{,}000 trials.

\paragraph{Same-GPU comparisons.}
Figure~\ref{fig:pkmv-speedups} presents speedups measured relative to the basic kernel on the \emph{same} GPU. Substantial gains are demonstrated for small-to-medium problems: $(H=2,d_{\text{head}}=4, N=64)$--($H=8,d_{\text{head}}=8, N=128$) across all GPUs, especially on GPUs optimized for tensor operations (T4, A100). The subsequent reduction to $\times1.00$ gains suggest bandwidth saturation in the GPU architecture.

\begin{figure}[width=.99\linewidth,cols=4,pos=t]
    \centering

    \begin{subfigure}{0.80\linewidth}
        \centering
        \includegraphics[width=\linewidth]{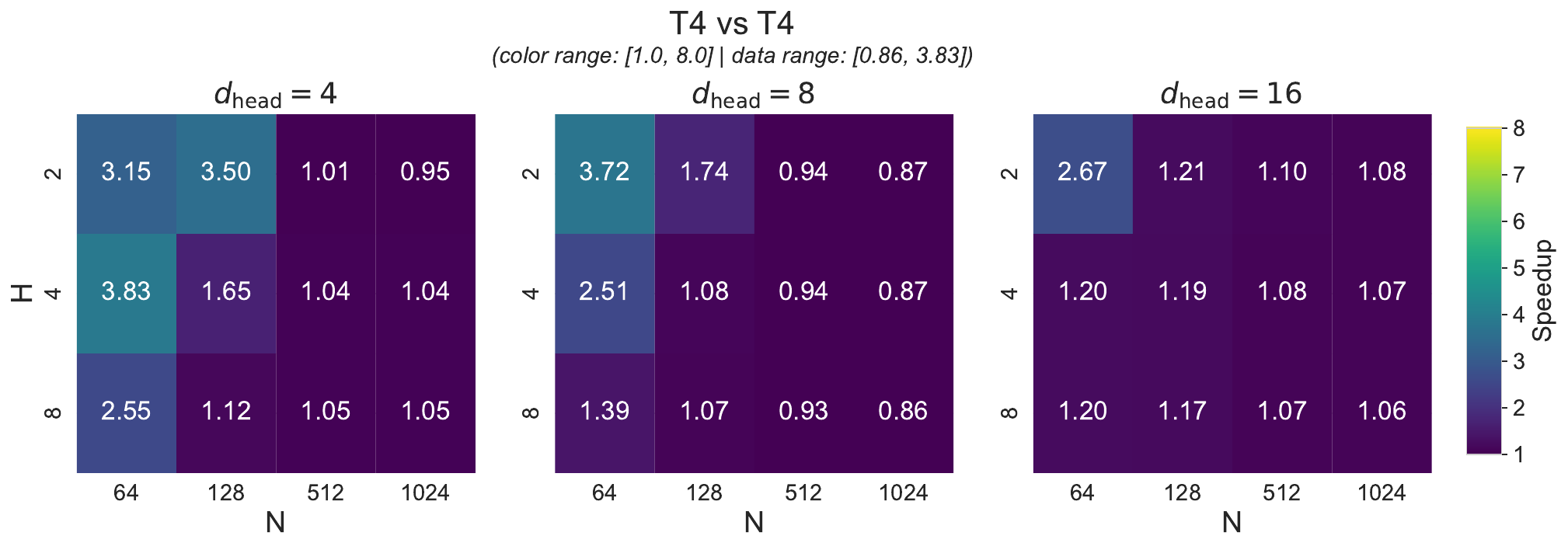}
        \caption{T4 vs T4 GPU}
        \label{fig:pkmv-speedup-t4}
    \end{subfigure}

    \vspace{1em}

    \begin{subfigure}{0.80\linewidth}
        \centering
        \includegraphics[width=\linewidth]{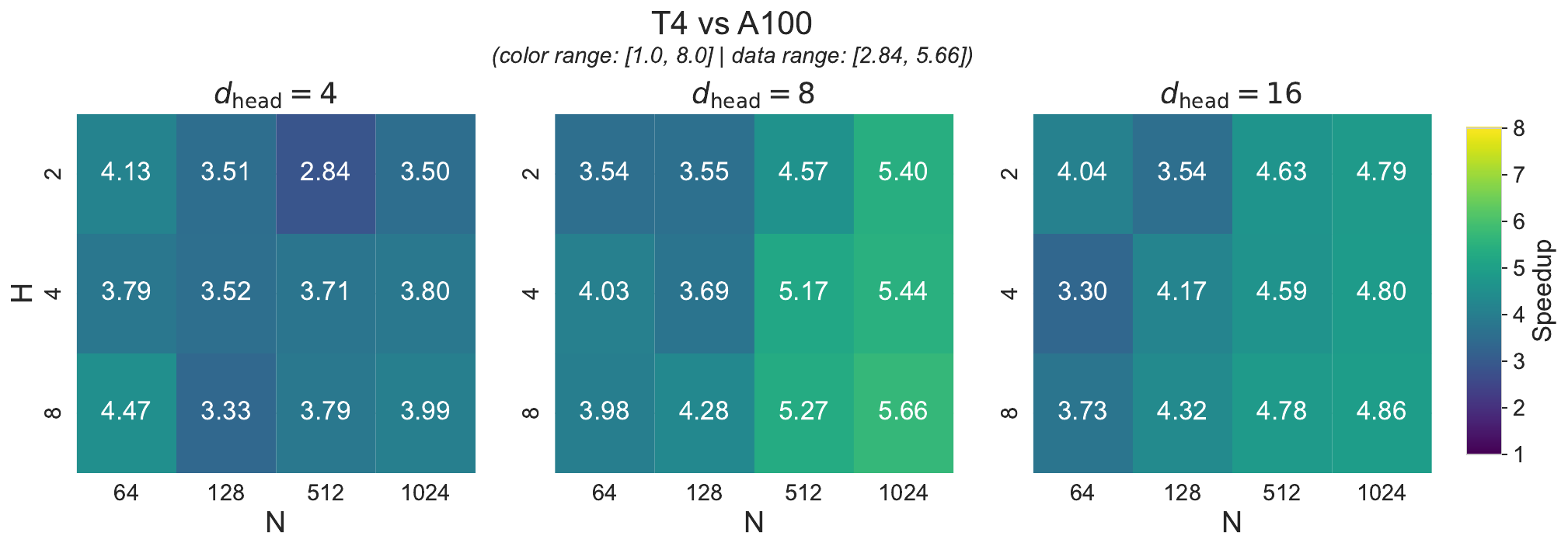}
        \caption{T4 vs A100 GPU}
        \label{fig:pkmv-speedup-a100}
    \end{subfigure}

    \vspace{1em}

    \begin{subfigure}{0.80\linewidth}
        \centering
        \includegraphics[width=\linewidth]{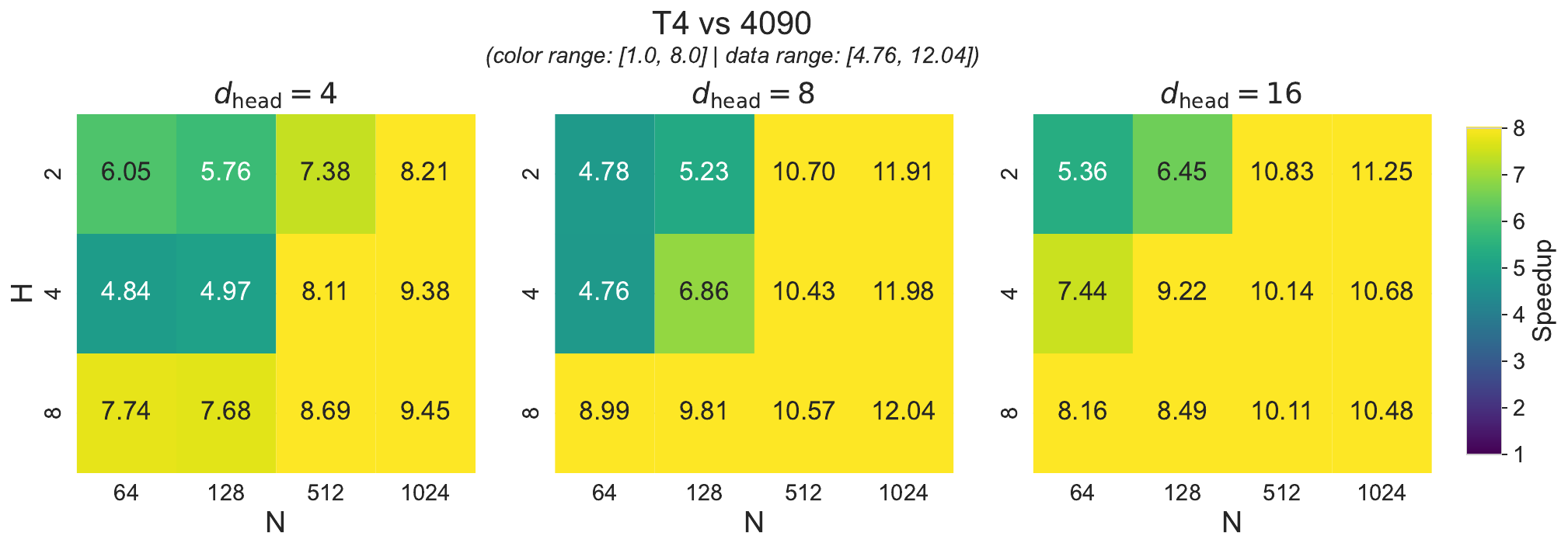}
        \caption{T4 vs RTX 4090}
        \label{fig:pkmv-speedup-4090}
    \end{subfigure}

    \caption{Speedup heatmaps for the basic vs efficient P$^2$-KMV kernel across different GPUs using T4 as the control. Color maps are adjusted for uniform comparison.}
    \label{fig:pkmv-speedups-T4control}
\end{figure}

\paragraph{Comparisons relative to T4.}
Figure~\ref{fig:pkmv-speedups-T4control} normalizes all results to the T4 as a common reference, illustrating how performance scales on newer architectures: the observed gains reflect both the efficient kernel and improved hardware capability. On the A100 and RTX~4090, the normalized speedups remain high even for larger problem sizes (e.g., beyond $H=8$, $d_{\text{head}}=8$, $N=128$), suggesting that the efficient P$^2$-KMV kernel continues to deliver benefits as tensor dimensions grow. It is worth noting that part of the apparent gain at larger scales may stem from the T4 degrading more rapidly than A100 or RTX~4090 at higher workloads; therefore, the figure should be interpreted as a relative scaling comparison, whereas absolute kernel efficiency is captured in the same-GPU results of Figure~\ref{fig:pkmv-speedups}.

\end{document}